\documentclass[Afour,sageh,times]{sagej}

\usepackage{moreverb,url}

\usepackage[colorlinks,bookmarksopen,bookmarksnumbered,citecolor=black,urlcolor=black]{hyperref}

\usepackage[utf8]{inputenc} 
\usepackage[T1]{fontenc}    
\usepackage{hyperref}       
\usepackage{url}            
\usepackage{booktabs}       
\usepackage{amsfonts}       
\usepackage{amsmath}
\usepackage{nicefrac}       
\usepackage{microtype}      
\usepackage[ruled]{algorithm2e}
\usepackage{xcolor}
\usepackage[nolist,nohyperlinks]{acronym}
\SetArgSty{textnormal}
\usepackage{graphicx}
\usepackage{caption}
\usepackage{subcaption}
\usepackage[export]{adjustbox}
\usepackage{float}
\usepackage{balance}
\usepackage{soul}
\usepackage[normalem]{ulem}

\newcommand{\sysstate}{\ensuremath{s}}
\newcommand{\action}{a}

\newcommand{\reward}{R}
\newcommand{\weight}{\ensuremath{\theta}}
\newcommand{\wspace}{\ensuremath{\Theta}}
\newcommand{\statefeat}{\ensuremath{\phi}}
\newcommand{\trajfeat}{\ensuremath{\Phi}}
\newcommand{\trace}{\ensuremath{\xi}}
\newcommand{\traj}{\ensuremath{\tau}}

\begin{acronym}
\acro{IRL}[IRL]{Inverse Reinforcement Learning}
\acro{FERL}{Feature Expansive Reward Learning}
\acro{ME-IRL}{Maximum Entropy Inverse Reinforcement Learning}
\acro{MDP}{Markov Decision Process}
\acro{POMDP}{Partially Observable Markov Decision Process}
\acro{MAP}{Maximum A Posterior}
\acro{EE}{End-Effector}
\acro{MSE}{Mean-Squared-Error}
\acro{GT}{Ground Truth}
\acro{GFL}{Guided Feature Learning}
\end{acronym}

\newcommand{\change}[1]{\textcolor{black}{#1}}
\newcommand{\changenew}[1]{\textcolor{black}{#1}}

\newcommand{\strike}[1]%
    {}

\newcommand\BibTeX{{\rmfamily B\kern-.05em \textsc{i\kern-.025em b}\kern-.08em
T\kern-.1667em\lower.7ex\hbox{E}\kern-.125emX}}

\def\volumeyear{2022}

\begin{document}

\runninghead{Bobu et. al.}

\title{Inducing Structure in Reward Learning by Learning Features}

\author{Andreea Bobu\affilnum{1}, Marius Wiggert\affilnum{1}, Claire Tomlin\affilnum{1}, and Anca D. Dragan\affilnum{1}}

\affiliation{\affilnum{1}University of California, Berkeley}

\corrauth{Andreea Bobu, Department of Electrical Engineering and Computer Science, University of California Berkeley, 2121 Berkeley Way, Berkeley, CA 94709}

\email{abobu@berkeley.edu}

\begin{abstract}
Reward learning enables robots to learn adaptable behaviors from human input. 
Traditional methods model the reward as a linear function of hand-crafted features, but that requires specifying all the relevant features \textit{a priori}, which is impossible for real-world tasks. To get around this issue, recent deep \ac{IRL} methods learn rewards directly from the raw state but this is challenging because the robot has to implicitly learn the features that are important \emph{and} how to combine them, simultaneously. Instead, we propose a divide and conquer approach:
focus human input specifically on learning the features separately, and only then learn how to combine them into a reward.
We introduce a novel type of human input for teaching features and an algorithm that utilizes it to learn complex features from the raw state space. The robot can then learn how to combine them into a reward using demonstrations, corrections, or other reward learning frameworks. We demonstrate our method in settings where all features have to be learned from scratch, as well as where some of the features are known. By first focusing human input specifically on the feature(s), our method decreases sample complexity and improves generalization of the learned reward over a deep \ac{IRL} baseline. We show this in experiments with a physical 7DOF robot manipulator, as well as in a user study conducted in a simulated environment.

\end{abstract}

\keywords{learning from humans, inverse reinforcement learning, feature learning}

\maketitle

\section{Introduction}

\begin{figure*}[h]
\includegraphics[width=0.99\textwidth]{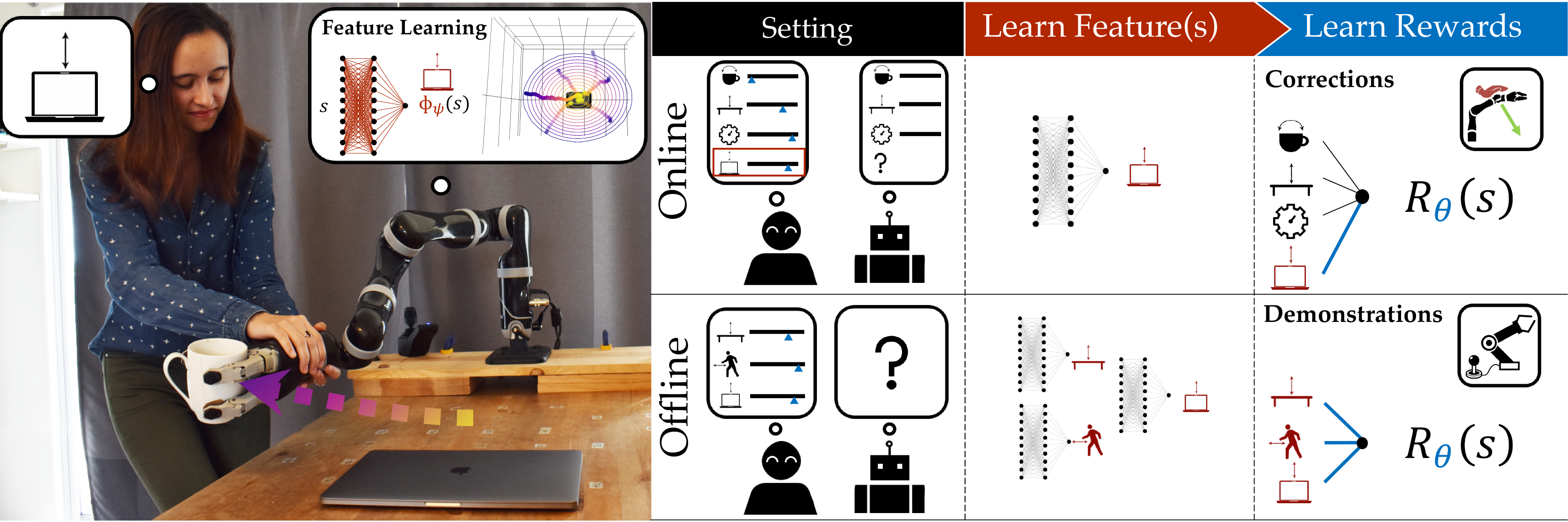}
\centering
\caption{(Left) The person teaches the robot the concept of \change{horizontal} distance from the laptop by providing a few feature traces. (Right-Top) In the online reward learning from corrections setting, once the robot detects that its feature set is incomplete, it queries the human for feature traces that teach it the missing feature and adapts the reward to account for it. (Right-Bottom) In the offline reward learning from demonstrations setting, the person has to teach the robot each feature separately one at a time using feature traces, and only then teach their combined reward.}
\label{fig:front_fig}
\end{figure*}

Whether it's semi-autonomous driving~\citep{sadigh2016infogather}, recommender systems~\citep{ziebart2008maximum}, or household robots working in close proximity with people~\citep{jain2015learning}, reward learning can greatly benefit autonomous agents to generate behaviors that adapt to new situations or human preferences.
Under this framework, the robot uses the person's input to learn a reward function that describes how they prefer the task to be performed.
For instance, in the scenario in Fig. \ref{fig:front_fig}, the human wants the robot to keep the cup away from the laptop to prevent spilling liquid over it; she may communicate this preference to the robot by providing a demonstration of the task or even by directly intervening during the robot's task execution to correct it.
After learning the reward function, the robot can then optimize it to produce behaviors that better resemble what the person wants.


In order to correctly interpret and efficiently learn from human input, traditional methods resorted to structuring
the reward as a (linear) function of carefully hand-engineered \textit{features} -- important aspects of the task \citep{ziebart2008maximum,abbeel2004apprenticeship,jain2015learning,bajcsy2017phri}.
Unfortunately, selecting the right space of features is notoriously challenging, even for expert system designers: knowing and specifying \textit{a priori} an exhaustive set of \emph{all} the features that might be relevant for the reward is impossible for most real-world tasks.
To bypass this feature specification problem, state-of-the-art deep \ac{IRL} methods~\citep{wulfmeier2016maxentirl, finn2016gcl,brown2020brex} learn rewards defined directly on the high-dimensional raw state (or observation) space, thereby implicitly constructing features automatically from task demonstrations.

In doing so, however, these approaches sacrifice the sample efficiency and generalizability that a well-specified feature set offers.
While using an expressive function approximator to extract features and learn their reward combination at once seems advantageous, many such functions can induce policies that explain the demonstrations.
Hence, to disambiguate between all these candidate functions, the robot requires a very large amount of (laborious to collect) data, and this data needs to be diverse enough to identify the true reward. 
For example, the human in the household robot setting in Fig. \ref{fig:front_fig} might want to demonstrate keeping the cup away from the laptop\change{, but from a single demonstration} 
\strike{Without a lot of demonstrations starting from different parts of the state space,} 
the robot could find many other explanations for the person's behavior: perhaps they always happened to keep the cup upright or they really like curved trajectories in general.

The underlying problem here is that demonstrations -- or task-specific input more broadly -- \change{are meant to teach the robot about the reward and not about the features per se, so these function approximators struggle to capture the right feature structure for the reward.}
\strike{Since demonstrations can then only provide implicit information about the features, these function approximators will inevitably need a lot of diverse data to capture the right feature structure for the reward.}  
In this work, we argue that the robot does not have to learn everything at once; instead, it can \emph{divide-and-conquer} the reward learning problem and focus on explicitly learning the features separately from learning how to combine them into the reward.
In our earlier example, if the robot were taught about the concept of distances to laptops separately, it would be able to quickly tell what the person wants from a single demonstration.

\strike{We introduce a new type of human input that is specifically designed to teach features, along with an algorithm for learning feature functions from it. We then showcase how the robot can incorporate feature learning to better capture human preferences in two scenarios. In an online reward learning setting like the one in Fig. \ref{fig:front_fig} (Right-Top), the robot already knows some features and only adds new ones when it detects something is missing. In the more challenging offline reward learning case in Fig. \ref{fig:front_fig} (Right-Bottom), the person has to teach the robot every feature as well as their reward combination.}

We make the following contributions:

\noindent\textbf{Learning features from a novel type of human input.} 
We present a method for learning complex non-linear features separately from the reward (Sec. \ref{sec:feature_learning}). We introduce a new type of human input \change{specifically designed to teach features}, which we call feature traces -- partial trajectories that describe the monotonic evolution of the value of the feature to be learned. To provide a feature trace, the person guides the robot from states where the feature is highly expressed to states where it is not, in a monotonic fashion. Looking at Fig. \ref{fig:front_fig} (Left), the person teaches the robot to avoid the laptop by giving a few feature traces
: she starts with the arm above the laptop and moves it away until comfortable with the distance from the object. We present an algorithm that harvests the structure inherent to feature traces and uses it to efficiently learn a feature relevant for the reward: in our example, the \change{horizontal} distance from the laptop. In experiments on a 7-DoF robot arm, we find that our method can learn high quality features closely resembling the ground truth (Sec. \ref{sec:feature_expert_exps}).

\noindent\textbf{Demonstrating our feature learning method in a user study on a simulated 7-DoF robot arm.} 
In a user study with the JACO2 (Kinova) robotic arm, we show that non-expert users can use our approach for learning features (Sec. \ref{sec:feature_user_exps}). The participants were able to provide feature traces to teach good features, and found our teaching protocol intuitive. 
Unfortunately, due to the current pandemic, we conducted the study online in a simulated environment; despite the inevitable degradation in input quality that this entails, the users were still able to teach features \change{that induced informative bias}.

\noindent\textbf{Analyzing generalization and sample complexity benefits of learning features for rewards.}
\change{We show how our method, which we call \ac{FERL} because it expands the feature set one by one, can improve reward learning sample complexity and generalization.}
First, we look at \change{an easier online reward learning setting like the one in Fig. \ref{fig:front_fig} (Right-Top)} where the robot knows part of the feature set from the get-go, but the person's preference also depends on other features not in the set (Sec. \ref{sec:online_FERL}). We show that, by learning the missing feature, the robot obtains a more generalizable reward than if it had trained a deep \ac{IRL} network directly from the raw state and the known set (Sec. \ref{sec:online_FERL_exps}). 
We then consider \change{the more challenging offline reward learning case in Fig. \ref{fig:front_fig} (Right-Bottom)} where the person teaches the reward from scratch, one feature at a time (Sec. \ref{sec:offline_FERL}). We find that the robot outperforms the baseline most of the time, with less clear results when the learned features are noisily taught by novice users in simulation (Sec \ref{sec:offline_FERL_exps}).


We note that this work is an extension of \citet{bobu2021ferl}, which was published at the International Conference on Human Robot Interaction. We build on this work by formalizing a general framework for feature-based reward learning, and instantiating it in a new offline learning setting where the person can teach each feature one by one before combining them into a reward. Not only is this offline setting more commonly encountered in reward learning, but it also showcases that our approach can be applied more generally to preference learning from different kinds of human input.

Overall, this work provides evidence that taking a divide-and-conquer approach focusing on learning important features separately before learning the reward improves sample complexity in reward learning. Although showcased in manipulation, our method can be used in any robot learning scenarios where feature learning is beneficial: in collaborative manufacturing users might care about the rotation of the object handed over, or in autonomous driving passengers may care about how fast to drive through curves.

\section{Related Work}
\label{sec:related}

Programming robot behavior through human input is a well-established paradigm. In this paradigm, the robot receives human input and aims to infer a policy or reward function that captures the behavior the human wants the robot to express. In imitation learning, the robot directly learns a policy that imitates demonstrations given by the human \citep{osa2018algorithmic}. The policy learns a \textit{correlation} between situations and actions but not \textit{why} a specific behavior is desirable. Because of that, imitation learning only works in the training regime whereas optimizing a learned reward, which captures \textit{why} a behavior is desirable, can generalize to unseen situations \citep{abbeel2004apprenticeship}. 

In the \ac{IRL} framework the robot receives demonstrations through teleoperation \citep{javdani2015shared,abbeel2004apprenticeship} or kinesthetic teaching \citep{argall2009survey} and learns a reward under which these demonstrations are optimal \citep{russell2002artificial, abbeel2004apprenticeship}. Recent research goes beyond demonstrations, utilizing other types of human input for reward learning such as corrections \citep{jain2015learning,bajcsy2017phri}, comparisons \citep{christiano2017preferences} and rankings \citep{brown2019extrapolating}, examples of what constitutes a goal \citep{fu2018variational}, or even specified proxy objectives \citep{HadfieldMenell2017InverseRD}. Depending on the interaction setting, the human input can be given all-at-once, iteratively, or on specific requests of the robot in an active learning setting \citep{lopes2009active, brown2018risk, sadigh2016infogather}.

All these methods require less human input if a parsimonious representation of the world, which summarizes raw state information in the form of relevant features, is available. This is because finite feature sets significantly reduce the space of possible functions which according to statistical learning theory reduces the information complexity of the learning problem \citep{vapnik2013nature}. In the following we discuss the the role of feature representations in reward learning and methods for learning features.

\subsection{Feature Representations in Reward Learning}
 Traditional reward learning methods rely on a set of carefully hand-crafted features that capture aspects of the environment a person may care about. These are selected by the system designer prior to the task~\citep{ziebart2008maximum,abbeel2004apprenticeship,jain2015learning,HadfieldMenell2017InverseRD,bajcsy2017phri}. If chosen well, this feature set introduces an inductive bias that enables the algorithms to find a good estimate of the human's preferences with limited input. Unfortunately, selecting such a set in the first place is notoriously challenging, even for experts like system designers. For one, defining a good feature function can be a time consuming trial-and-error process, especially if the feature is meant to capture a complex aspect of the task~\citep{wulfmeier2016maxentirl}. Moreover, the chosen feature space may not be expressive enough to represent everything that a person might want (and is giving input about) \citep{bobu2020quantifying, haug2018teaching}. When this is the case, the system may misinterpret human guidance, perform unexpected or undesired behavior, and degrade in overall performance \citep{amodei2016, russell2002artificial, haug2018teaching}.

To tackle these challenges that come with hand-designing a feature set, state-of-the-art deep \ac{IRL} methods use the raw state space directly and shift the burden of extracting behavior-relevant aspects of the environment onto the function approximator \citep{finn2016gcl,wulfmeier2016maxentirl}. The objective of \ac{IRL} methods is to learn a reward which induces behavior that matches the state expectation of the demonstrations. The disadvantage of such approaches is that they require large amounts of highly diverse data to learn a reward function which generalizes across the state space. This is because with expressive function approximators there exists a large set of functions that could explain the human input, i.e. many reward functions induce policies that match the demonstrations' state expectation. The higher dimensional the state, the more human input is needed to disambiguate between those functions sufficiently to find a reward function which accurately captures human preferences and thereby generalizes to states not seen during training and not just replicates the demonstrations' state expectations. Thus, when venturing sufficiently far away from the demonstrations the learned reward in \ac{IRL} does not generalize which can lead to unintended behavior ~\citep{Reddy2020SQILIL,fu2018learning}.

It has been shown that providing linear feature functions as human input can reduce the risk of unintended behavior \citep{haug2018teaching}. In our work we argue that generalization with limited input can be achieved without requiring hand-crafted features if the robot explicitly learns features, instead of attempting to learn them implicitly from demonstrations.


\subsection{Learning Features}
In \ac{IRL} researchers have explored the direction of inferring a set of relevant features directly from task demonstrations. This can take the form of joint Bayesian inference on both reward and feature parameters \citep{choi2013bayesian} or projecting the raw state space to lower dimensions via PCA on demonstrated trajectories \citep{vernaza2012efficient}. 
\changenew{There are also methods that add features iteratively to learn a non-linear reward, such as \citet{levine2010feature}, which constructs logical conjunctions of primitive integer features,
and \citet{ratliff2007boosting}, which trains regression trees to distinguish expert from non-expert trajectories in a base feature space. 
\citet{levine2010feature} performs well in discrete-state MDPs, but is not suitable for continuous state spaces, does not operate on raw states but rather a hand-engineered set of integer component features, and requires the reward structure to be expressible as logical conjunctions.
Meanwhile, \citet{ratliff2007boosting} allows for larger state spaces and arbitrary continuous rewards, but still relies on engineering a relevant set of base features and severely underperforms in the case of non-expert human input when compared to more recent IRL techniques~\citep{levine2011nonlinear, wulfmeier2016maxentirl}.
Because of these shortcomings, IRL researchers have opted recently for either completely hand-specifying the features or using deep IRL for extracting them automatically from the raw continuous state space with non-expert demonstrations~\citep{fu2018learning, finn2016gcl}.}

Rather than relying on demonstrations for everything, we propose to first learn complex non-linear features leveraging explicit human input about relevant aspects of the task (Sec. \ref{sec:feature_learning}). Based on these features, a reward can be inferred with minimal input (Sec. \ref{sec:reward_learning_method}). Our results show that adding structure in such a targeted way can enhance both the generalization of the learned reward and data-efficiency of the method.



\section{Problem Formulation}
\label{sec:formulation}

We consider a robot $R$ operating in the presence of a human $H$ from whom it is trying to learn to perform a task, ultimately seeking to enable autonomous execution.
In the most general setting, both $H$ and $R$ are able to affect the evolution of the \change{continuous} state $\sysstate \in \mathbb{R}^d$ \change{(i.e. robot joint poses or object poses)} over time through their respective \change{continuous} actions $\action_H$ and $\action_R$ via a dynamics function $f$: 
\begin{equation}
    \sysstate^{t+1} = f(\sysstate^t, \action_H^t, \action_R^t)\enspace,
\end{equation}
with $\action_H\in\mathcal{A}_H$ and $\action_R\in\mathcal{A}_R$, and $\mathcal{A}_H$ and $\mathcal{A}_R$ compact sets. 
Thus, when executing a task, the robot follows a trajectory $\traj = [\sysstate^0, \action_H^0, \action_R^0, \sysstate^1, \action_H^1, \action_R^1, \dots, \sysstate^T, \action_H^T, \action_R^T]$.

We assume that the human has some consistent internal preference ordering between different trajectories $\traj$, which affects the actions $\action_H$ that they choose. In principle, these human preferences could be captured by a reward function $\reward^*(\traj)$. 
Unfortunately, the robot does not have access to $\reward^*$, so to learn how to perform the task it must attempt to infer it.
Since $\reward^*$ may encode arbitrary preference orderings deeming the inference problem intractable, we assume that the robot reasons over a parameterized approximation $\reward_\weight$ induced by parameters $\weight\in\wspace$. The robot's goal is, thus, to estimate the human's preferred $\weight$ from their actions $\action_H$. 

Even with this parameterization, the space of possible reward functions is infinite-dimensional.
One way to represent it using a finite $\weight$ is through the means of a finite family of basis functions $\trajfeat_i$, also known as \textit{features} \citep{Ng2000inverse}: $\reward_\weight(\vec\trajfeat(\traj))$, where $\vec\trajfeat$ is the set of chosen features $\trajfeat_i$.
Consistent with classical utility theories \citep{von1945theory}, we may decompose trajectory features $\trajfeat_i$ into state features $\statefeat_i$ and approximate the trajectory's reward through a cumulative return over time:
\begin{equation}
    \change{\reward_\weight(\traj) = \reward_\weight(\vec\trajfeat(\traj)) = \sum_{(\sysstate, \action_H, \action_R) \in \traj}r_{\weight}\big(\vec\statefeat(s, \action_H, \action_R)\big)\enspace.}
    \label{eq:rewardstructure}
\end{equation}
This restriction to a finite set of features $\vec\statefeat$ is essentially a truncation of the infinite collection of basis functions spanning the full reward function space.
Thus, the features we choose to represent the reward dramatically impact the reward functions that can be learned altogether. Importantly, this observation holds regardless of the representation power that $r_\theta$ has (linear combination, neural network, etc).
Motivated by recovering a reward function $r_\theta$ that captures the person's preferences as best as possible, we are, thus, interested in the question of how to choose the feature representation $\vec\statefeat$.


We assume the robot has access to a (possibly empty) initial set of features $\vec\phi$.
In Sec. \ref{sec:feature_learning}, we propose a protocol via which the robot can learn a novel feature to add to its existing set by soliciting feature-specific human input.
We then describe classic offline \ac{IRL} and its adaptation to situations where the human is teaching the reward from scratch (Sec. \ref{sec:offline_FERL}); our framework enables them to teach one feature at a time before teaching the reward on top using task demonstrations.
Lastly, we present the online variant, where the robot executes the task according to a reward function defined on an incomplete feature set and the human intervenes to correct it (Sec. \ref{sec:online_FERL}); our method allows them to explicitly focus on teaching the missing feature(s) and adding them to the set before the reward is updated.

\section{Algorithmic Approach: Feature Learning}
\label{sec:feature_learning}

We first look at learning individual feature functions.
\change{In this paper, we focus on state features (ignoring actions from the feature representation), which we define as arbitrary complex mappings $\phi(\sysstate): \mathbb{R}^d \to \mathbb{R}^+$.
As such, in regions of the state space where the feature is highly expressed, this function has high positive values, whereas for states where the feature is not expressed, $\phi$ is closer to zero.} 


One natural idea for learning this mapping is treating it as a regression problem and asking the human for regression labels $(\sysstate, \phi(\sysstate))$ directly.
Unfortunately, to learn anything useful, the robot would need a very large set of labels from the person, which would be too \change{effortful for them to provide}.
Even worse, humans are notoriously unreliable at quantifying their preferences with any degree of precision~\citep{Braziunas2008elicitation}, so their labels might result in arbitrarily noisy regressions.
Hence, we need a type of human input that balances being informative and not placing too much burden on the human.

\subsection{Feature Traces}
\label{sec:feature_traces}

To teach a non-linear representation of $\phi$ with little data, we introduce \textit{feature traces} $\trace = \sysstate_{0:n}$, a novel type of human input defined as a sequence of $n$ states that are monotonically decreasing in feature value, i.e. $\phi(\sysstate_i) \geq \phi(\sysstate_j), \forall i < j$. 
This \change{approach} relaxes the need for accurate state labeling, while simultaneously providing a combinatorial amount of state comparisons (see Sec. \ref{sec:feature_training} for details) \change{from each trace $\trace$}.

When learning a feature, the robot can query the human for a set $\Xi$ of $N$ traces.
The person gives a trace $\trace$ by simply moving the system from any start state $\sysstate_0$ to an end state $\sysstate_n$, noisily ensuring monotonicity. 
Our method, thus, only requires an interface for communicating ordered \change{feature values} over states: kinesthetic teaching is useful for household or small industrial robots, while teleoperation and simulation interfaces may be better for larger robotic systems.

To illustrate how a human might offer feature traces in practice, let's turn to Fig. \ref{fig:front_fig} (Left). Here, the person is teaching the robot to keep the mug away from the laptop \change{(i.e. not above)}. The person starts a trace at $\sysstate_0$ by placing the end-effector \change{directly above}
the object center, then leads the robot away from the laptop to $\sysstate_n$. Our method works best when the person tries to be informative, i.e. covers diverse areas of the space: the traces illustrated move radially in all directions and start at different heights. 
While for some features, like distance from an object, it is easy to be informative, for others, like slowing down near objects, it might be more difficult. We explore how easy it is for users to be informative in our study in Sec. \ref{sec:feature_user_exps}, with encouraging findings, and discuss alleviating existing limitations in Sec. \ref{sec:discussion}. 

The power of feature traces lies in their inherent structure. 
Our algorithm, thus, makes certain assumptions to harvest this structure for learning. First, we assume that the feature values of states along the collected traces $\trace \in \Xi$ are monotonically decreasing.
\change{Secondly, we assume that by default the human starts all traces in states $\sysstate_0$ with the highest feature value across the domain, then leads the system to states $\sysstate_{n}$ with the lowest feature value.}
\change{In some situations, this assumption might unnecessarily limit the kinds of feature traces the human can provide.
For example, the person might want to start somewhere where the feature is only ``half'' expressed relative to the feature range of the domain.
Because of this, we optionally allow the human to provide \textit{relative values} $v_{0}, v_{n} \in [0,1]$\footnote{Since specifying decimal fractions is difficult, the person gives percentages between 0 and 100 instead.} to communicate that the traces start/end at values that are fractions of the feature range of the domain.}

\subsection{Learning a Feature Function}
\label{sec:feature_training}

To allow for arbitrarily complex non-linear features, we approximate a feature by a neural network \change{$\phi_{\psi}(\sysstate) : \mathbb{R}^d \to \mathbb{R}^+$}. 
\change{We incorporate the assumptions in the previous section by training $\phi_{\psi}$ as a discriminative function with respect to the state ordering in feature traces $\trace\in\Xi$, and also 
encouraging the starts $\sysstate_0$ and ends $\sysstate_n$ across all traces to have the same high and low values, respectively. For ease of exposition, we present our feature learning technique without the relative values $v_0$ and $v_n$ first, then later describe how to modify the algorithm to include them.} 

\subsubsection{Monotonicity Along Feature Traces.}

\change{
First, due to the monotonicity assumption along any feature trace $\trace_k = (\sysstate_0^k, \sysstate_1^k, \dots, \sysstate_n^k)$, when training $\phi_\psi$ we want to encourage feature values to decrease monotonically along every trace, i.e. $\phi_{\psi}(\sysstate_i^k) \geq \phi_{\psi}(\sysstate_j^k),  \forall j>i, k$. 
For this purpose, we convert the set of collected traces $\trace_k\in\Xi$ into a dataset of \textit{ordered} tuples $(\sysstate_i^k, \sysstate_j^k) \in \mathcal{T}_{ord}$, where every first element appears earlier in the trace than the second element (hence its feature value should be higher). 
This results in $\binom{(n+1)}{2}$ tuples per trace, which we can use for training $\phi_\psi$.}

\change{We train the discriminative function $\phi_\psi$ as a predictor for whether a state $\sysstate$ has a higher feature value than another state $\sysstate'$, which we represent as a softmax-normalized distribution:
}
%
\begin{equation}
    \change{P(\phi_\psi(\sysstate) > \phi_\psi(\sysstate')) = P(\sysstate \succ \sysstate') = \frac{e^{\phi_\psi(\sysstate)}}{e^{\phi_\psi(\sysstate)} + e^{\phi_\psi(\sysstate')}} \enspace,}
    \label{eq:softmax}
\end{equation}
%
%
%
\change{where we define the shorthand notation $\sysstate \succ \sysstate'$ for $\phi_\psi(\sysstate) > \phi_\psi(\sysstate')$.}
\change{We choose $\psi$ to minimize a negative log-likelihood loss $L_{ord}(\psi)$ operating on the ordered tuples dataset:}
\begin{align}
    L_{ord}(\psi) &= - \sum_{(\sysstate, \sysstate') \in \mathcal{T}_{ord}} \log(P(\sysstate \succ \sysstate')) \\
    &= -\sum_{(\sysstate, \sysstate') \in \mathcal{T}_{ord}} \log\frac{e^{\phi_{\psi}(\sysstate)}}{e^{\phi_{\psi}(\sysstate)} + e^{\phi_{\psi}(\sysstate')}} \enspace.
    \label{eq:loss_ord}
\end{align}




\change{Intuitively, this loss spaces out the feature values $\phi_\psi$ such that they decrease monotonically along every trace; however, this alone does not constrain the traces to have the same start and end values, respectively.}

\subsubsection{Start/End Feature Value Equivalence.}

\change{To encourage all traces to start and end in the same high and low feature values, we need an additional loss term encoding $\phi_\psi(\sysstate_{0}^i) = \phi_\psi(\sysstate_{0}^j)$ and $\phi_\psi(\sysstate_{n}^i) = \phi_\psi(\sysstate_{n}^j)$ for all $\trace_i, \trace_j \in \Xi$.
We thus convert the set of collected traces $\Xi$ into another dataset $\mathcal{T}_{equiv}$ of \textit{equivalence} tuples $(\sysstate_{0}^i, \sysstate_{0}^j), (\sysstate_{n}^i, \sysstate_{n}^j) \; \forall \; \trace_i, \trace_j \in \Xi, \;i \neq j,\; i > j$.
This results in $2 \binom{N}{2}$ tuples where the states of the tuple $(\sysstate, \sysstate')$ should have the same feature value, i.e. $\phi_\psi(\sysstate) = \phi_\psi(\sysstate')$. We denote this relationship as $\sysstate \sim \sysstate'$ to simplify notation.}

\change{When training $\phi_\psi$, the predictor should not be able to distinguish which state has a higher feature value, hence $P(\phi_\psi(\sysstate) > \phi_\psi(\sysstate')) = 0.5$.
As such, we introduce a second loss function $L_{equiv}(\psi)$ that minimizes the negative log-likelihood of both $\sysstate$ having a higher feature value than $\sysstate'$ and $\sysstate'$ having a higher feature value than $\sysstate$:}
\begin{align}
    L_{equiv}(\psi) &= - \sum_{(\sysstate, \sysstate') \in \mathcal{T}_{equiv}} \log(P(\sysstate \succ \sysstate')) + \log(P(\sysstate' \succ \sysstate)) \\
    &= -\sum_{(\sysstate, \sysstate') \in \mathcal{T}_{equiv}} \log \frac{e^{\phi_{\psi}(\sysstate)+\phi_{\psi}(\sysstate')}}{(e^{\phi_{\psi}(\sysstate)} + e^{\phi_{\psi}(\sysstate')})^2} \enspace.
    \label{eq:loss_equiv}
\end{align}

\change{This loss ensures the state space around feature trace starts and ends have similar feature values, respectively.}
\footnote{\change{One could choose other losses to ensure equivalence of start and end values such as a p-norm $||\phi_\psi(\sysstate) - \phi_\psi(\sysstate')||_p$. 
We experimented with $p=2$ but it produced inferior results.}}

\change{We now have a total dataset $\mathcal{T} = \mathcal{T}_{ord} \cup \mathcal{T}_{equiv}$ of $|\mathcal{T}| = \sum_{i=1}^{N}\binom{(n^i+1)}{2} + 2 \binom{N}{2}$ tuples, which is already significantly large for a small set of feature traces.}
%
\change{We can use it to optimize a loss $L(\psi)$ that combines the ordered and equivalence losses:}
%
\begin{equation}
    \change{L(\psi) = L_{ord}(\psi) + \lambda L_{equiv}(\psi)\enspace,}
    \label{eq:loss_total}
\end{equation}
\change{where $\lambda$ is a hyperparameter trading off the two loss functions.}
%
%

Given the loss function in Eq. \eqref{eq:loss_total}, we can use any automatic differentiation package to compute its gradients and update $\psi$ via gradient descent.
\change{Note that $L_{equiv}$ is akin to a binary cross-entropy loss with a target of 0.5, whereas $L_{ord}$ is similar to a binary cross-entropy loss with a target of 1.}
\change{This form of loss function has been shown to be effective for preference learning~\citep{christiano2017preferences,Ibarz2018reward}.}
The key differences here are that our loss is over feature functions not rewards, and that preferences are state orderings provided via feature traces not trajectory comparisons. \change{Additionally, in practice we normalize the feature functions to make their subsequent reward weights reflect importance relative to one another.} We present the full feature learning algorithm using feature traces in Alg. \ref{alg:feature_learning}.

\begin{algorithm}
\DontPrintSemicolon
\textbf{Input:} $N$ number of queries, $K$ iterations. \\
\For{$i\leftarrow 1$ \KwTo $N$}{
    Query feature trace $\trace$ as in Sec. \ref{sec:feature_traces}.\\
    $\Xi \gets \Xi \cup \trace$.
    }
\change{Convert $\Xi$ to datasets $\mathcal{T}_{ord}$ and $\mathcal{T}_{equiv}$ as in Sec. \ref{sec:feature_training}}.\\
Initialize $\statefeat_\psi$ randomly.\\
\For{iteration $k\leftarrow 1$ \KwTo $K$}{
    \change{Sample tuples batch $\hat{\mathcal{T}}_{ord}\in\mathcal{T}_{ord}$.}\\
    \change{Sample tuples batch $\hat{\mathcal{T}}_{equiv}\in\mathcal{T}_{equiv}$.}\\
    \change{Estimate $L(\psi)$ using $\hat{\mathcal{T}}_{ord}$, $\hat{\mathcal{T}}_{equiv}$, and Eq. \eqref{eq:loss_total}.}\\
    Update parameter $\psi$ via gradient descent on $L(\psi)$.\\
}
\Return{\change{normalized} $\statefeat_\psi$}
\caption{Feature Learning via Feature Traces}
\label{alg:feature_learning}
\end{algorithm}

\subsubsection{Incorporating Relative Values.}

\change{So far, we have assumed that all feature traces have starts and ends of the same high and low feature value, respectively. The optional relative values $v_{0}, v_{n}$ can relax this assumption to enable the human to provide richer traces and teach more complex feature functions, e.g. where no monotonic path from the highest to lowest feature value exists. 
By default, $v_0 = 1$ communicating that the trace starts at the highest feature value of the domain, and $v_n = 0$ signifying that the trace ends at the lowest feature value.
By allowing $v_0$ and $v_n$ to be something different from their defaults, the person can provide traces that start at higher feature values or end at lower ones. We describe how to include these relative values in the feature training procedure in App. \ref{app:relative_values}.}



\section{Algorithmic Approach: Reward Learning}
\label{sec:reward_learning_method}

Now that we have a method for learning relevant features, we discuss how the robot can include this capability in reward learning frameworks.
For exposition, we chose two reward learning frameworks -- learning from demonstrations (offline) and from corrections (online) -- but we stress that features learned with our method are applicable to any other reward learning method that admits features (e.g. comparisons, scalar feedback, state of the world, etc.). 

\subsection{Offline \ac{FERL}}
\label{sec:offline_FERL}

We first consider the scenario where the human is attempting to teach the robot a reward function from scratch, i.e. the robot starts off with an empty feature set $\vec\statefeat$. For instance, imagine a system designer trying to engineer the robot's reward before deployment, or an end user resetting it and custom designing the reward for their home.
We can think of this as an offline reward learning setting, where the person provides inputs to the robot before it starts executing the task.
Here, we focus on learning from demonstrations, although our framework can be adapted to any other offline reward learning strategy.

In standard learning from demonstrations, deep \ac{IRL} uses a set of demonstrations to train a reward function directly from the raw state, in an end-to-end fashion.
Under our divide-and-conquer framework, we redistribute the human input the robot asks for: first ask for feature traces $\trace$ focusing explicitly on learning $F$ features one by one via Alg. \ref{alg:feature_learning}, and only then collect a few demonstrations $\traj\in\mathcal{D}^*$ to learn the reward on top of them.
Alg. \ref{alg:S-FeRL} summarizes the full procedure.

\begin{algorithm}
\DontPrintSemicolon
\textbf{Input:} Demonstration set $\mathcal{D}^*$, $F$ number of features, $K$ iterations, $\alpha$ learning rate. \\
Initialize empty feature set $\vec\phi=[]$.\\
\For{$f\leftarrow 1$ \KwTo $F$}{
    Learn feature $\phi_f$ using Alg. \ref{alg:feature_learning}.\\
        $\vec{\phi} \gets (\vec{\phi}, \phi_f)$.
    }
Initialize $\weight$ randomly.\\
\For{iteration $k\leftarrow 1$ \KwTo $K$}{
    Generate samples $\mathcal{D}^\weight$ using current reward $\reward_\weight$.\\
    Estimate gradient $\nabla\mathcal{L}$ using $\mathcal{D}^*$, $\mathcal{D}^\weight$ in Eq. \eqref{eq:loglikelihood_gradient}.\\
    Update parameter $\weight$ using gradient $\nabla\mathcal{L}$ in Eq. \eqref{eq:sferl_update}.\\
}
\Return{optimized reward parameters $\weight$}
\caption{Offline \ac{FERL}}
\label{alg:S-FeRL}
\end{algorithm}

\subsubsection{Creating the Feature Set.}

Since the robot starts off with an empty feature set $\vec\statefeat$, the person has to teach it every relevant feature one at a time. To do so, they follow the procedure in Alg. \ref{alg:feature_learning}, that is they collect a set of feature traces $\trace\in\Xi$ for the current feature, then use them to train $\statefeat_\psi$. The person can add this new feature to the robot's existing set:
\begin{equation}
    \vec\phi \gets (\vec\phi, \phi_\psi)\enspace,
    \label{eq:feature_add}
\end{equation}
and repeat the procedure for as many features $F$ as they want.

\change{After being equipped with a new set of features taught by the human, the robot can undergo standard learning from demonstration procedures to recover the person's preferences. 
We now review Maximum Entropy IRL~\citep{ziebart2008maximum} for completion of the offline reward learning exposition.}

\subsubsection{Offline Reward Learning.}

To teach the robot the desired reward function $\reward_\weight$, the person collects a set of demonstrations $\traj\in\mathcal{D}^*$ for how to perform the task by directly controlling the state $\sysstate$ through their input $\action_H$. During a demonstration, the robot is put in gravity compensation mode or teleoperated, to allow the person full control over the desired trajectory.
The robot interprets the set of demonstrations $\mathcal{D}^*$ as evidence about the human's preferred $\weight$ parameter, and uses them to estimate it and, thus, to learn the reward function.

In order to reason about the human's preferences, the robot needs to be equipped with a model $P(\traj \mid \weight)$ for how those preferences affect their choice of demonstrations. For example, if the human were assumed to act optimally, the model would place all the probability on the set of trajectories that perfectly optimize the reward $\reward_\weight$. However, since humans are not perfect, we relax this assumption and model them as being \emph{noisily-optimal}, choosing trajectories that are approximately aligned with their preferences. We follow the Boltzmann noisily-rational decision model:
\begin{equation}
    P(\traj \mid \weight, \beta) = \frac{e^{\beta R_{\weight}(\traj)}}{\int_{\bar{\traj}}e^{\beta R_{\weight}(\bar{\traj})} d\bar{\traj}} \enspace,
    \label{eq:observation_model}
\end{equation}
where the human picks trajectories proportional to their exponentiated reward \citep{baker2007goal,von1945theory}. Here, $\beta \in [0, \infty)$ controls how much the robot expects to observe human input consistent with its reward model. 
For now, we use the Maximum Entropy \ac{IRL} \citep{ziebart2008maximum} version of this observation model where $\beta$ is fixed to 1, so for notation simplicity we refer to this model as $P(\traj \mid \weight)$. 
Later in Sec. \ref{sec:online_FERL}, we will allow $\beta$ to vary and make use of it in the online version of our framework.

In maximum entropy \ac{IRL}, to recover the $\weight$ parameter we maximize the log-likelihood $\mathcal{L(\weight)}$ of the observed data under the above model~\citep{jaynes1957infotheory}. To see how, let's start by writing down the log-likelihood formula:
\begin{equation}
\begin{aligned}
    \mathcal{L}(\weight) =& \log\prod_{\traj\in\mathcal{D}^*} P(\traj \mid \weight) = \sum_{\traj\in\mathcal{D}^*} \log \frac{e^{ \reward_{\weight}(\traj)}}{\int_{\bar{\traj}}e^{\reward_{\weight}(\bar{\traj})} d\bar{\traj}} \\
    =&  \sum_{\traj\in\mathcal{D}^*} \reward_\weight(\traj) - |\mathcal{D}^*|\log \int_{\bar{\traj}}e^{\reward_{\weight}(\bar{\traj})} d\bar{\traj} \enspace.
\end{aligned}
\end{equation}

Computing the integral over trajectories is intractable in real-world problems, so sample-based approaches to maximum entropy \ac{IRL} estimate it with samples $\traj\in\mathcal{D}'$ drawn from a background distribution $q(\traj)$:
\begin{equation}
\begin{aligned}
    \mathcal{L}(\weight) \approx \sum_{\traj\in\mathcal{D}^*} \reward_\weight(\traj) - |\mathcal{D}^*|\log \frac{1}{|\mathcal{D}'|} \sum_{\bar{\traj}\in\mathcal{D}'}\frac{e^{\reward_{\weight}(\bar{\traj})}}{q(\bar\traj)} \enspace.
\end{aligned}
\end{equation}
The distribution $q(\traj)$ is chosen often times to be uniform; instead, we follow~\citet{finn2016gcl} and generate samples in those regions of the trajectory space that are good according to the current estimate of the reward function, i.e. $q(\traj)\propto e^{\reward_{\weight}(\traj)}$. We denote $\traj\in\mathcal{D}^\weight$ such a set sampled under $\weight$.

We may now find $\weight$ by maximizing the log-likelihood $\mathcal{L}(\weight)$ using gradient-based optimization on the above objective. The gradient then takes the following form:
%
%
\begin{equation}
    \nabla \mathcal{L} = \frac{1}{|\mathcal{D}^*|}\sum_{\traj\in\mathcal{D}^*}\nabla \reward_\weight(\traj) - \frac{1}{|\mathcal{D}^\weight|}\sum_{\bar\traj\in\mathcal{D}^\weight}\nabla \reward_\weight(\bar\traj)\enspace.
    \label{eq:general_loglikelihood_gradient}
\end{equation}

At this point, a standard deep \ac{IRL} baseline could use any automatic differentiation package to compute the gradient and update the reward parameters directly from the raw trajectory state. Instead, consistent with prior work on reward learning with feature sets, we represent the reward as a linear combination of the learned features $\vec\statefeat$:
\begin{equation}
    \reward_\weight(\traj) = \weight^T\vec\trajfeat(\traj) = \sum_{(\sysstate, \action_H, \action_R) \in \traj}\weight^T\vec\statefeat(s)\enspace.
    \label{eq:linear_reward}
\end{equation}
Note that the linear reward assumption is not necessary for our algorithm to work. While in theory the reward could be modeled as non-linear, our divide-and-conquer approach is motivated by keeping the reward parameter space small while still effectively capturing the person's preferences.

For the linear case, the gradient becomes the difference between the observed demonstration feature values and the expected feature values dictated by the sampled trajectories:
%
%
\begin{equation}
    \nabla \mathcal{L} = \frac{1}{|\mathcal{D}^*|}\sum_{\traj\in\mathcal{D}^*} \vec\trajfeat(\traj) - \frac{1}{|\mathcal{D}^\weight|}\sum_{\bar\traj\in\mathcal{D}^\weight} \vec\trajfeat(\bar\traj)\enspace.
    \label{eq:loglikelihood_gradient}
\end{equation}

Lastly, we compute an estimate $\hat\weight$ by iteratively computing the gradient $\nabla\mathcal{L}$ and updating the parameters until convergence:
\begin{equation}
    \hat\weight' = \hat\weight - \alpha \left( \frac{1}{|\mathcal{D}^*|}\sum_{\traj\in\mathcal{D}^*} \vec\trajfeat(\traj) - \frac{1}{|\mathcal{D}^\weight|}\sum_{\bar\traj\in\mathcal{D}^\weight} \vec\trajfeat(\bar\traj)\right) \enspace,
    \label{eq:sferl_update}
\end{equation}
where $\alpha$ is the learning rate chosen appropriately.
The final reward learning procedure, thus, consists of $K$ iterations of generating samples $\mathcal{D}^\weight$ under the current reward, using them to estimate the gradient in Eq. \ref{eq:loglikelihood_gradient}, and updating the parameter $\weight$ via gradient descent with Eq. \ref{eq:sferl_update}.


%

\subsection{Online \ac{FERL}}
\label{sec:online_FERL}

In Sec. \ref{sec:offline_FERL}, we saw that our method allows the person to specify a reward by sequentially teaching features and adding them to the robot's feature set before using demonstrations to combine them.
However, in many situations the system designer or even the user teaching the features might not consider all aspects relevant for the task \textit{a priori}.
As such, we now consider an online reward learning version of our previous scenario, where the person provides inputs to the robot during the task execution and its feature space may or may not be able to correctly interpret them.

We assume the robot has access to an initial feature set $\vec\phi$, and is tracking a trajectory $\traj$ optimizing its current estimate of the reward function $\reward_\weight$ in Eq. \eqref{eq:linear_reward}.
If the robot is not executing the task according to the person's preferences, the human can intervene with input $a_H$. For instance, $a_H$ might be an external torque that the person applies to change the robot's current configuration. Or, they might stop the robot and kinesthetically demonstrate the task, resulting in a trajectory. Building on prior work, we assume the robot can evaluate whether its existing feature space can explain the human input (Sec.~\ref{sec:confidence_estimation}). If it can, the robot directly updates its reward function parameters $\weight$, also in line with prior work~\citet{bajcsy2017phri,ratliff2006MMP} (Sec. \ref{sec:online_reward_update}).
If it can not, the human can teach the robot a new feature\footnote{Because feature learning was triggered by an intervention, it is fair to assume that the human knows what aspect of the task they were trying to correct.} $\statefeat_\psi$ just like in Sec. \ref{sec:offline_FERL} and augment its feature set $\vec\phi \gets (\vec\phi, \phi_\psi)$. The robot can then go back to the original human input $a_H$ that previously could not be explained by the old features and use it to update its estimate of the reward parameters $\weight$. Algorithm \ref{alg:E-FeRL} summarizes the full procedure.

\begin{algorithm}
\DontPrintSemicolon
\textbf{Input:} Features $\vec\phi=[\phi_1, \dots, \phi_f]$, initial parameters $\weight$, confidence threshold $\epsilon$. \\
Plan initial trajectory $\traj$ by optimizing $\reward_\weight$.\\
\While{executing $\traj$}{
    \If{$\action_H$}{
    Estimate confidence $\hat\beta$ from $\action_H$ using Eq. \eqref{eq:beta}.\\
         \If{$\hat\beta < \epsilon$}{
        Learn feature $\statefeat_{new}$ using Alg. \ref{alg:feature_learning}.\\
        $\vec\statefeat \gets (\vec\statefeat, \statefeat_{new}), \weight \gets (\weight, 0.0)$.
         }
        Get induced trajectory $\traj_H$ from Eq. \eqref{eq:deformation}.\\
        Update parameter $\weight$ using $\traj_H$ in Eq. \eqref{eq:eferl_update}.\\
        Replan trajectory $\traj$ by optimizing new $\reward_\weight$.
    }
}
\caption{Online \ac{FERL}}
\label{alg:E-FeRL}
\end{algorithm}

\begin{figure*}
\centering
\begin{subfigure}{.16\textwidth}
  \centering
  \includegraphics[width=\textwidth,left]{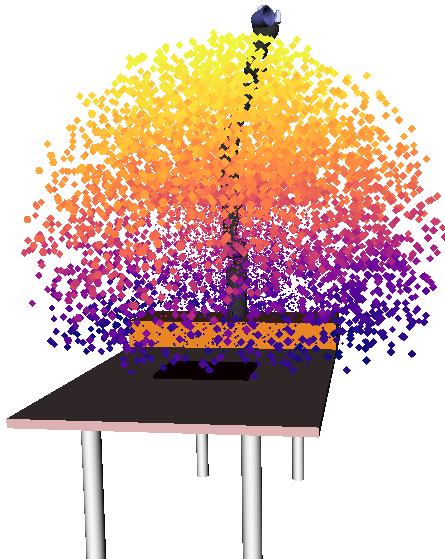}
\end{subfigure}%
\begin{subfigure}{.31\textwidth}
  \centering
  \includegraphics[width=\textwidth,left]{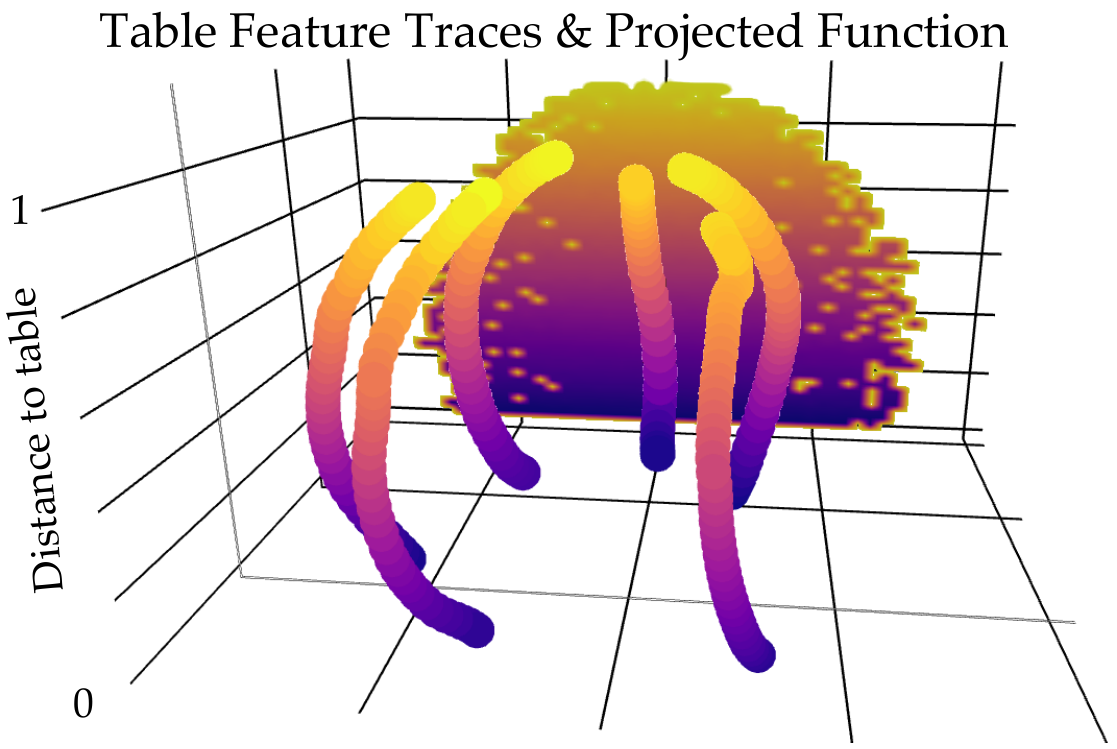}
\end{subfigure}
\begin{subfigure}{.23\textwidth}
  \centering
  \includegraphics[width=\textwidth,left]{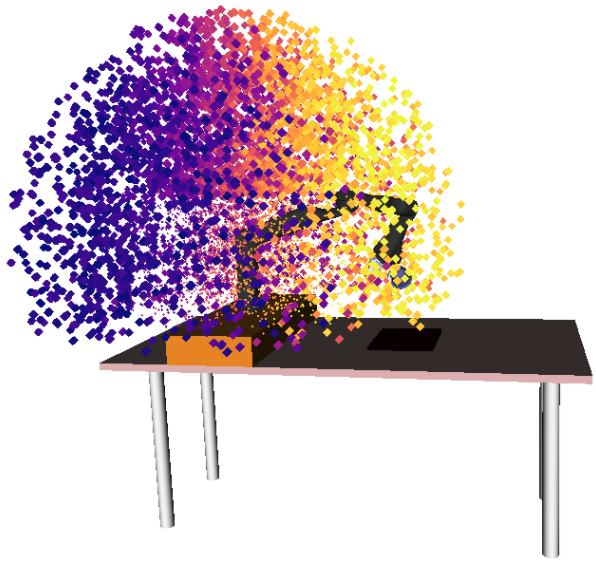}
\end{subfigure}
\begin{subfigure}{.27\textwidth}
  \centering
  \includegraphics[width=\textwidth,left]{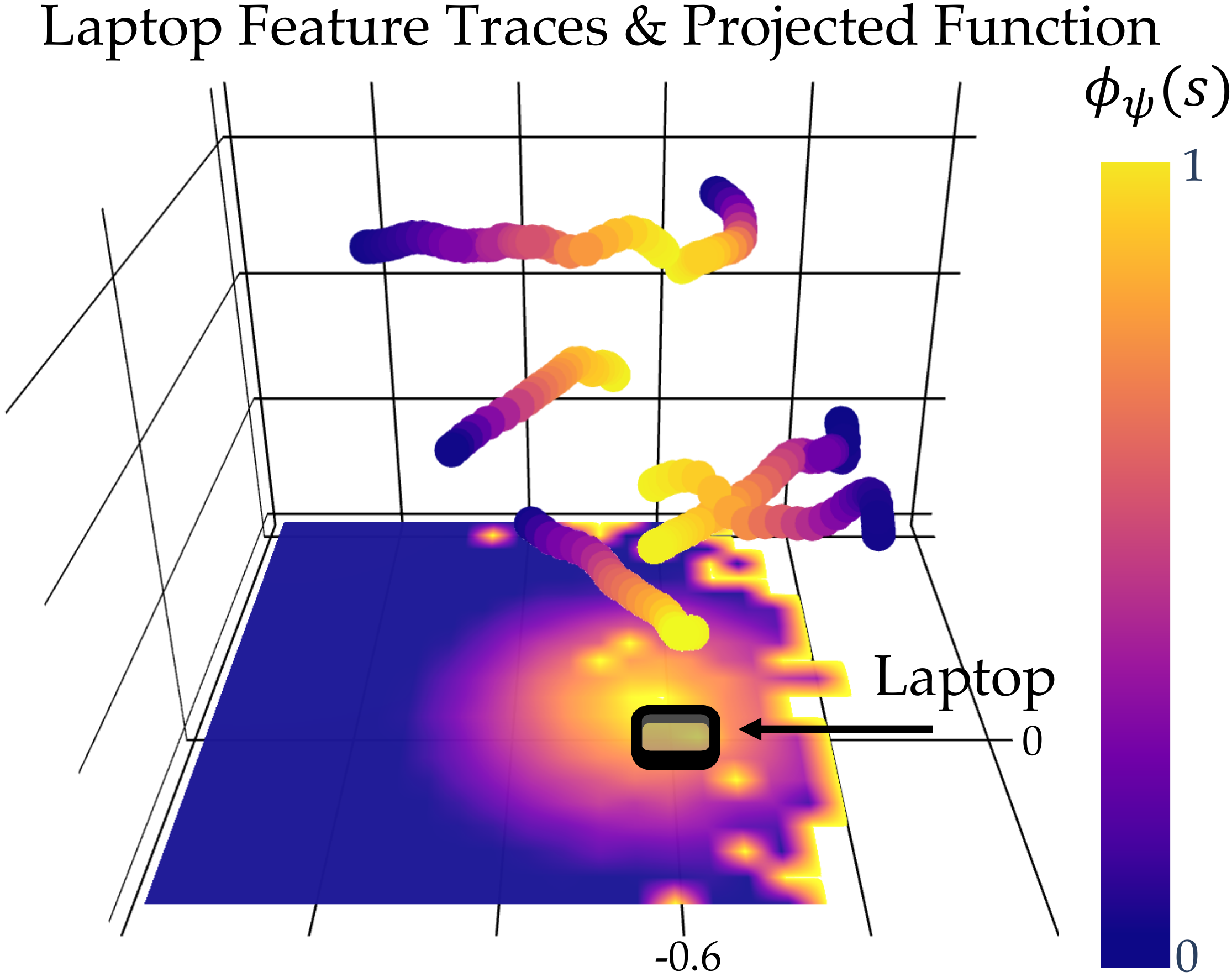}
\end{subfigure}
\caption{Visualization of the experimental setup, learned feature values $\phi_{\psi}(s)$, and training feature traces $\trace$ for \textit{table} (up) and \textit{laptop} (down). We display the feature values $\phi_{\psi}(\sysstate)$ for states $\sysstate$ sampled from the reachable set of the 7-DoF arm, as well as their projections onto the $yz$ and $xy$ planes.} 
\label{fig:Laptop_Feature_Exp}
\end{figure*}

\subsubsection{Online Reward Update.}
\label{sec:online_reward_update}

Whether it needs to learn a new feature $\statefeat_\psi$ or not, the robot has to then use the human input $a_H$ to update its estimate of the reward parameters $\weight$. Here, any prior work on online reward learning from user input is applicable, but we highlight one example to complete the exposition.

For instance, take the setting where the human's input $\action_H$ was an external torque, applied as the robot was tracking a trajectory $\traj$ that was optimal under its current reward $\reward_\weight$. Prior work \citet{bajcsy2017phri} has modeled this as inducing a deformed trajectory $\tau_H$, by propagating the change in configuration to the rest of the trajectory: 
\begin{equation}
    \traj_H = \traj + \mu A^{-1}\tilde{\textbf{\action}}_H\enspace,
    \label{eq:deformation}
\end{equation}
where $\mu>0$ scales the magnitude of the deformation, $A$ defines a norm on the Hilbert space of trajectories\footnote{We used a norm $A$ based on acceleration, consistent with \citet{bajcsy2017phri}, but other norm choices are possible as well.} and dictates the deformation shape~\citep{deformation}, and $\tilde{\textbf{\action}}_H$ is $\action_H$ at the interaction time and 0 otherwise. 

If we think of $\traj_H$ as the human observation and of $\traj$ as the expected behavior according to the current reward function \citep{bajcsy2017phri}, we arrive at a natural alternation of the update rule in Eq. \eqref{eq:sferl_update}:
\begin{equation}
    \hat\weight' = \hat\weight - \alpha\Big(\vec\trajfeat(\traj_H) - \vec\trajfeat(\traj)\Big) \enspace.
    \label{eq:eferl_update}
\end{equation}
Intuitively, the robot updates its estimate $\hat\weight$ in the direction of the feature change induced by the human's correction $\action_H$ from $\tau$ to $\tau_H$.

If instead, the human intervened with a full demonstration, work on online learning from demonstrations (Sec. 3.2 in \citet{ratliff2006MMP}) has derived the same update with $\tau_H$ now being the human demonstration. In our implementation, we use corrections and follow \citet{bajcsy2018onefeature}, which shows that people more easily correct one feature at a time, and only update the $\weight$ index corresponding to the feature that changes the most (after feature learning this is the newly learned feature). After the update, the robot replans its trajectory using the new reward.

\subsubsection{Confidence Estimation.} 
\label{sec:confidence_estimation}

The robot can learn a new feature from the person because we assumed it has the capacity to detect that a feature is missing in the first place. 
We alluded earlier in Sec. \ref{sec:offline_FERL} how this ability might be enabled by manipulating the $\beta$ parameter in the observation model in Eq. \eqref{eq:observation_model}. We now expand on this remark.

In the presented Boltzmann model, $\beta$ controls how much the robot expects to observe human input consistent with its reward structure, and, thus, its feature space. A high $\beta$ suggests that the input is consistent with the robot's feature space, whereas a low $\beta$ may signal that no reward function captured by the feature space can explain the input.
As such, inspired by work in \citet{fridovich-keil2019confidence,fisac2018probabilistically,bobu2020quantifying}, instead of keeping $\beta$ fixed like in the maximum entropy \ac{IRL} observation model, we reinterpret it as a \emph{confidence} in the robot's features' ability to explain human data. 

When the human input $\action_H$ is a correction, following \citet{bobu2020quantifying}, the robot estimates $\hat\beta$ by considering how efficient the human input $\action_H$ is in achieving the induced trajectory features $\vec\trajfeat(\traj_H)$. Accordingly, $\hat\beta$ is inversely proportional to the difference between the actual human input and the input that would have produced $\vec\trajfeat(\traj_H)$ optimally:
\begin{equation}
    \hat{\beta} \propto \frac{1}{\|\action_H\|^2 - \|\action_H^*\|^2} \enspace ,
    \label{eq:beta}
\end{equation}
where we obtain $\action_H^*$ by solving the optimization problem presented in \citet{bobu2020quantifying} Eq. (21).

Intuitively, if the person's input is close to the optimal $\action_H^*$, then it achieves the induced features $\vec\trajfeat(\traj_H)$ efficiently, resulting in high confidence $\hat{\beta}$. If, however, there is a far more efficient alternative input -- the difference between $\action_H$ and $\action^*_H$ is large --, $\hat\beta$ will be small: the person probably intended to give input about a feature the robot does not know about.  

Alternatively, if the human input $\action_H$ is a demonstration, like in the classical \ac{IRL} presented in Sec. \ref{sec:offline_FERL}, also following \citet{bobu2020quantifying}, we may estimate $\hat\beta$ via a Bayesian belief update: $b'(\weight, \beta) \propto P(\traj \mid \weight, \beta)b(\weight, \beta)$. Once again, in our implementation we used corrections, but the work in \citet{bobu2020quantifying} shows confidence estimation can easily be adapted to learning from demonstrations if desired.

To detect a missing feature, the robot simply needs a confidence threshold $\epsilon$.
If $\hat\beta$ is above the threshold, the robot is confident in its feature space, so it updates the reward as usual; if $\hat\beta < \epsilon$, its features are insufficient and the robot asks the person to be taught a new one.

\section{Experiments: Learning Features}
\label{sec:feature_learning_exp}

Before testing \ac{FERL} in the two reward learning settings of interest, we first analyze our method for learning features in experiments with a robotic manipulator. In Sec. \ref{sec:feature_expert_exps}, we inspect how well \ac{FERL} can learn six different features of varying complexity by using real robot data collected from an expert -- a person familiar with how the algorithm works. We then conduct an online user study in simulation in Sec. \ref{sec:feature_user_exps} to test whether non-experts -- people not familiar with \ac{FERL} but taught to use it -- can teach the robot good features.

\subsection{Expert Users}
\label{sec:feature_expert_exps}

\begin{figure}
\includegraphics[width=0.48\textwidth]{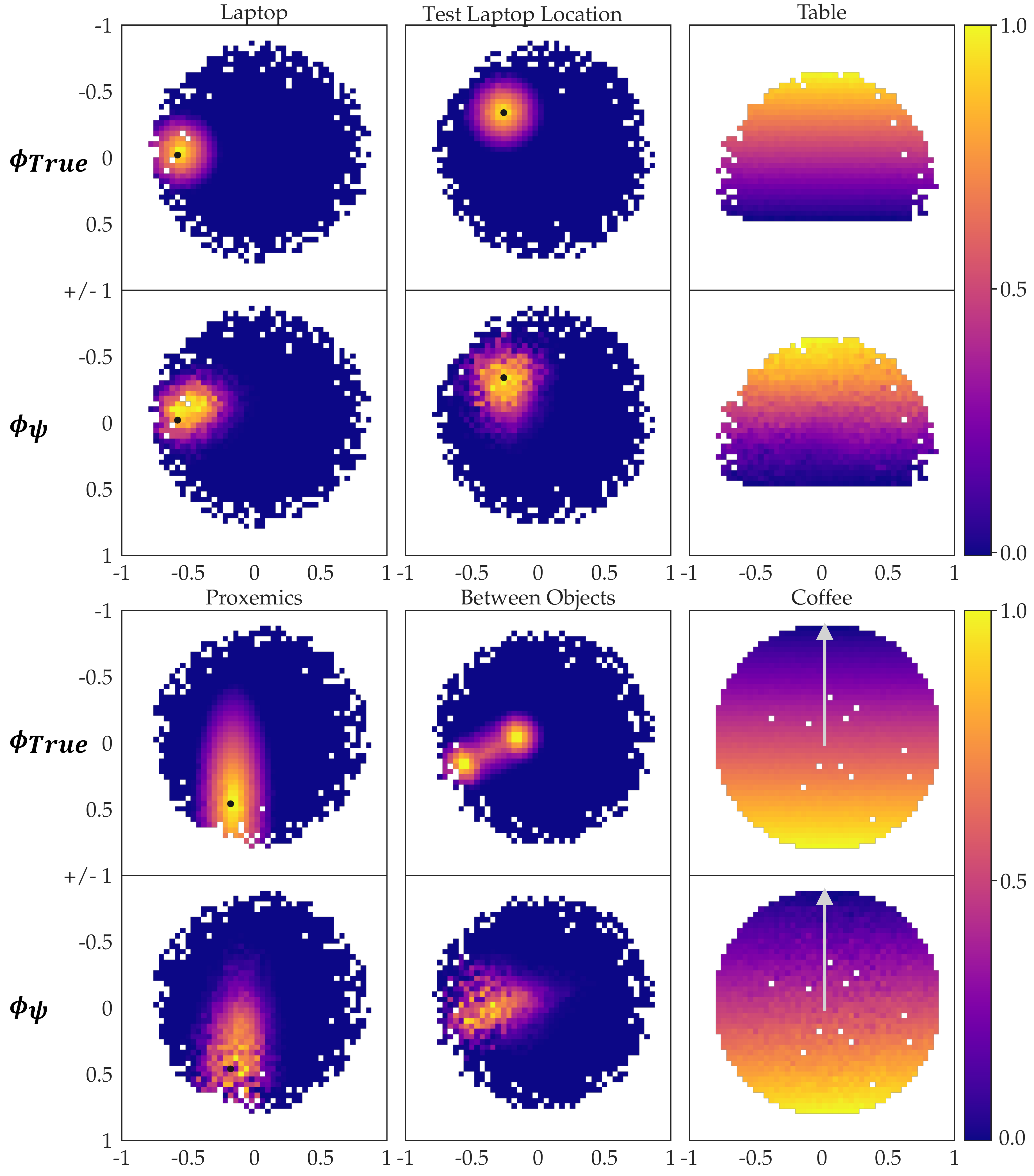}
\centering
\caption{The plots display the ground truth $\phi_{\text{True}}$ (top rows) and learned feature values $\phi_{\psi}$ (bottom rows) over $\mathcal{S}_{\text{Test}}$, averaged and projected onto a representative 2D subspace: the $xy$-plane, the $yz$-plane (table), and the $xz$ orientation plane for \textit{coffee} (the arrow represents the cup upright).}
\label{fig:FERL_Qual}
\end{figure}

We have argued that feature traces are useful in teaching the robot features explicitly. In our first set of experiments, we look at how good the learned features are, and how their quality varies with the amount of feature traces provided.

\subsubsection{Experimental Design.}
We conduct our experiments on a 7-DoF JACO robotic arm. We investigate six features in the context of personal robotics:
%
\begin{enumerate}
\item \textit{table}: distance of the \ac{EE} to the table \change{(T), as a $z$-coordinate difference: $EE^z - T^z$ (superscript denotes pose coordinate selection)};
\item \textit{coffee}: \change{coffee cup upright orientation, defined by how far the \ac{EE} is from pointing up: $1 - EE^R \cdot [0, 0, 1]$ (superscript denotes pose rotation matrix)};
\item \textit{laptop}: 0.3 meter $xy$-plane distance of the \ac{EE} to a laptop \change{(L)}, to avoid passing over the laptop:
\change{$\max\{0.3 - \|EE^{xy} - L^{xy}\|_2, 0\}$};
%
%
\item \textit{test laptop location}: same as \textit{laptop} but the test position differs from the training ones;
\item \textit{proxemics}: \change{non-symmetric 0.3 meter $xy$-plane distance between the \ac{EE} and the human (H), to keep the \ac{EE} away from them, three times as much when moving in front of them than on their side : $\max\{0.3 - \sqrt{\big(\frac{EE^y-H^y}{3}\big)^2 + (EE^x-H^x)^2}, 0\}$};
%
%
\item \textit{between objects}: 0.2 meter $xy$-plane distance of the \ac{EE} to two objects, \change{$O_1$ and $O_2$ -- the feature penalizes being above either object, and, to a lesser extent, passing in between the objects as defined by a distance to the imaginary line $O_1O_2$: $\max\{0.2 - \min\{0.8*\|O_1O_2^{xy} - EE^{xy}\|_2, \|O_1^{xy} - EE^{xy}\|_2, \|O_2^{xy} - EE^{xy}\|_2\}, 0\}$.}
\end{enumerate}

\change{Most features can be taught with the default relative values $v_n=0$ and $v_0=1$, but \textit{between objects}} requires some traces with explicit values $v_0, v_n$. We approximate all features $\phi_{\psi}$ by neural networks (2 layers, 64 units each), and train them on a set of traces $\Xi$ using stochastic gradient descent (see App. \ref{app:FERL_implementation} for training details details). 

For each feature, we collected a set $\mathcal{F}$ of 20 feature traces (40 for the complex \textit{test laptop location} and \textit{between objects}) from which we sample subsets $\Xi \in \mathcal{F}$ for training. 
We decide for each feature what an informative and intuitive set of traces would be, i.e. how to choose the starting states to cover enough of the space (details in App. \ref{app:FERL_traces}). 
As described in Sec. \ref{sec:feature_training}, the human teacher starts at a state where the feature is highly expressed, e.g. for \textit{laptop} that is the \ac{EE} positioned above the laptop. They then move the \ac{EE} away until the distance is equal to the desired radius. They do this for a few different directions and heights to give a diverse dataset.

Our raw state space consists of the 27D $xyz$ positions of all robot joints and objects in the scene, as well as the rotation matrix of the \ac{EE}. We assume known object positions but they could be obtained from a vision system. 
It was surprisingly difficult to train on both positions and orientations due to spurious correlations in the raw state space, hence we show results for training only on positions or only on orientations. This speaks to the need for methods that can handle correlated input spaces, which we expand on in App. \ref{app:rawstate}.

\paragraph{Manipulated Variables.}

We are interested in seeing trends in how the quality of the learned features changes with more or less data available. Hence, we manipulate the number of traces $N$ the learner gets access to. 

\paragraph{Dependent Measures.}

After training a feature $\phi_{\psi}$, we measure error compared to the ground truth feature $\phi_{\text{True}}$ that the expert tries to teach, on a test set of states $\mathcal{S}_{\text{Test}}$.
To form $\mathcal{S}_{\text{Test}}$, we uniformly sample 10,000 states from the robot's reachable set. Importantly, many of these test points are far from the training traces, probing the generalization of the learned features $\phi_{\psi}$. We measure error via the \ac{MSE}, $\text{MSE} = \frac{1}{|\mathcal{S}_{\text{Test}}|} \sum_{\mathcal{S}_{\text{Test}}}||\phi_{\psi}(\sysstate) - \phi_{\text{True}}(\sysstate)||^2$. To ground the \ac{MSE} values, we normalize them with the mean \ac{MSE} of a randomly initialized untrained feature function, $\text{MSE}_{\text{norm}} = \frac{\text{MSE}}{\text{MSE}_{\text{random}}}$, hence a value of 1.0 is random performance. For each $N$, we run 10 experiments sampling different feature trace sets $\Xi$ from $\mathcal{F}$, and calculate $\text{MSE}_{\text{norm}}$.

\paragraph{Hypotheses.} \hfill

\noindent\textbf{H1:} With enough data, \ac{FERL} learns good features.

\noindent\textbf{H2:} \ac{FERL} learns increasingly better features with more data.

\noindent\textbf{H3:} \ac{FERL} becomes less input-sensitive with more data.

\begin{figure}
\includegraphics[width=0.48\textwidth]{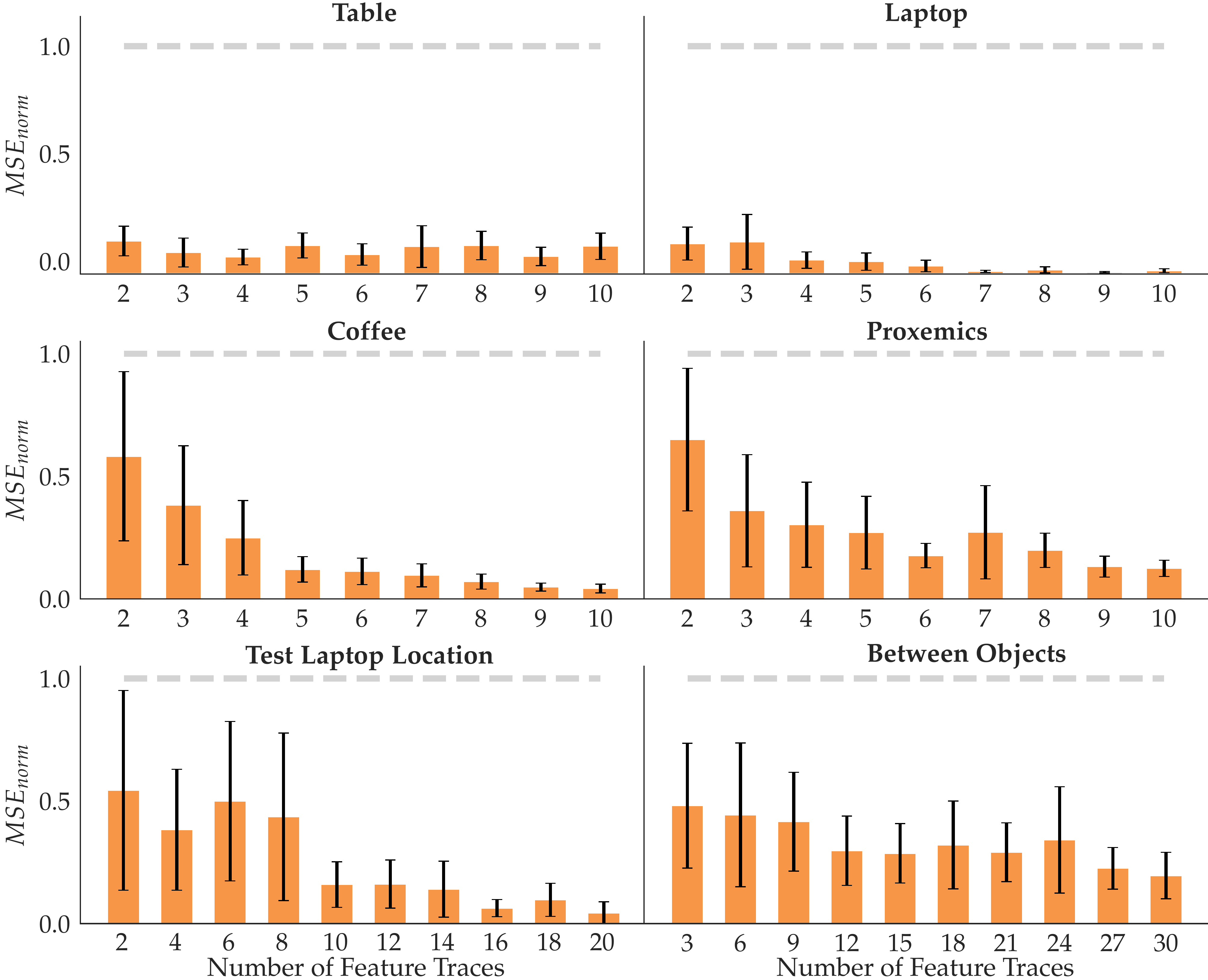}
\centering
\caption{For each feature, we show the $\text{MSE}_{\text{norm}}$ mean and standard error across 10 random seeds with an increasing number of traces (orange) compared to random (gray).}
\label{fig:FERL_MSE}
\end{figure}

\subsubsection{Qualitative Results.}
We first inspect the results qualitatively, for $N\!=\!10$. In Fig. \ref{fig:Laptop_Feature_Exp} we show the learned \textit{table} and \textit{laptop} features $\phi_{\psi}$ by visualizing the position of the \ac{EE} for all 10,000 points in our test set. The color of the points encodes the learned feature values $\phi_{\psi}(\sysstate)$ from low (blue) to high (yellow): \textit{table} is highest when the \ac{EE} is farthest, while \textit{laptop} peaks when the \ac{EE} is above the laptop.
In Fig. \ref{fig:FERL_Qual}, we illustrate the \ac{GT} feature values $\phi_{\text{True}}$ and the trained features $\phi_{\psi}$ by projecting the test points on 2D sub-spaces and plotting the average feature value per 2D grid point.
For Euclidean features we used the \ac{EE}'s \textit{xy}-plane or  \textit{yz}-plane (\textit{table}), and for \textit{coffee} we project the $x$-axis basis vector of the \ac{EE} after forward kinematic rotations onto the \textit{xz}-plane (arrow up represents the cup upright). White pixels are an artifact of sampling.

We observe that $\phi_{\psi}$ resembles $\phi_{\text{True}}$ very well for most features. Our most complex feature, \textit{between objects}, does not recreate the \ac{GT} as well, although it does learn the general shape. 
However, we note in App. \ref{app:betweenobjects} that in smaller raw input space it is able to learn the fine-grained GT structure.
This implies that spurious correlation in input space is a problem, hence for complex features more data or active learning methods to collect informative traces are required.

\subsubsection{Quantitative Analysis.} 
Fig. \ref{fig:FERL_MSE} displays the means and standard errors across 10 seeds for each feature with increasing amount of data $N$. To test H1, we look at the errors with the maximum amount of data. Indeed, \ac{FERL} achieves small errors, put in context by the comparison with the error a random feature incurs (gray line). This is confirmed by an ANOVA with random vs. \ac{FERL} as a factor and the feature ID as a covariate, finding a significant main effect ($F(1,113)=372.0123,p<.0001$). In line with H2, most features have decreasing error with increasing data. Indeed, an ANOVA with $N$ as a factor and feature ID as a covariate found a significant main effect ($F(8,526)=21.1407,p<.0001$). Lastly, supporting H3, we see that the standard error on the mean decreases when \ac{FERL} gets more data. To test this, we ran an ANOVA with the standard error as the dependent measure and $N$ as a factor, finding a significant main effect ($F(8,45)=3.098,p=.0072$).

\begin{figure}
  \centering
  \includegraphics[width=0.43\textwidth]{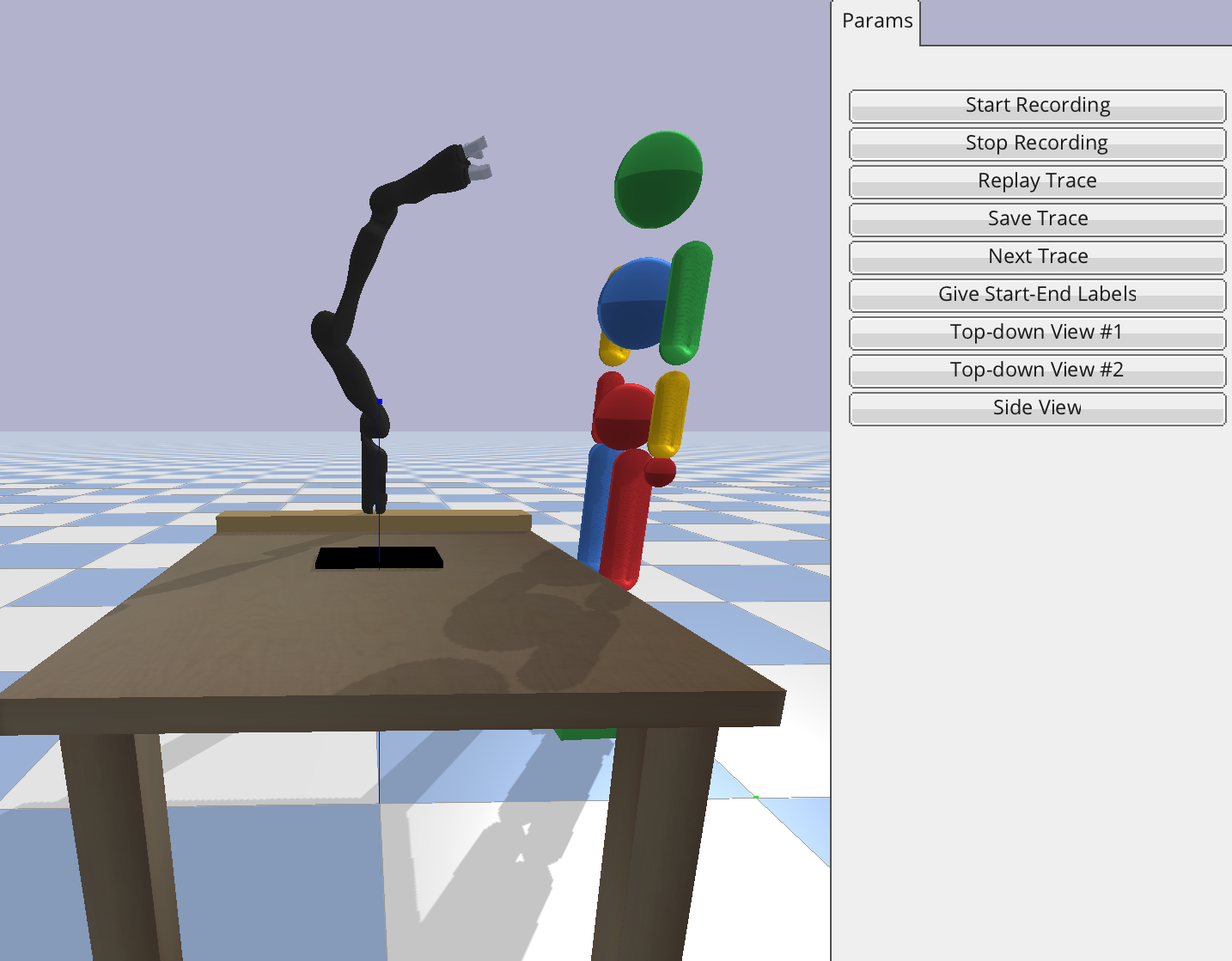}
   \caption{The pybullet simulator interface used in the user study, replicating our lab setup with the JACO robot.}
\label{fig:simulator}
\end{figure}

\subsubsection{Summary.} The qualitative and quantitative results support our hypotheses and suggest that our method requires few traces to reliably learn features $\phi_{\psi}$ that generalize well to states not seen during training. We also find that the more complex a feature, the more traces are needed for good performance: while \textit{table} and \textit{laptop} perform well with just $N\!=\!4$, some other features, like \textit{between objects}, require more traces. Active learning approaches that disentangle the learned function by querying traces at parts of the state space that are confusing could further reduce the amount of data required.

\subsection{User Study}

\label{sec:feature_user_exps}

In the previous section, we have demonstrated that experts can teach the robot good feature functions.
We now design a user study to test how well non-expert users can teach features with \ac{FERL} and how easily they can use the \ac{FERL} protocol. 

\subsubsection{Experimental Design.}
Due to COVID, we replicated our set-up from Fig. \ref{fig:front_fig} (Left) in a pybullet simulator~\citep{coumans2019} in which users can move a 7 DoF-JACO robotic arm using their cursor. Through the interface in Fig. \ref{fig:simulator}, the users can drag the robot to provide feature traces, and use the buttons for recording, saving, and discarding them.

The user study is split into two phases: familiarization and teaching. In the first phase, we introduce the user to the task context, the simulation interface, and how to provide feature traces through an instruction video and a manual. Next, we describe and 3D visualize the familiarization task feature \textit{human} (0.3 meter $xy$-plane distance of the \ac{EE} to the human position), after which we ask them to provide 10 feature traces to teach it. Lastly, we give the users a chance to see what they did well and learn from their mistakes by showing them a 3D visualization of their traces and the learned feature. \change{See App. \ref{app:user_instructions} for more details on the user training.}

In the second phase, we ask users to teach the robot three features from Sec. \ref{sec:feature_expert_exps}: \textit{table}, \textit{laptop}, and \textit{proxemics}. This time, we don't show the learned features until after all three tasks are complete.

\paragraph{Manipulated Variables.}
We manipulate the \textit{input type} with three levels: \textit{Random}, \textit{Expert}, and \textit{User}. For \textit{Random}, we randomly initialize 12 feature functions per task; for \textit{Expert}, the authors collected 20 traces per task in the simulator, then \textit{randomly} subsampled 12 sets of 10 that lead to features of similar MSEs to the ones in the physical setup before;
for \textit{User}, each person provided 10 traces per task.

\begin{figure}
\includegraphics[width=.48\textwidth]{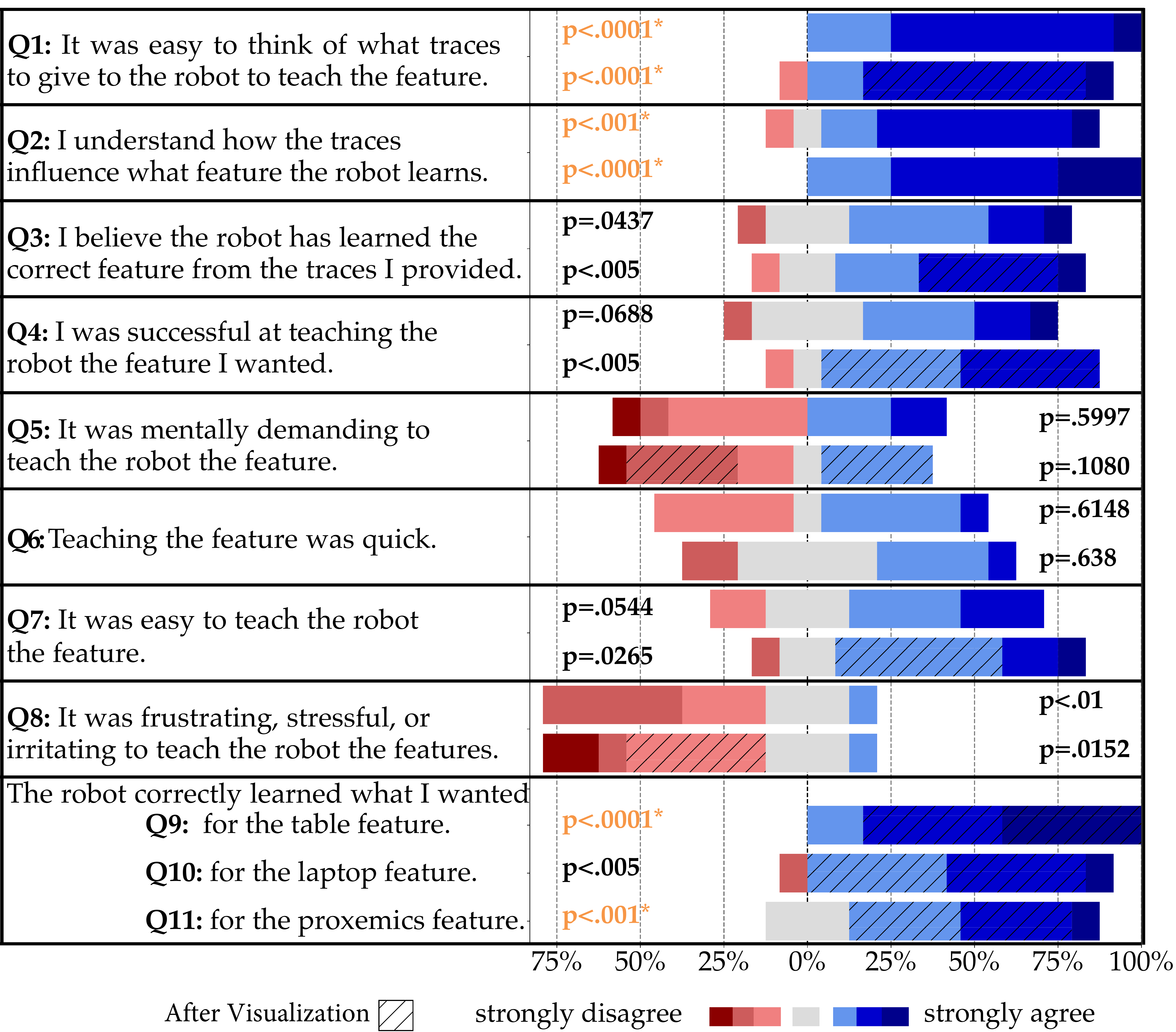}
\centering
\caption{Questions, answer distributions, and p-values 
(2-sided t-test against the middle score 4) from the user study. The p-values in orange are significant after adjusted for multiple comparisons using the Bonferroni correction.}
\label{fig:study_qual}
\end{figure}

\paragraph{Dependent Measures.}
Our objective metric is the learned feature's \ac{MSE} compared to the \ac{GT} feature on $\mathcal{S}_{\text{Test}}$, similar to Sec. \ref{sec:feature_expert_exps}. Additionally, to assess the users’ interaction experience we administered the subjective 7-point Likert scale survey from Fig. \ref{fig:study_qual}, with some items inspired by NASA-TLX~\citep{HART1988NASATLX}. After they provide the feature traces for all 3 tasks, we ask the top eight questions in Fig. \ref{fig:study_qual}. The participants then see the 3D visualizations of their feature traces and learned features, and we survey all 11 questions as in Fig. \ref{fig:study_qual} to see if their assessment changed.




\paragraph{Participants.}
We recruited 12 users (11 male, aged 18-30) from the campus community to interact with our simulated JACO robot and provide feature traces for the three tasks. All users had technical background, so we caution that our results will speak to \ac{FERL}'s usability with this population rather than the general population.

\paragraph{Hypotheses.} \hfill

\noindent\textbf{H4:} \ac{FERL} learns good features from non-expert user data. 

\noindent\textbf{H5:} Users find it easy to think of traces to give the robot, believe they understand how these traces influence the learned feature, believe they were successful teachers, and find our teaching protocol intuitive (little mental/physical effort, time, or stress).

\begin{figure}
  \centering
  \includegraphics[width=0.4\textwidth]{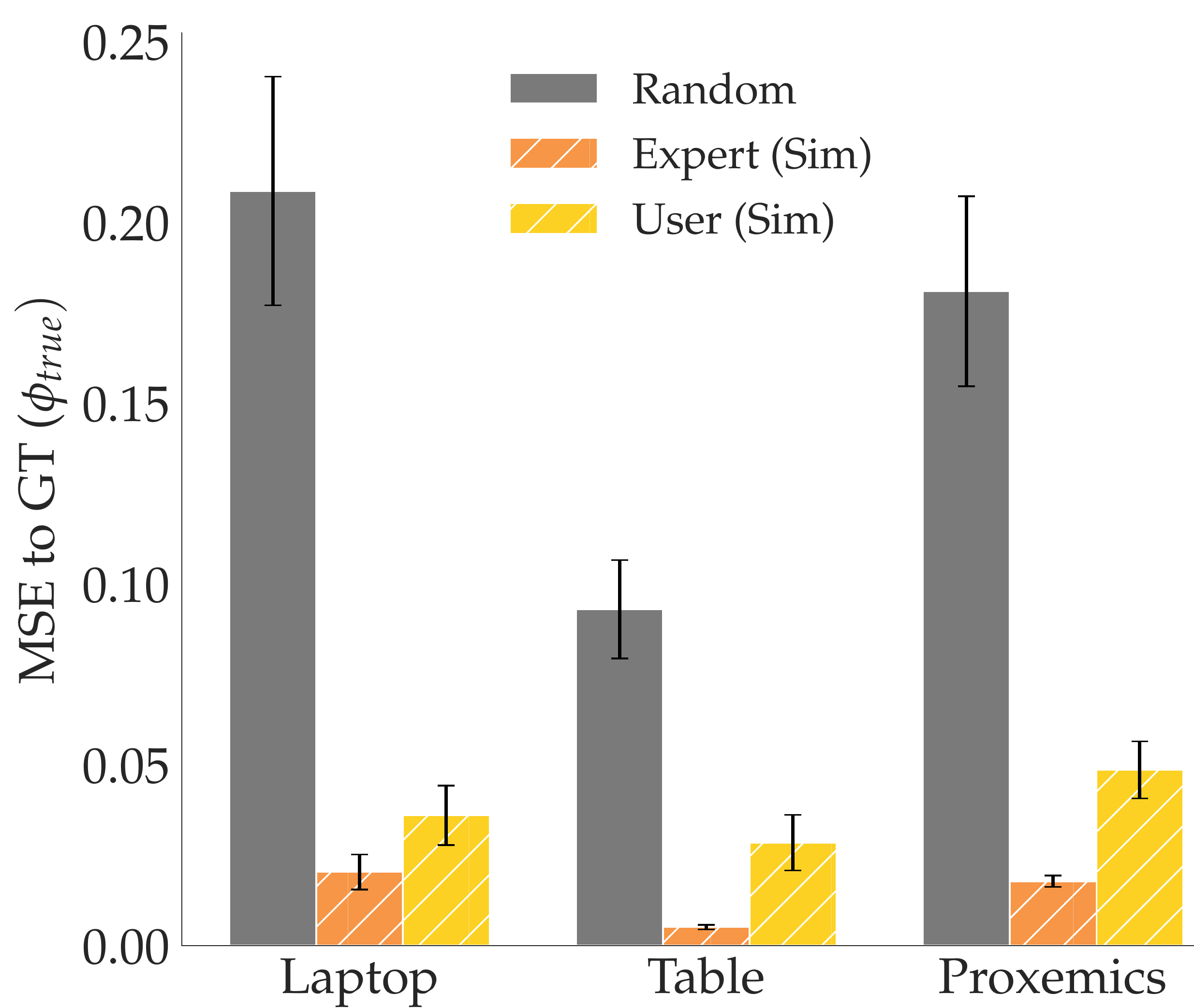}
  \caption{MSE to GT for the three features learned from expert (orange) and user (yellow) traces provided in simulation, and randomly (gray) initialized feature for comparison.}
   \label{fig:study_featureMSE}
\end{figure}

\subsubsection{Analysis}

\paragraph{Objective.} Fig. \ref{fig:study_featureMSE} summarizes the results by showing how the \ac{MSE} varies with each of our input types, for each task feature. Right off the bat, we notice that in line with H4, the MSEs for the user features are much closer to the expert level than to random. We ran an ANOVA with input type as a factor and task as a covariate, finding a significant main effect (F(2, 103) = 132.7505, p < .0001). We then ran a Tukey HSD post-hoc, which showed that the MSE for \textit{Random} input was significantly higher than both \textit{Expert} (p < .0001) and \textit{User} (p < .0001), and found no significant difference between \textit{Expert} and \textit{User} (p = .0964).
While this does not mean that user features are as good as expert features (we expect some degradation in performance when going to non-experts), it shows that they are substantially closer to them than to random, i.e. the user features maintain a lot of signal despite this degradation.

\paragraph{Subjective.} In Fig. \ref{fig:study_qual}, we see the Likert survey scores before and after the users saw the teaching results. For every question, we report 2-sided t-tests against the neutral score 4. These results support H5, although the evidence for finding the teaching protocol intuitive is weaker, and participants might have a bias to be positive given they are in a study.
In fact, several participants mentioned in their additional remarks that they had a good idea of what traces to give, and the only frustrating part was the GUI interface, which was necessary because in-person studies are not possible during the COVID pandemic
("I had a pretty good mental model for what I wanted to show, but found it frustrating doing that with a mouse", "I know what it wants, but the interface makes it difficult to give those exact traces"); performing the experiment as it was originally intended with the real robot arm would have potentially alleviated this issue ("With manual control of the arm it would have been a lot easier.").

Looking before and after the visualization, we find a trend: seeing the result seems to reinforce people’s belief that they were effective teachers (Q3, Q4), also noticed in their comments ("Surprising how well it learned!", "Surprised that with limited coverage it generalized pretty well."). Also, in support of H4, we see significant evidence that users thought the robot learned the correct feature (Q9-Q11).

Lastly, we wanted to know if there was a correlation between subjective scores and objective performance. We isolated the ``good teachers'' -- the participants who scored better than average on all 3 feature tasks in the objective metric, and compared their subjective scores to the rest of the teachers. By running a factorial likelihood-ratio test for each question, we found a significant main effect for good teachers: they are more certain that the robot has learned a correct feature even before seeing the results (Q3, p = .001), are more inclined to think they were successful (Q4, p = .0203), and find it significantly easier to teach features (Q7, p = .0202).

\subsubsection{Summary.}

Both the objective and subjective results provide evidence that non-expert users can teach the robot reasonable features using our \ac{FERL} protocol. In addition, participants found our teaching protocol intuitive, suggesting that feature traces can be useful for teaching features outside of the system designer's setting. In the following sections, we explore whether both expert and non-expert features can be used to improve reward learning generalization.

\section{Experiments: Online \ac{FERL}}
\label{sec:online_FERL_exps}

Now that we have tested our method for learning features with both experts and non-experts, we analyze how the learned features affect reward learning. In this section, we start with the easier setting where the robot already has a feature set that it is using for online reward learning, but the human might provide input about a missing feature.

\subsection{Expert Users}
\label{sec:online_FERL_expert_exps}

\begin{figure}
\begin{subfigure}{.47\textwidth}
  \centering
  \includegraphics[width=\textwidth]{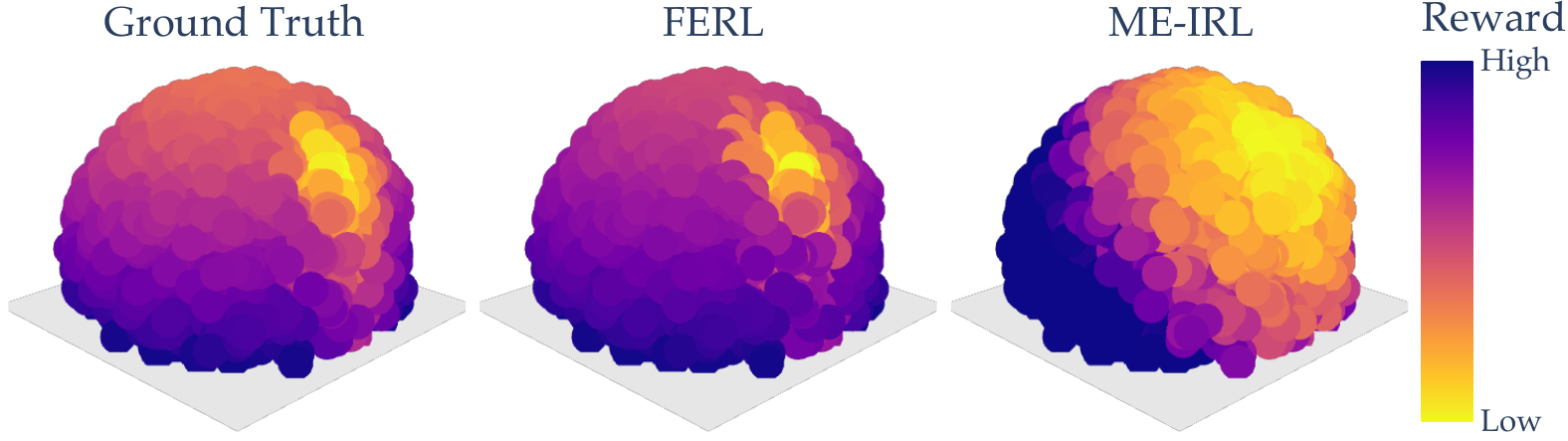}
\end{subfigure}%
\\
\begin{subfigure}{.47\textwidth}
  \centering
  \includegraphics[width=\textwidth]{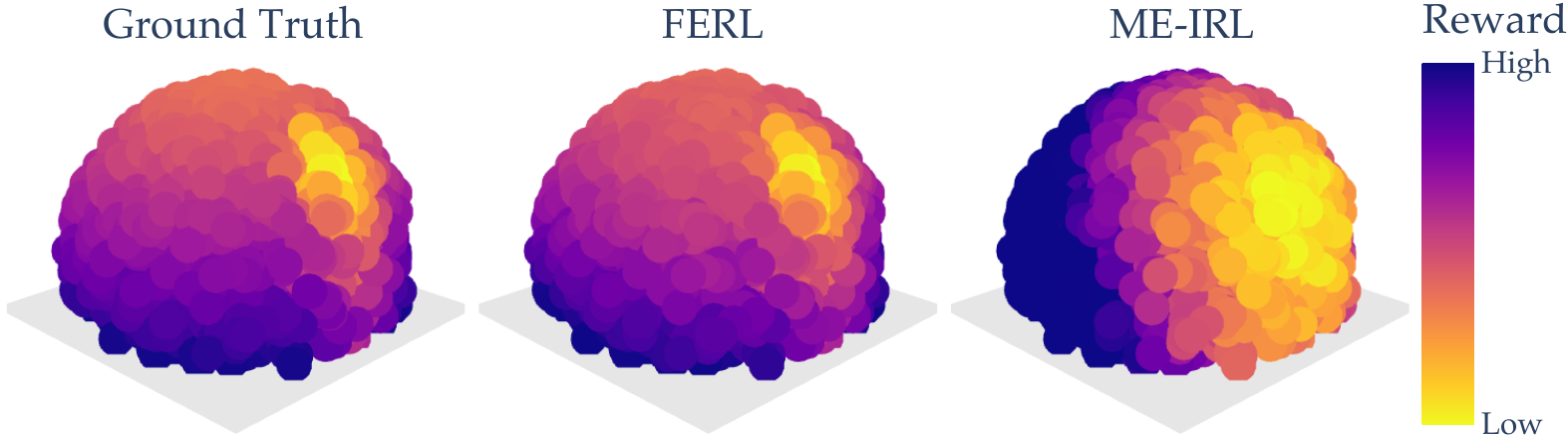}
\end{subfigure}%
\\
\begin{subfigure}{.47\textwidth}
  \centering
  \includegraphics[width=\textwidth]{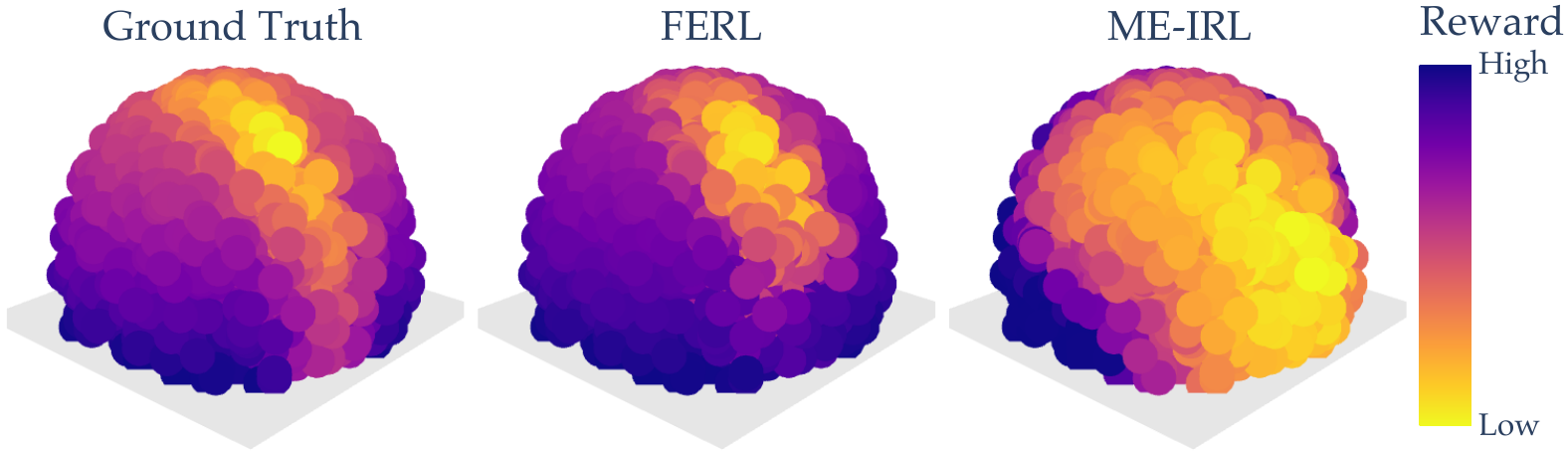}
\end{subfigure}
\caption{Visual comparison of the ground truth, online \ac{FERL}, and ME-IRL rewards for \textit{Laptop Missing} (top), \textit{Table Missing} (middle) and \textit{Proxemics Missing} (bottom).}
\label{fig:VisualComparison}
\end{figure}

When the robot receives human input that cannot be explained by its current set of features, we hypothesize that adding \ac{FERL} features to it can induce structure in the reward learning procedure that helps better recover the person's preferences. We first test this hypothesis with expert user data.

\subsubsection{Experimental Design.} We run experiments on the same JACO robot arm in three settings in which two features are known ($\phi_{\text{coffee}}, \phi_{\text{known}}$) and one is unknown. In all tasks, the true reward is $r_{\text{true}}\!=\!(0, 10, 10)(\phi_{\text{coffee}}, \phi_{\text{known}}, \phi_{\text{unknown}})^T$.
We include $\phi_{\text{coffee}}$ with zero weight to evaluate if the methods can learn to ignore irrelevant features.
In task 1, $\phi_{\text{laptop}}$ is unknown and the known feature is $\phi_{\text{table}}$; in task 2, $\phi_{\text{table}}$ is unknown and $\phi_{\text{laptop}}$ is known; and in task 3, $\phi_{\text{proxemics}}$ is unknown and $\phi_{\text{table}}$ is known. We name the tasks \textit{Laptop Missing}, \textit{Table Missing}, and \textit{Proxemics Missing}, respectively.

\paragraph{Manipulated Variables.} We manipulate the \textit{learning method} with 2 levels: \ac{FERL} and an adapted \ac{ME-IRL} baseline\footnote{\change{We chose \ac{ME-IRL} as it is the state-of-the-art method for learning rewards and does not rely on base feature engineering, as explained in Section \ref{sec:related}.} 
\change{We also tried a linear variant of \ac{ME-IRL} optimizing the reward parameters on top of random features modeled as neural networks. However, we found the performance of this alternate baseline to be consistently inferior to that of the deep \ac{ME-IRL} (see App. \ref{app:randomfeatures}), so we only compare against the deep variant.}} \citep{finn2016gcl, wulfmeier2016maxentirl} learning a deep reward function from demonstrations. 
We model the \ac{ME-IRL} reward function $r_{\omega}$ as a neural network with 2 layers, 128 units each. For a fair comparison, we gave $r_{\omega}$ access to the known features: once the 27D Euclidean input is mapped to a neuron, a last layer combines it with the known feature vector.

Also for a fair comparison, we took great care to collect a set of demonstrations for \ac{ME-IRL} designed to be as informative as possible: we chose diverse start and goal configurations for the demonstrations, and focused some of them on the unknown feature and some on learning a combination between features (see App. \ref{app:MEIRL_demos}). 
Moreover, \ac{FERL} and \ac{ME-IRL} rely on different input types: \ac{FERL} on feature traces $\trace$ and pushes $a_H$ and \ac{ME-IRL} on a set of near-optimal demonstrations $\mathcal{D}^*$. To level the amount of data each method has access to, we collected the traces $\Xi$ and demonstrations $\mathcal{D}^*$ such that \ac{ME-IRL} has more data points: the average number of states per demonstration/trace were 61 and 31, respectively.


Following Eq. \eqref{eq:general_loglikelihood_gradient}, the gradient of the \ac{ME-IRL} objective with respect to the reward parameters $\omega$ can be estimated by: $\nabla_{\omega}\mathcal{L} \!\approx\! \frac{1}{|\mathcal{D}^*|}\sum_{\tau \in \mathcal{D}^*} \!\!\nabla_{\omega}R_{\omega}(\tau) \!-\! \frac{1}{|\mathcal{D}^{\omega}|}\sum_{\tau \in \mathcal{D}^{\omega}}\!\! \nabla_{\omega}R_{\omega}(\tau)$ \citep{wulfmeier2016maxentirl,finn2016gcl}. 
Here, $R_{\omega}(\tau)\!=\!\sum_{s \in \tau}\!r_{\omega}(\sysstate)$ is the parametrized reward, $\mathcal{D}^*$ the expert demonstrations, and $\mathcal{D}^{\omega}$ are trajectory samples from the $r_{\omega}$ induced near optimal policy.
We use TrajOpt \citep{schulman2013trajopt} to obtain the samples $\mathcal{D}^{\omega}$ (see App. \ref{app:MEIRL_implementation} for details).
For practical considerations and implementation details of the online version of \ac{FERL} we used, see App. \ref{app:online_FERL_implementation}.

\begin{figure}
  \centering
  \includegraphics[width=0.48\textwidth]{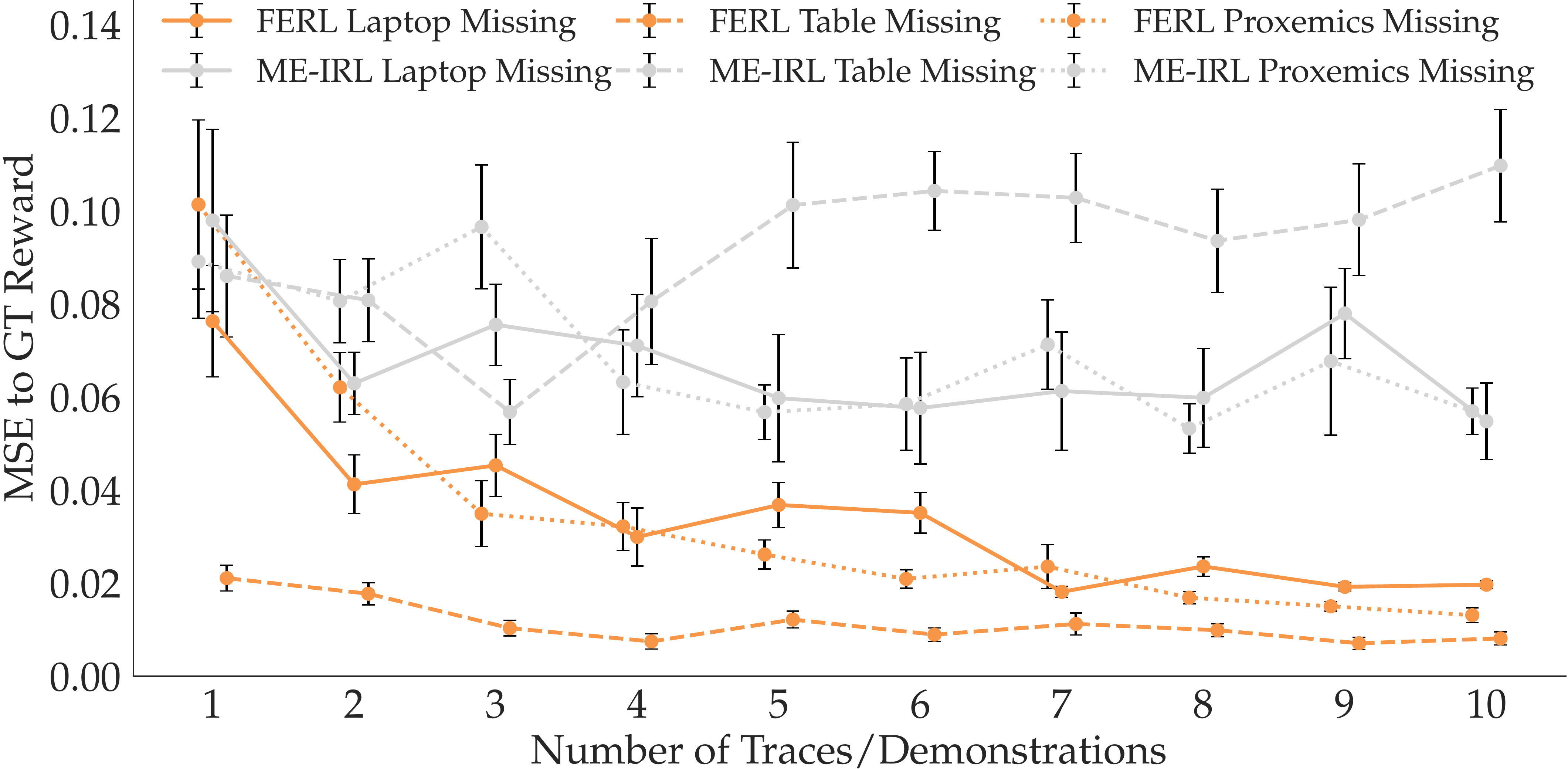}
 \caption{MSE of online \ac{FERL} and \ac{ME-IRL} to \ac{GT} reward across all three tasks. \ac{FERL} learns rewards that better generalize to the state space.}
   \label{fig:MSEComparison}
\end{figure}

\paragraph{Dependent Measures.}

We compare the two reward learning methods across \change{three} metrics commonly used in the \ac{IRL} literature~\citep{choi2011inverse}: 1) \textit{Reward Accuracy}: how close to \ac{GT} the learned reward is, 2) \textit{Behavior Accuracy}: how well do the behaviors induced by the learned rewards compare to the \ac{GT} optimal behavior, measured by evaluating the induced trajectories on \ac{GT} reward, \change{and 3) \textit{Test Probability}: how likely trajectories generated by the \ac{GT} reward are under the learned reward models.}

\change{For \textit{Reward Accuracy}, note that any affine transformation of a reward function would result in the same induced behaviors, so simply measuring the \ac{MSE} between the learner's reward and the \ac{GT} reward may not be informative. 
As such, we make reward functions given by different methods comparable by computing each learner's reward values on $\mathcal{S}_{\text{Test}}$ and normalizing the resulting set of rewards to be in $[0,1]$. 
This allows us to compute the \ac{MSE} on $\mathcal{S}_{\text{Test}}$ between each method and the \ac{GT}. 
Similarly to Sec. \ref{sec:feature_expert_exps}, we report this metric by varying the number of traces / demonstrations each learner gets access to.}
For \textit{Behavior Accuracy} \change{and \textit{Test Probability}}, we train \ac{FERL} and \ac{ME-IRL} with a set of 10 traces / demonstrations. 
\change{For \textit{Behavior Accuracy}, we use} TrajOpt~\citep{schulman2013trajopt} to produce optimal trajectories for 100 randomly selected start-goal pairs under the learned rewards. We evaluate the trajectories with the \ac{GT} reward $r_{\text{true}}$ and divide by the reward of the \ac{GT} induced trajectory for easy relative comparison.
\change{For \textit{Test Probability}, we generate 100 optimal trajectories using the \ac{GT} reward, then evaluate their likelihood under the Boltzmann model in Eq. \eqref{eq:observation_model} with each learned reward. \changenew{To approximate the intractable integral in Eq. \eqref{eq:observation_model}, we sample\footnote{\changenew{To obtain dynamically feasible trajectories, we sampled random objectives given by linear combinations of various features, and optimized them with TrajOpt. While this sampling strategy cannot be justified theoretically, it works well in practice: the resulting optimized trajectories are a heuristic for sampling diverse and interesting trajectories in the environment.}} sets of 100 trajectories for every start-goal pair corresponding to the optimal trajectories.}  For a fair comparison, we use the normalized rewards once again, and fit the maximum likelihood coefficient $\hat\beta$ for each model.}

\paragraph{Hypotheses.} \hfill

\noindent\textbf{H6:} Online \ac{FERL} learns rewards that better generalize to the state space than \ac{ME-IRL}. 

\noindent\textbf{H7:} Online \ac{FERL} performance is less input-sensitive than \ac{ME-IRL}'s.

\subsubsection{Qualitative Comparison.} In Fig. \ref{fig:VisualComparison}, we show the learned \ac{FERL} and \ac{ME-IRL} rewards as well as the \ac{GT} for all three tasks evaluated at the test points.
As we can see, by first learning the missing feature and then the reward on the extended feature vector, \ac{FERL} is able to learn a fine-grained reward structure closely resembling the \ac{GT}. Meanwhile, \ac{ME-IRL} learns some structure capturing where the laptop or the human is, but not enough to result in a good trade-off between the active features.

\subsubsection{Quantitative Analysis.} To compare \textit{Reward Accuracy}, we show in Fig. \ref{fig:MSEComparison} the \ac{MSE} mean and standard error across 10 seeds, with increasing training data. We visualize results from all 3 tasks, with \ac{FERL} in orange and \ac{ME-IRL} in gray. \ac{FERL} is closer to GT than \ac{ME-IRL} no matter the amount of data, supporting H6. To test this, we ran an ANOVA with learning method as the factor, and with the task and data amount as covariates, and found a significant main effect (F(1, 595) = 335.5253, p < .0001).

Additionally, the consistently decreasing MSE in Fig. \ref{fig:MSEComparison} for \ac{FERL} suggests that our method gets better with more data; in contrast, the same trend is inexistent with \ac{ME-IRL}. Supporting H7, the high standard error that \ac{ME-IRL} displays implies that it is highly sensitive to the demonstrations provided and the learned reward likely overfits to the expert demonstrations. We ran an ANOVA with  standard error as the dependent measure, focusing on the $N=10$ trials which provide the maximum data to each method, with the learning method as the factor and the task as a covariate. We found that the learning method has a significant effect on the standard error (F(1, 4) = 12.1027, p = .0254). With even more data, this shortcoming of IRL might disappear; however, this would pose an additional burden on the human, which our method successfully alleviates. 

\begin{figure}

   \centering
  \includegraphics[width=0.44\textwidth]{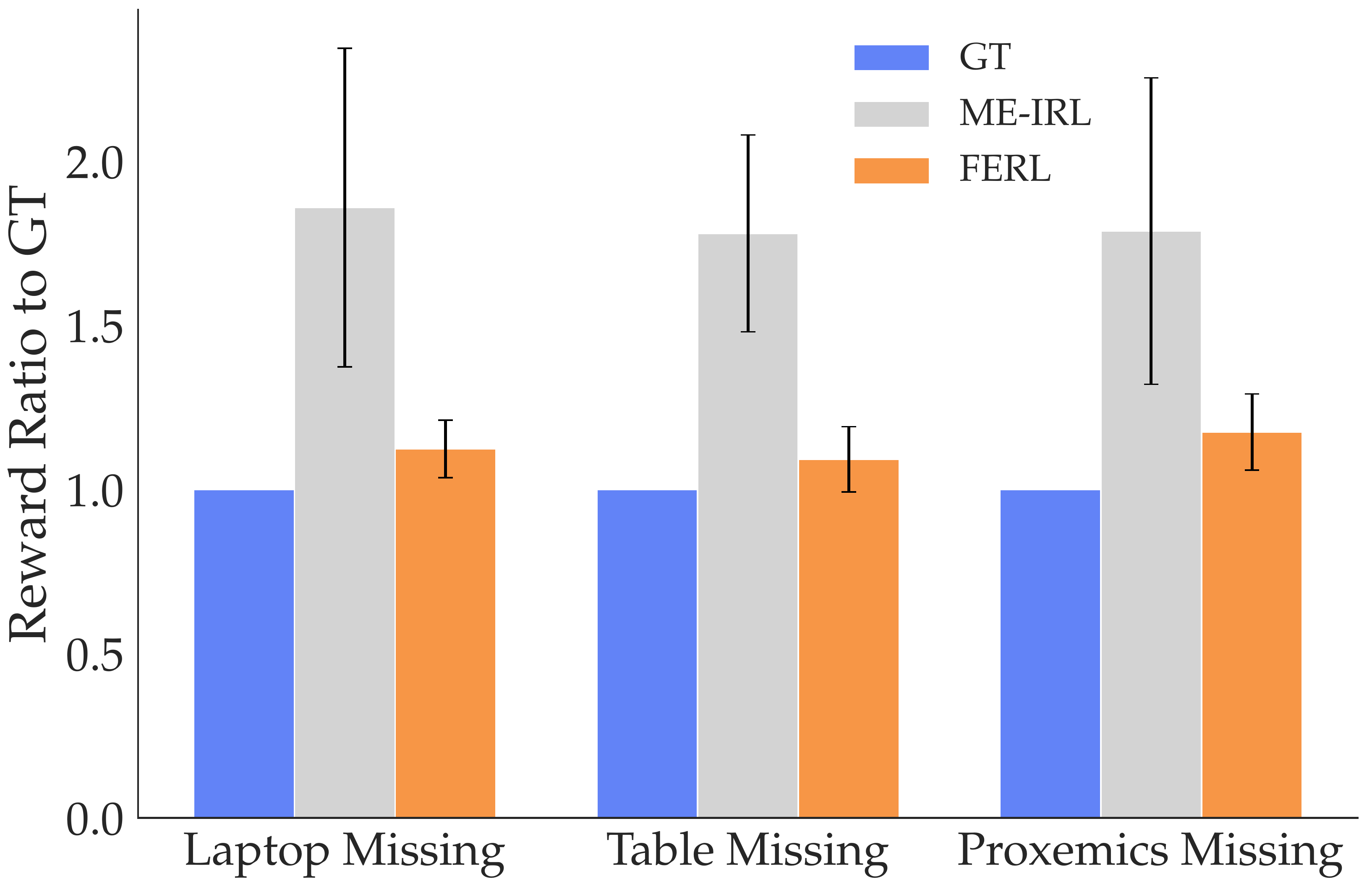}
   \caption{Induced trajectories' reward ratio for the two methods compared to \ac{GT}. \ac{ME-IRL} struggles to generalize across all tasks.}
   \label{fig:onlineFERL_InducedTrajectories}

\end{figure}

We also looked at \textit{Behavior Accuracy} for the two methods. Fig. \ref{fig:onlineFERL_InducedTrajectories} illustrates the reward ratios to \ac{GT} for all three tasks. The \ac{GT} ratio is 1 by default, and the closer to 1 the ratios are, the better the performance because all rewards are negative. The figure further supports H6, showing that \ac{FERL} rewards produce trajectories that are preferred under the \ac{GT} reward over \ac{ME-IRL} reward trajectories. An ANOVA using the task as a covariate reveals a significant main effect for the learning method (F(1, 596) = 14.9816, p = .0001).

\begin{figure}
   \centering
  \includegraphics[width=0.44\textwidth]{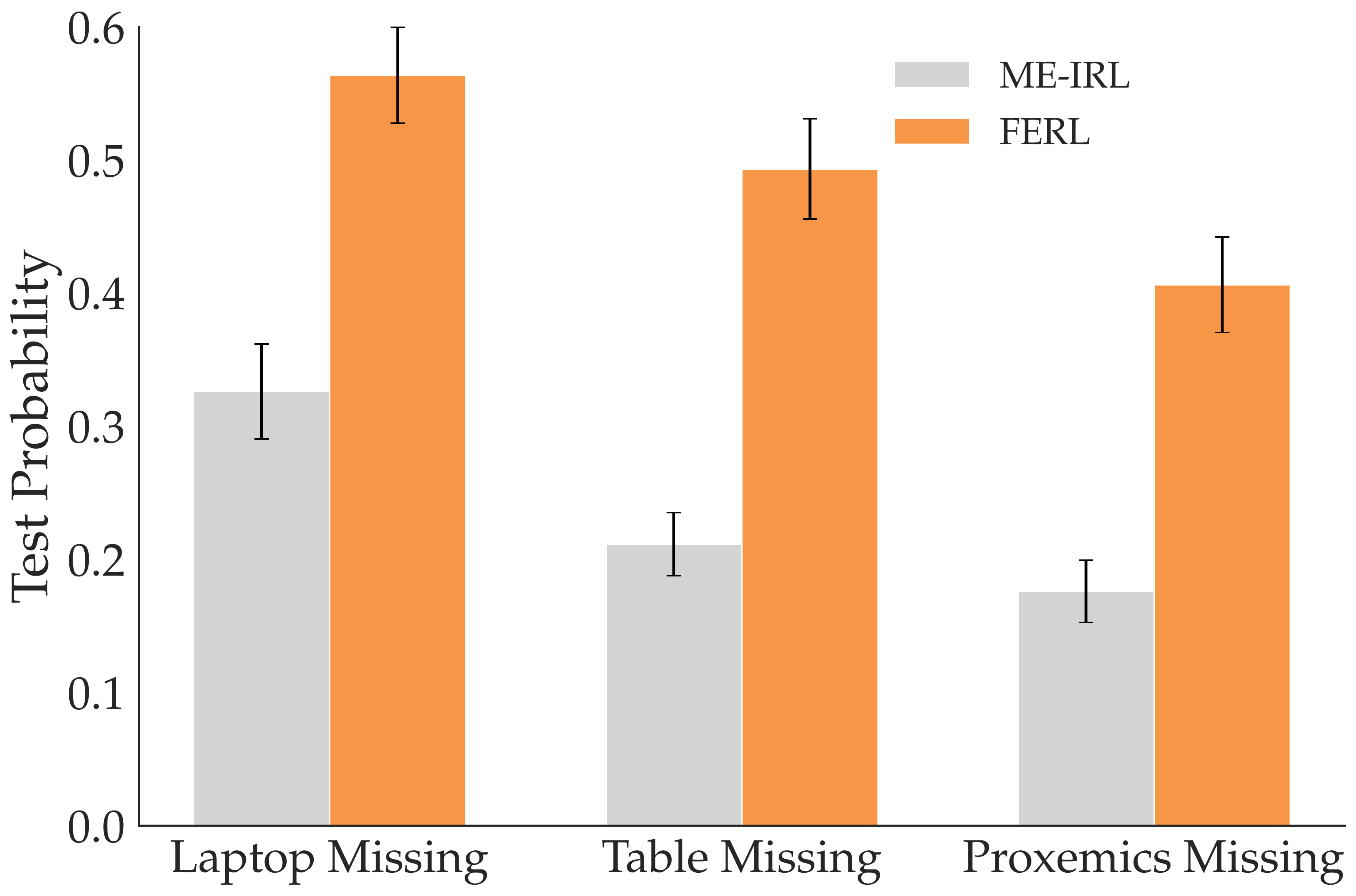}
   \caption{\change{Probability assigned by the two methods to a set of optimal trajectories under the Boltzmann assumption. The trajectories are more likely under \ac{FERL} than \ac{ME-IRL}, suggesting \ac{FERL} is the more accurate reward model.}}
   \label{fig:onlineFERL_Probabilities}
\end{figure}

\change{Lastly, we compare how likely a test set of trajectories given by optimizing the \ac{GT} reward is under the two models. A more accurate reward model should give higher probabilities to the demonstrated trajectories under the Boltzmann noisily-rational assumption in Eq. \ref{eq:observation_model}.
Fig. \ref{fig:onlineFERL_Probabilities} illustrates that \ac{FERL} does indeed assign higher likelihood to the test trajectories than \ac{ME-IRL}, which is consistent with H6.}

\subsubsection{Summary.}

The rewards learned with \ac{FERL} qualitatively capture more structure than \ac{ME-IRL} ones, but they also quantitatively get closer to the \ac{GT}. Using \ac{FERL} features -- at least when the robot is missing one feature -- seems to induce useful structure in the reward learning process that guides the robot to better capture the person's preferences. These results hold when the person teaching the missing feature is an expert user; we next look at the case where a novice interacts with the robot instead.

\subsection{Non-expert Users}
\label{sec:online_FERL_user_exps}

The objective results in Sec. \ref{sec:feature_user_exps} show that while users' performance degrades from expert performance, they are still able to teach features with a lot of signal. 
We now want to test how important the user-expert feature quality gap is when it comes to using these features for online reward learning.

\subsubsection{Experimental Design.}
For this experiment, we had a similar setup to the one in Sec. \ref{sec:online_FERL_expert_exps}, only that we performed reward learning with \ac{FERL} using the user-taught simulation features from the user study. We wanted to see if the divide-and-conquer approach employed by \ac{FERL} results in better rewards than \ac{ME-IRL} even when using noisy simulation data.

\paragraph{Manipulated Variables.}
We manipulate the \textit{learning method}, \ac{FERL} or \ac{ME-IRL}, just like in Sec. \ref{sec:online_FERL_expert_exps}. Because corrections and demonstrations would be very difficult in simulation, we use for \ac{ME-IRL} the expert data from the physical robot. For \ac{FERL}, we use the user data from the simulation, and the expert corrections that teach the robot how to combine the learned feature with the known ones. Note that this gives \ac{ME-IRL} an advantage, since its data is both generated by an expert, and on the physical robot. Nonetheless, we hypothesize that the advantage of the divide-and-conquer approach is stronger.


\paragraph{Dependent Measures.} We use the same objective metric as \textit{Reward Accuracy} in the expert comparison in Sec. \ref{sec:online_FERL_expert_exps}: the learned reward \ac{MSE} to the \ac{GT} reward on $\mathcal{S}_{\text{Test}}$.

\paragraph{Hypothesis.} \hfill

\textbf{H8:} Online FERL learns more generalizable rewards than ME-IRL even when using features learned from data provided by non-experts in simulation.

\begin{figure}
  \centering
  \includegraphics[width=0.44\textwidth]{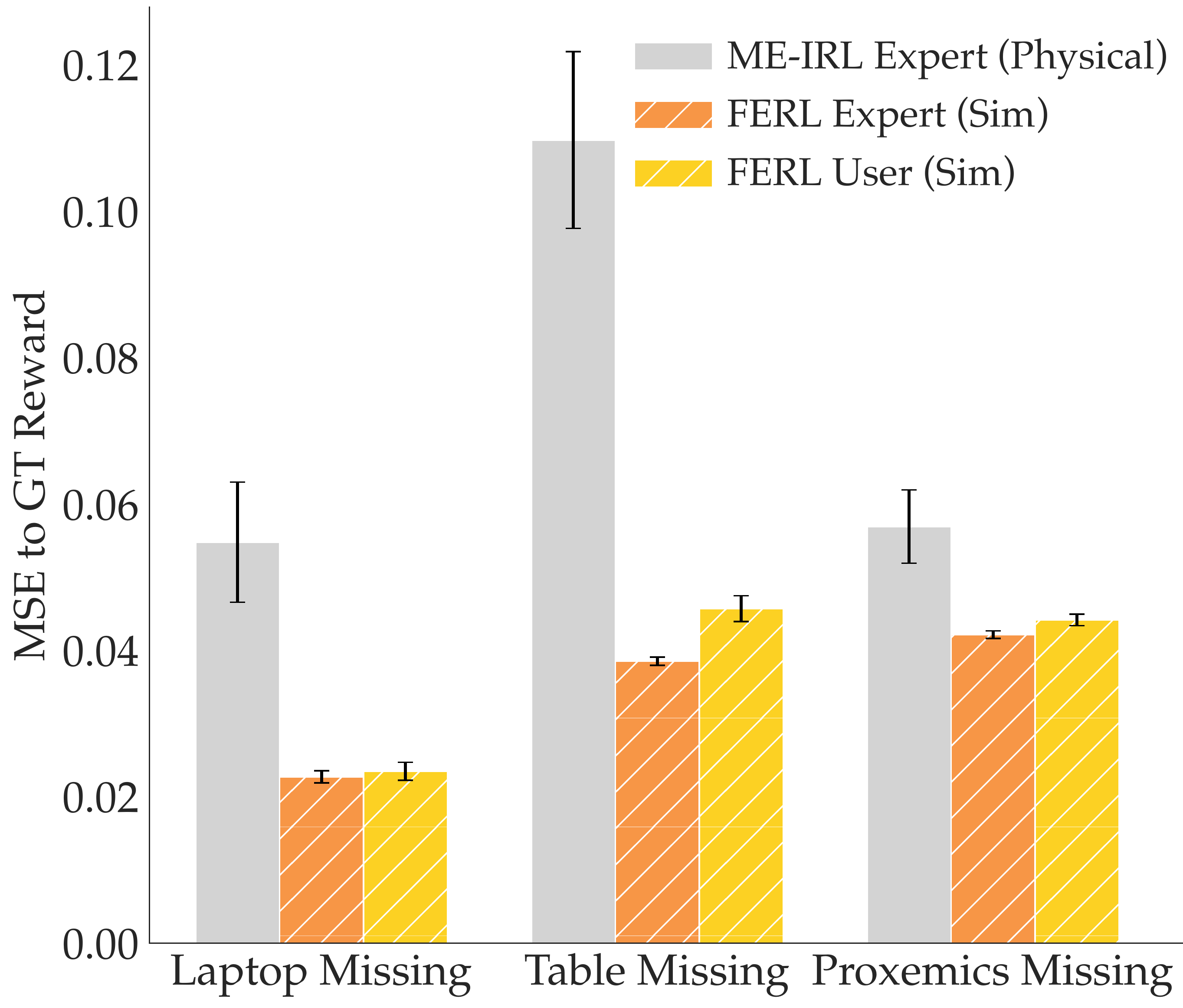}
   \caption{MSE to GT reward for the three tasks, comparing \ac{ME-IRL} from expert physical demonstrations (gray) to online \ac{FERL} from expert (orange) and non-expert (yellow) features learned in simulation and combined via corrections.}
   \label{fig:study_rewardMSE}
\end{figure}

\subsubsection{Analysis.}

Fig. \ref{fig:study_rewardMSE} illustrates our findings for the reward comparison. In the figure, we added \ac{FERL} with expert-taught simulation features for reference: we \change{randomly} subsampled sets of 10 from 20 expert traces \change{collected by the authors}, and trained 12 expert features for each of our 3 task features. We see that, \textit{even though \ac{ME-IRL} was given the advantage of using physical expert demonstrations, it still severely underperforms when compared to FERL with both expert and user features learned in simulation.} This finding is crucial because it underlines the power of our divide-and-conquer approach in online reward learning: even when given imperfect features, the learned reward is superior to trying to learn everything implicitly from demonstrations.

We verified the significance of this result with an ANOVA with the learning method as a factor and the task as a covariate. We found a significant main effect for the learning method (F(1, 62) = 41.2477, p < .0001), supporting our H8.

\subsubsection{Summary.} Despite the degradation in feature quality we see in user features when compared to expert ones, we find that the structure they do maintain is advantageous in online reward learning. This suggests that the online instantiation of \ac{FERL} can be used even by non-experts to better teach the robot their preferences.

\section{Experiments: Offline \ac{FERL}}
\label{sec:offline_FERL_exps}

\begin{figure}
\begin{subfigure}{.47\textwidth}
  \centering
  \includegraphics[width=\textwidth]{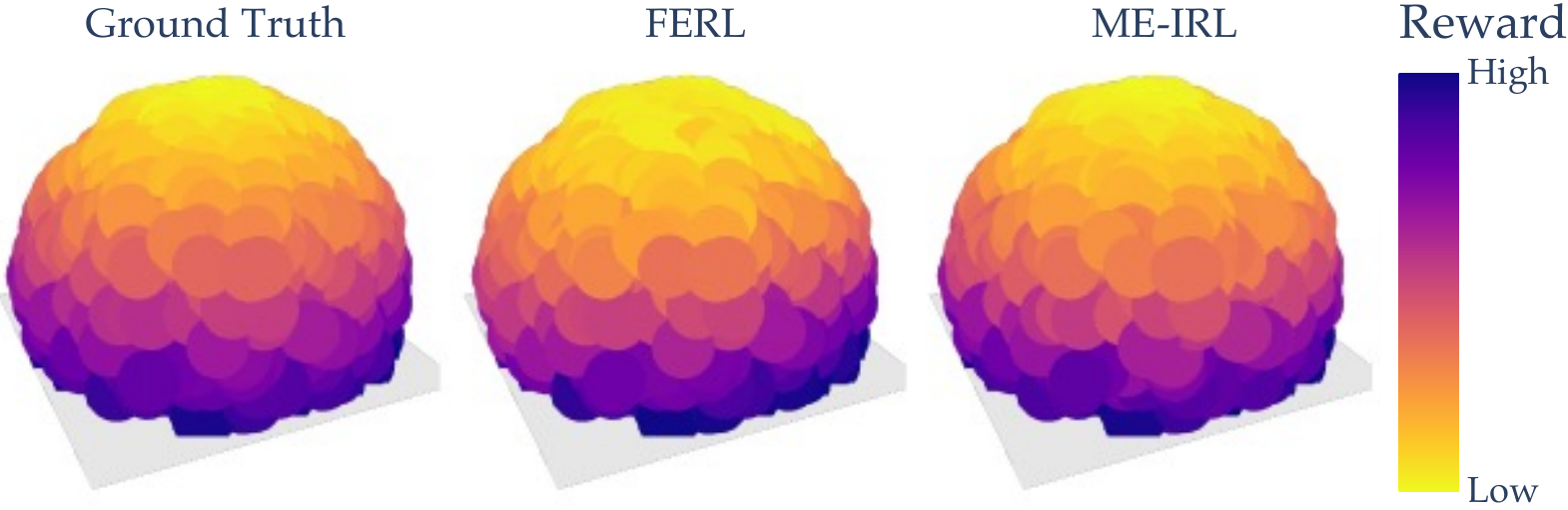}
\end{subfigure}%
\\
\begin{subfigure}{.47\textwidth}
  \centering
  \includegraphics[width=\textwidth]{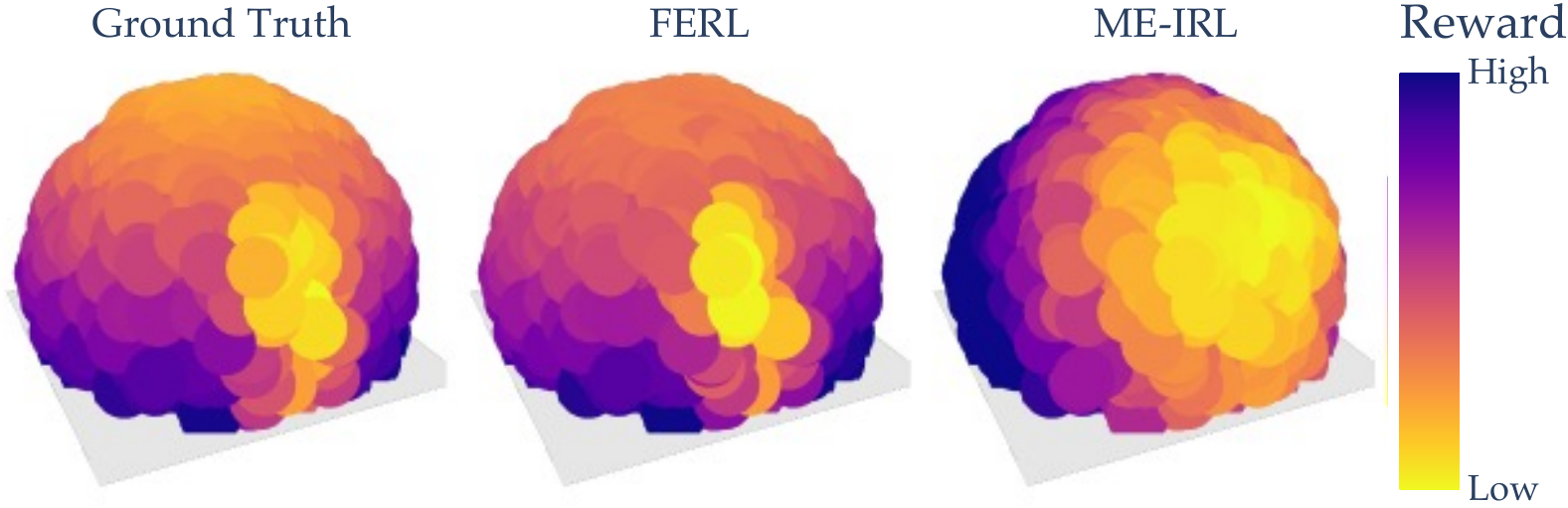}
\end{subfigure}%
\\
\begin{subfigure}{.47\textwidth}
  \centering
  \includegraphics[width=\textwidth]{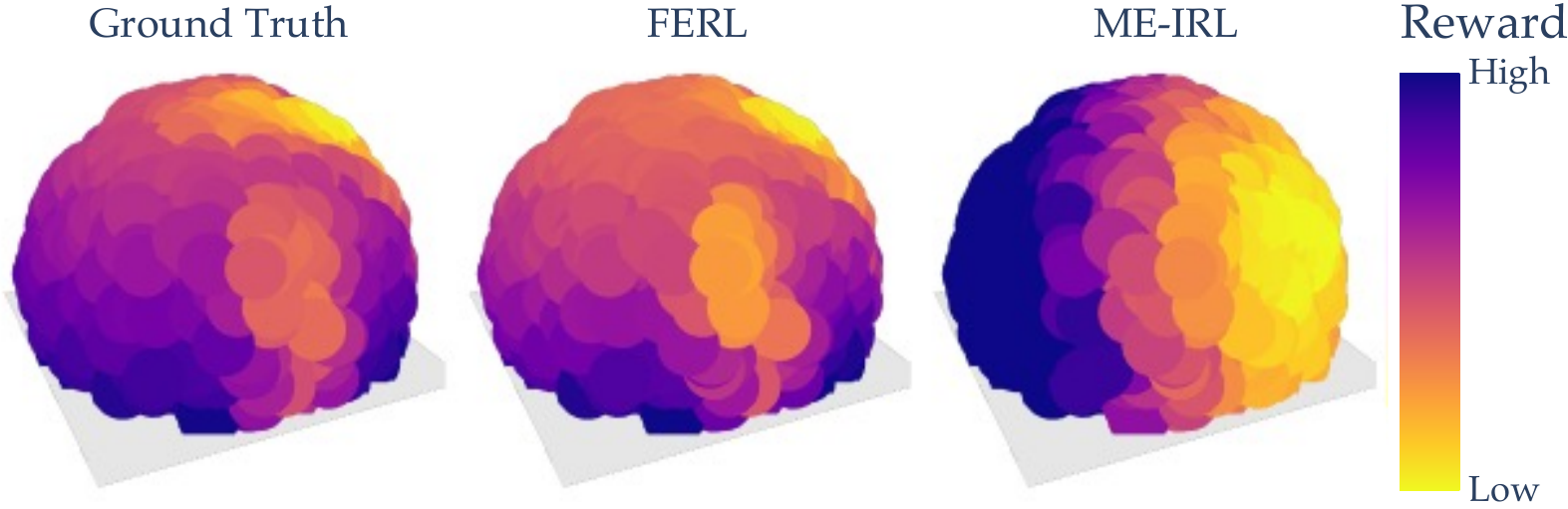}
\end{subfigure}
\caption{Visual comparison of the ground truth, offline \ac{FERL}, and ME-IRL rewards for \textit{One Feature} (top), \textit{Two Features} (middle) and \textit{Three Features} (bottom).}
\label{fig:OfflineVisualComparison}
\end{figure}

In the online reward learning setting, the robot was already equipped with a starting feature set, and we tested how learning missing features affects the reward. We now look at the scenario where the robot's reward must be programmed entirely from scratch, teaching each feature separately before combining them into a reward via demonstrations.

\subsection{Expert Users}
\label{sec:offline_FERL_expert_exps}

\begin{figure*}
\centering
\begin{subfigure}{.33\textwidth}
  \centering
  \includegraphics[width=\textwidth,left]{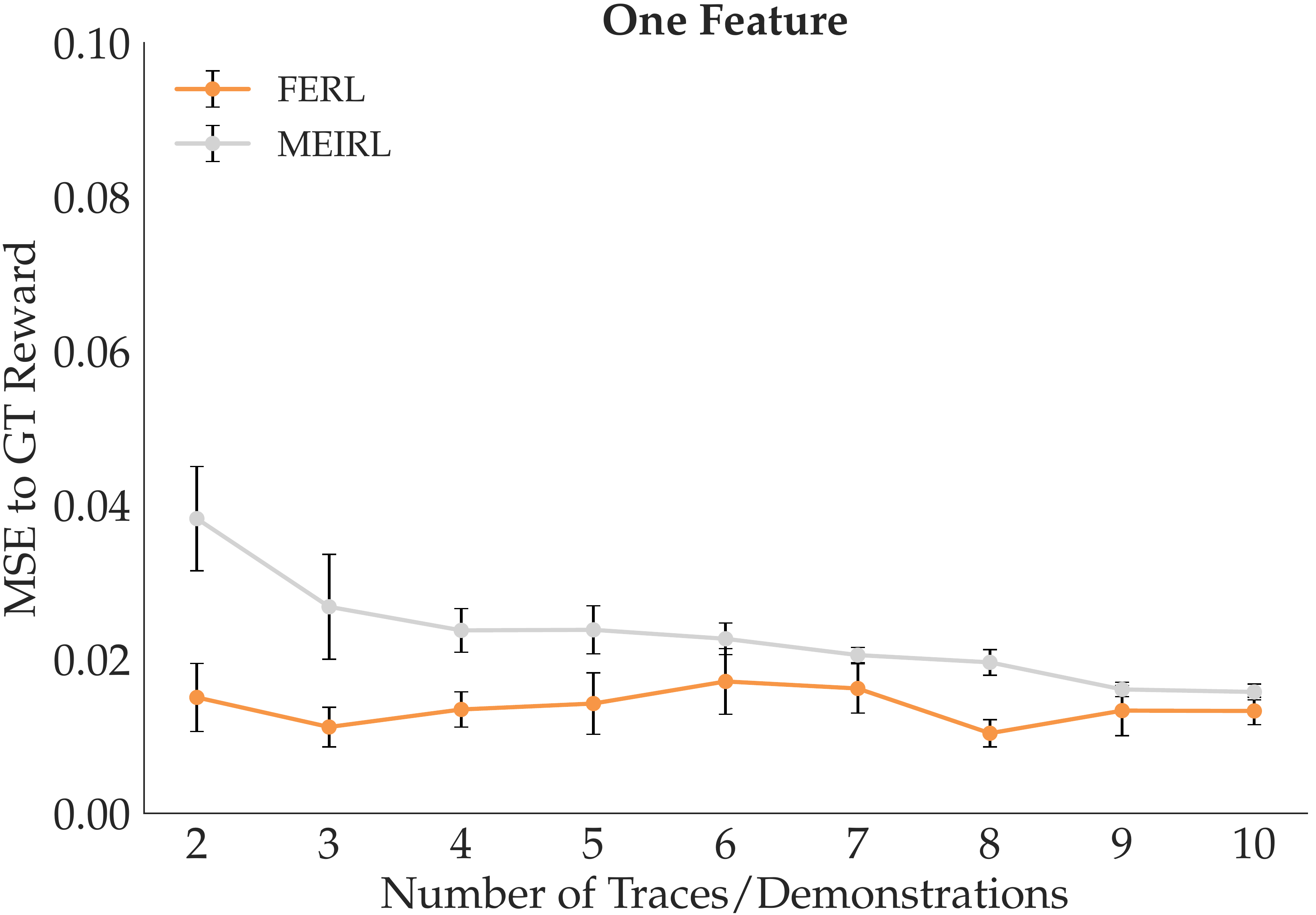}
\end{subfigure}%
\begin{subfigure}{.33\textwidth}
  \centering
  \includegraphics[width=\textwidth,left]{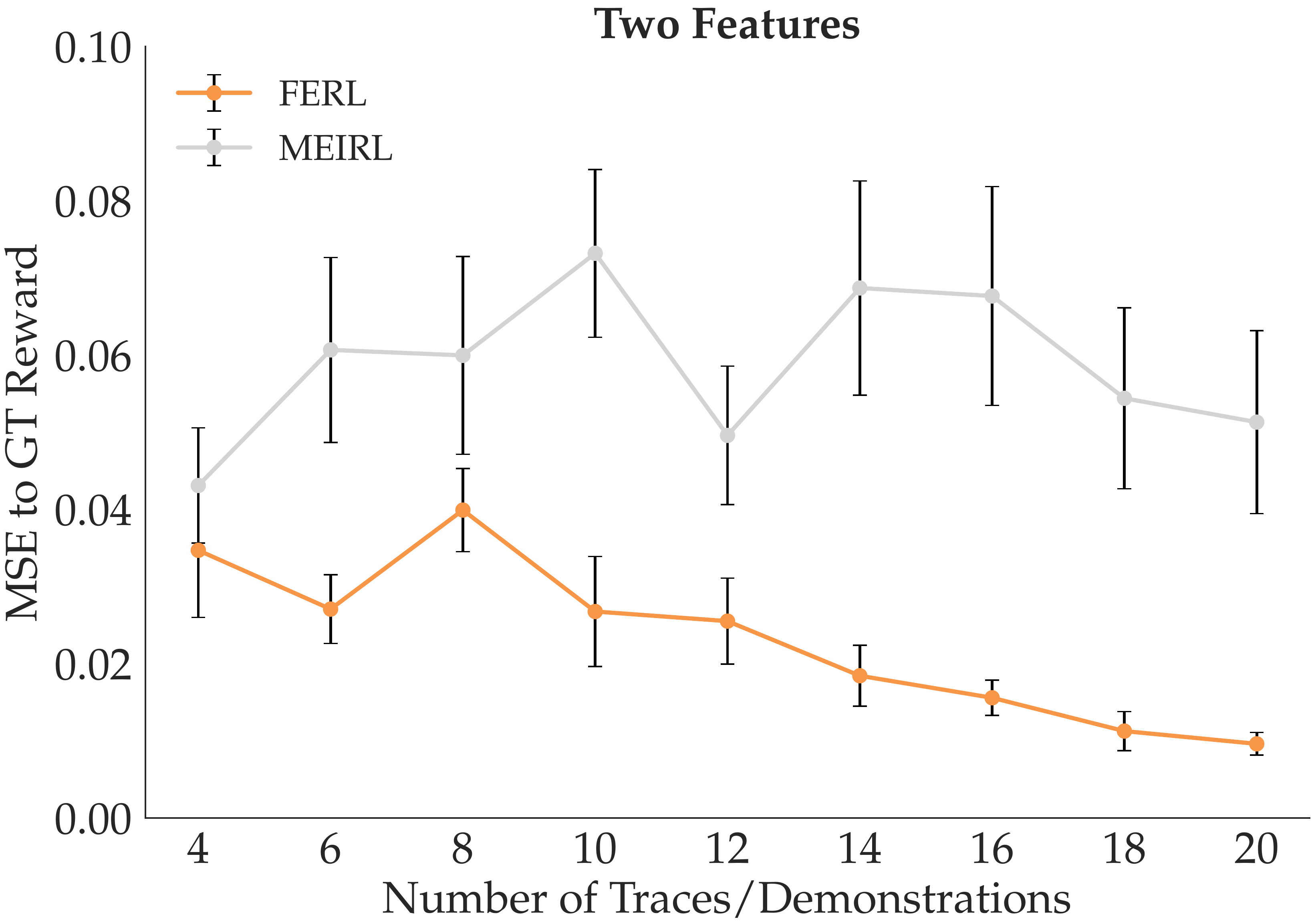}
\end{subfigure}%
\begin{subfigure}{.33\textwidth}
  \centering
  \includegraphics[width=\textwidth,left]{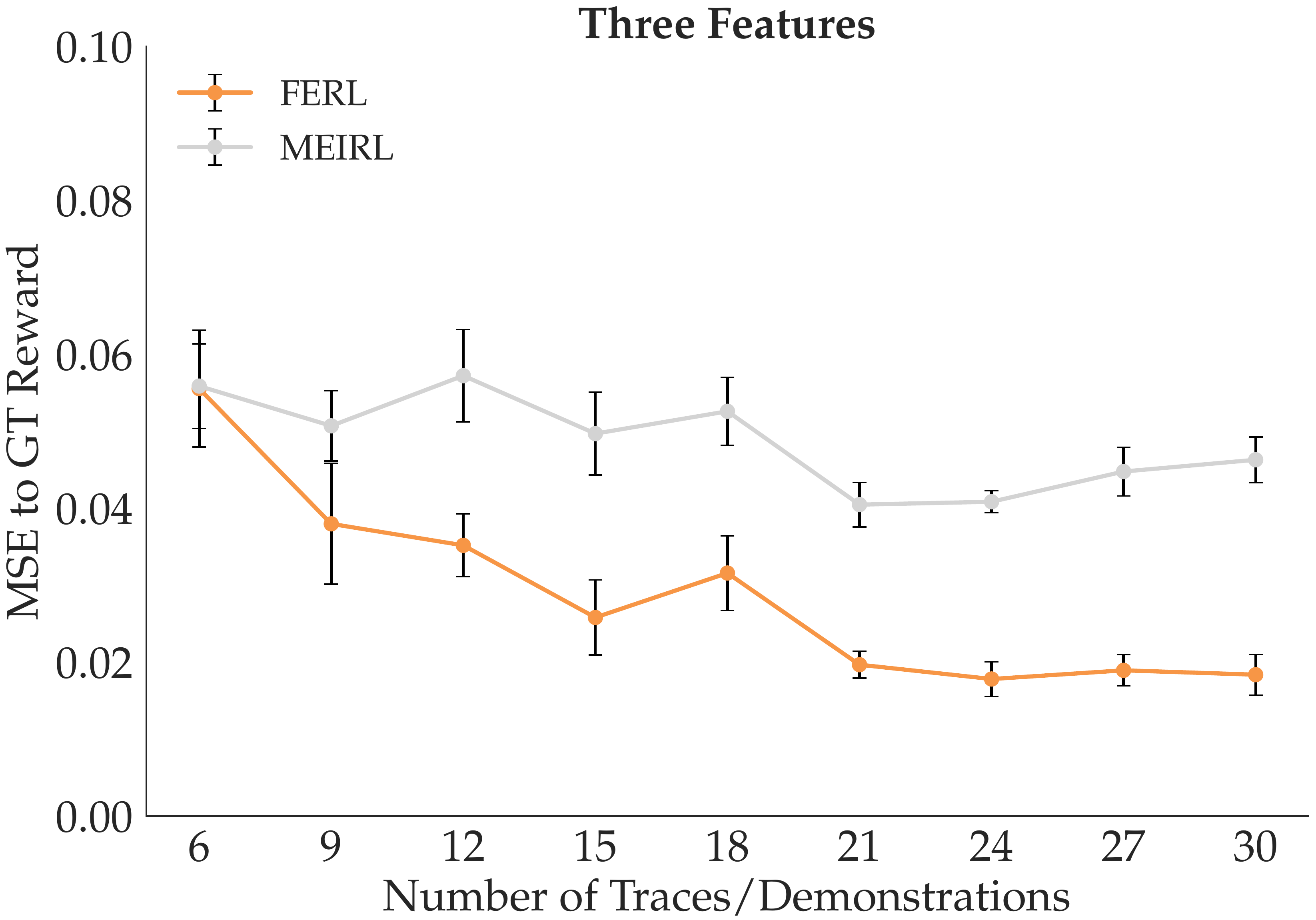}
\end{subfigure}
\caption{MSE of offline \ac{FERL} and \ac{ME-IRL} to \ac{GT} reward for \textit{One Feature} (Left), \textit{Two Features} (Middle), and \textit{Three Features} (Right). In most data regimes, \ac{FERL} learns rewards that better generalize to the state space.} 
\label{fig:offline_FERL_Cost_MSE}
\end{figure*}

We have argued that learned features can induce useful structure that speeds up reward learning. We test how the reward is affected when the entire structure is built up from the expert features taught from real robot data in Sec. \ref{sec:feature_expert_exps}.

\subsubsection{Experimental Design.}

We run experiments on the robot arm in three settings of increasing complexity: in the first, the true reward depends on a single feature, and every subsequent task adds another feature to the reward. 
In task 1, the true reward depends on only $\phi_{\text{table}}$.
In task 2, we add the $\phi_{\text{laptop}}$ feature, and in task 3 the $\phi_{\text{proxemics}}$ feature. In both tasks 2 and 3, the reward equally combines the two and three features, respectively. 
Task 1 should be easy enough for even an end-to-end \ac{IRL} method to solve, especially since it relies on the simplest feature that we have considered. Meanwhile, tasks 2 and 3 require learning rewards that are more structurally complex.
We name the three tasks \textit{One Feature}, \textit{Two Features}, and \textit{Three Features}, respectively.

\paragraph{Manipulated Variables.}

We manipulated the \textit{learning method} with 2 levels: \ac{FERL} and \ac{ME-IRL}. While in Sec. \ref{sec:online_FERL_exps} \ac{ME-IRL} had access to the known features, this time the reward network is a function mapping directly from the 27D Euclidean input space only. For practical considerations and implementation details of the offline version of \ac{FERL} we used, see App. \ref{app:offline_FERL_implementation}.

For a fair comparison, we once again took great care in how we collected the demonstrations \ac{ME-IRL} learns from. Just like before, we chose diverse start and goal configurations, and focused some of the demonstrations on each individual feature, and, when it applies, on each combination of features (see App. \ref{app:MEIRL_demos}). Importantly, while \ac{ME-IRL} uses a set of near-optimal demonstrations $\mathcal{D}^*$, \ac{FERL} requires both demonstrations and feature traces $\trace$. To level the amount of data each method has access to, we distributed the demonstrations and traces \ac{FERL} has access to such that \ac{ME-IRL} has more data points. The average number of states per demonstration/trace were 64 and 31, respectively, so if we keep the number of \ac{ME-IRL} demonstrations and \ac{FERL} traces the same, \ac{FERL} has a non-zero budget of demonstrations to use for cases with more than one demonstration ($N>1$).

\paragraph{Dependent Measures.} We use the same objective metrics as \textit{Reward Accuracy}, \textit{Behavior Accuracy}, \change{and \textit{Test Probability}} in Sec. \ref{sec:online_FERL_expert_exps}. For \textit{Reward Accuracy}, we vary the number $N$ of traces / demonstrations each learner gets but skip $N=1$ because \ac{FERL} would have an unfair advantage in the amount of data given.
We give \ac{ME-IRL} up to 10, 20, and 30 demonstrations for the three tasks, respectively. Meanwhile, we give \ac{FERL} up to 10 traces for each feature, and 1, 2, and 3 demonstrations for each task, respectively. Overall, \ac{FERL} would use up to 10 traces and one demonstration, up to 20 traces and 2 demonstrations, and up to 30 traces and 3 demonstrations, while \ac{ME-IRL} would be given 10, 20, and 30 demonstrations for each task, respectively.
For \textit{Behavior Accuracy} \change{and \textit{Test Probability}}, we train \ac{FERL} with 10 traces per feature and 1, 2, or 3 demonstrations, and \ac{ME-IRL} with 10, 20, and 30 demonstrations, respectively. Just like in Sec. \ref{sec:online_FERL_expert_exps}, for \textit{Behavior Accuracy} we produce optimal trajectories for 100 randomly selected start-goal pairs under the learned rewards and evaluate them under the \ac{GT} reward.
\change{Meanwhile, for \textit{Test Probability}, we generate 100 optimal trajectories using the \ac{GT} reward, then evaluate their likelihood under the learned models.}

\paragraph{Hypothesis.} \hfill

\noindent\textbf{H9:} Offline \ac{FERL} learns rewards that better generalize to the state space than \ac{ME-IRL}.

\subsubsection{Qualitative Comparison.}

In Fig. \ref{fig:OfflineVisualComparison}, we show the learned \ac{FERL} and \ac{ME-IRL} rewards as well as the \ac{GT} for all three tasks evaluated at the test points.
The figure illustrates that by first learning each feature separately and then the reward that combines them, \ac{FERL} is able to learn a fine-grained reward structure closely resembling the \ac{GT}. 
For the easiest task, \textit{One Feature}, \ac{ME-IRL} does recover the \ac{GT} appearance, but this is unsurprising since the \textit{table} feature is very simple.
For the other more complex two tasks, just like in the online case, \ac{ME-IRL} learns some structure capturing where the laptop or the human is, but not enough to result in a good trade-off between the features.

\subsubsection{Quantitative Analysis.}

To compare \textit{Reward Accuracy}, we show in Fig. \ref{fig:offline_FERL_Cost_MSE} the \ac{MSE} mean and standard error across 10 seeds, with increasing training data. We visualize results from all 3 tasks side by side, with \ac{FERL} in orange and \ac{ME-IRL} in gray. For \textit{One Feature}, as expected, \ac{ME-IRL} does eventually learn a good reward with enough data. However, for the other more complex tasks that combine multiple features, \ac{ME-IRL} underperforms when compared to our method.
Overall, across the tasks, \ac{FERL} is closer to GT than \ac{ME-IRL} no matter the amount of data, supporting H9. To test this, we ran an ANOVA with learning method as the factor, and with the task and data amount as covariates, and found a significant main effect (F(1, 535) = 148.8431, p < .0001).

\begin{figure}

  \centering
  \includegraphics[width=0.44\textwidth]{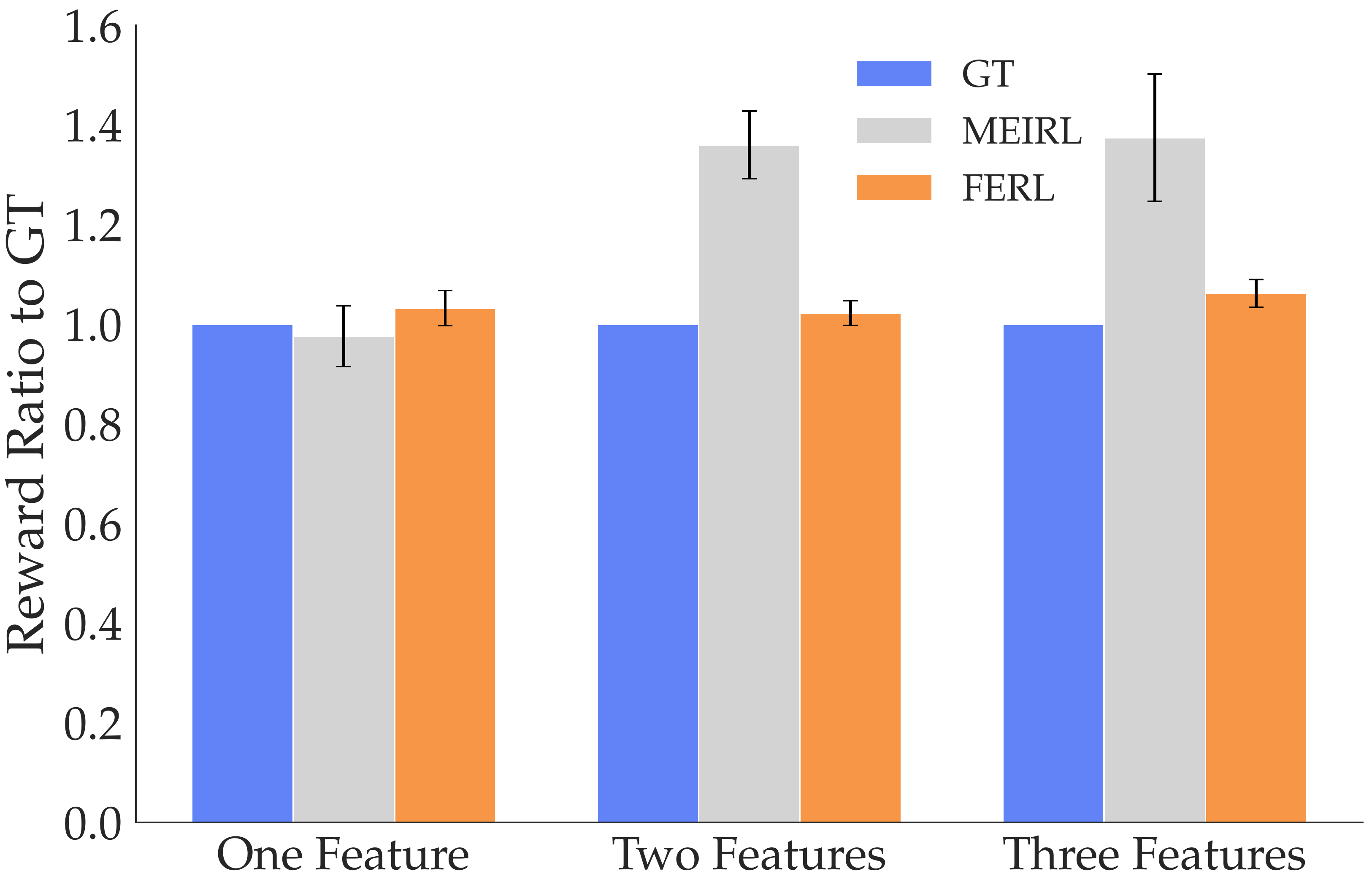}
  \caption{Induced trajectories' reward ratio for the two methods compared to \ac{GT}. While \ac{ME-IRL} generalizes for the single feature task, it struggles with the more complex multiple feature tasks.}
  \label{fig:offline_InducedTrajectories}

\end{figure}

\begin{figure}

   \centering
  \includegraphics[width=0.44\textwidth]{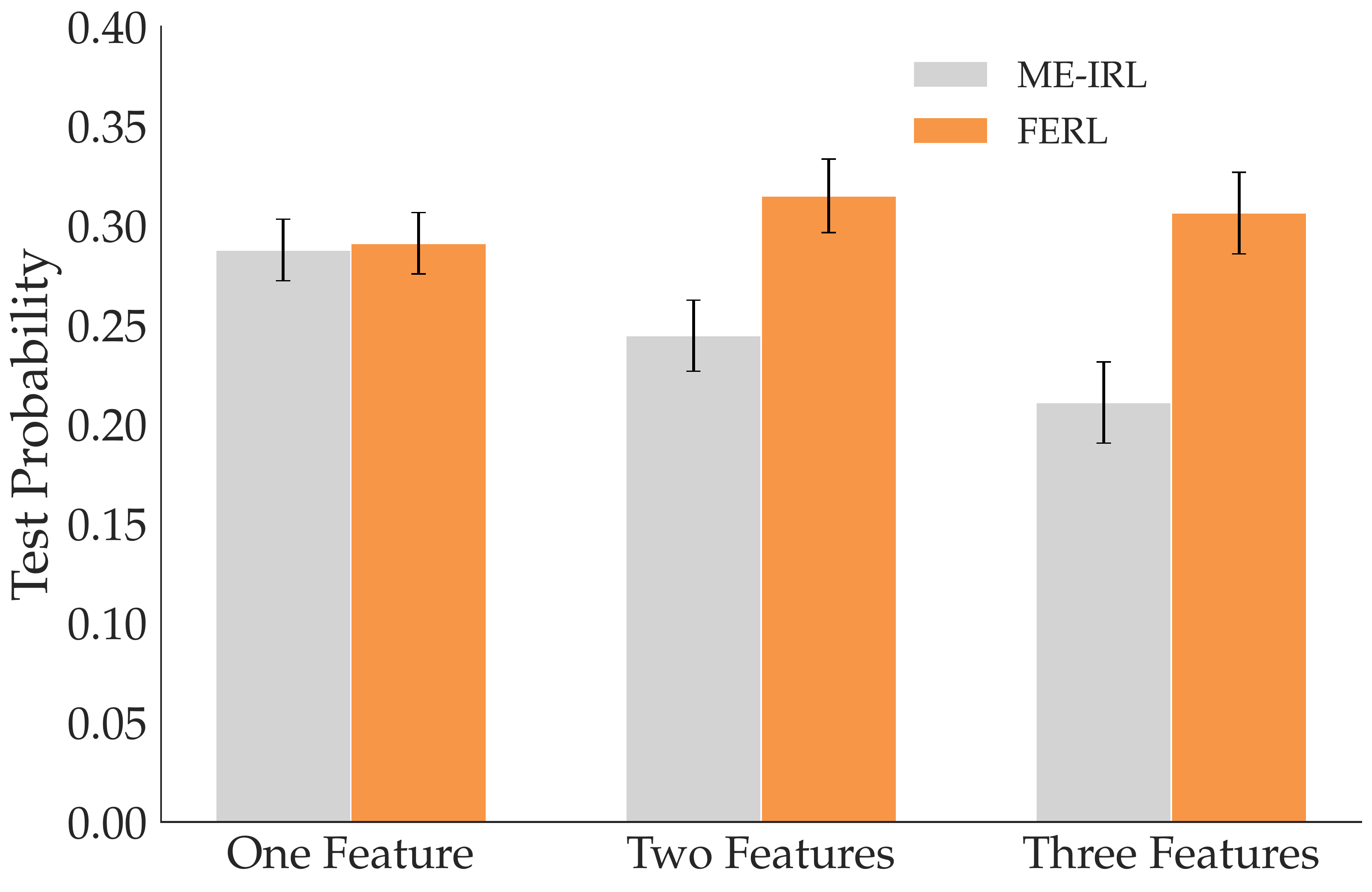}
   \caption{\change{Probability assigned by the two methods to a set of optimal trajectories under the Boltzmann assumption. For the more complex multiple feature tasks, the trajectories are more likely under \ac{FERL} than \ac{ME-IRL}.}}
   \label{fig:offlineFERL_Probabilities}

\end{figure}

For comparing \textit{Behavior Accuracy}, Fig. \ref{fig:offline_InducedTrajectories} illustrates the reward ratios to \ac{GT} for all three tasks. When the reward consists of a single very simple feature, \ac{ME-IRL} performs just as well as our method. However, when the reward structure more complexly combines multiple features, \ac{ME-IRL} does not produce as good trajectories under the \ac{GT} reward as \ac{FERL}, supporting H8. 
We ran an ANOVA using the learning method as a factor and the task as a covariate and did not find a significant main effect, probably due to the \textit{One Feature} results. To verify this theory, we re-ran the ANOVA using only the data from the more complex \textit{Two Features} and \textit{Three Features} tasks, and did, in fact, find a significant main effect (F(1, 397) = 5.7489, p = .0097).
\change{Results with the \textit{Test Probability} metric paint a similar picture. Fig. \ref{fig:offlineFERL_Probabilities} shows that for the easy \textit{One Feature} case, both methods perform comparably, but when the reward is more complex (\textit{Two Features} and \textit{Three Features}), \ac{FERL} outperforms \ac{ME-IRL} and assigns higher probability to the test trajectories.}

\subsubsection{Summary.} The results in this section suggest that while \ac{ME-IRL} is capable of recovering very simple reward structures, it does not perform as well as using \ac{FERL} features for complex rewards. This observation applies when the features are taught by experts, so we now test what happens if we instead use non-expert user features.

\subsection{Non-expert Users}
\label{sec:offline_FERL_user_exps}

In Sec. \ref{sec:online_FERL_user_exps}, we saw that user-taught \ac{FERL} features have enough structure to help the robot recover the human's preferences in online reward setting where the original feature set is incomplete. However, there we only had one missing feature. In this section, we test the more challenging scenario, where we learn a reward from scratch using the noisy user features learned in simulation.

\subsubsection{Experimental Design.}

For this experiment, we had a similar setup as in Sec. \ref{sec:online_FERL_user_exps} -- using the user-taught simulation features for learning the reward -- only this time we tested the offline instantiation of \ac{FERL}.
Given that now we combine multiple noisy features together into a reward, we wanted to see how our divide-and-conquer approach fares against the \ac{ME-IRL} baseline.

\paragraph{Manipulated Variables.}
We manipulate the \textit{learning method}, \ac{FERL} or \ac{ME-IRL}, just like in Sec. \ref{sec:offline_FERL_expert_exps}. Like in Sec. \ref{sec:online_FERL_user_exps}, we use demonstrations collected from the expert on the physical robot for \ac{ME-IRL}. For \ac{FERL}, we use the user data from the simulation, and the expert demonstrations that teach the robot how to combine the learned feature into a reward. Note that this gives \ac{ME-IRL} an advantage, since all its data is both generated by an expert, and on the physical robot.

\paragraph{Dependent Measures.} We use the same objective metric as \textit{Reward Accuracy} in the expert comparison in Sec. \ref{sec:offline_FERL_expert_exps}: the learned reward \ac{MSE} to the \ac{GT} reward on $\mathcal{S}_{\text{Test}}$.

\paragraph{Hypotheses.} \hfill

\noindent\textbf{H10:} Offline FERL learns more generalizable rewards than ME-IRL even when using features learned from data provided by non-experts in simulation.

\subsubsection{Analysis.}

\begin{figure}
  \centering
  \includegraphics[width=0.48\textwidth]{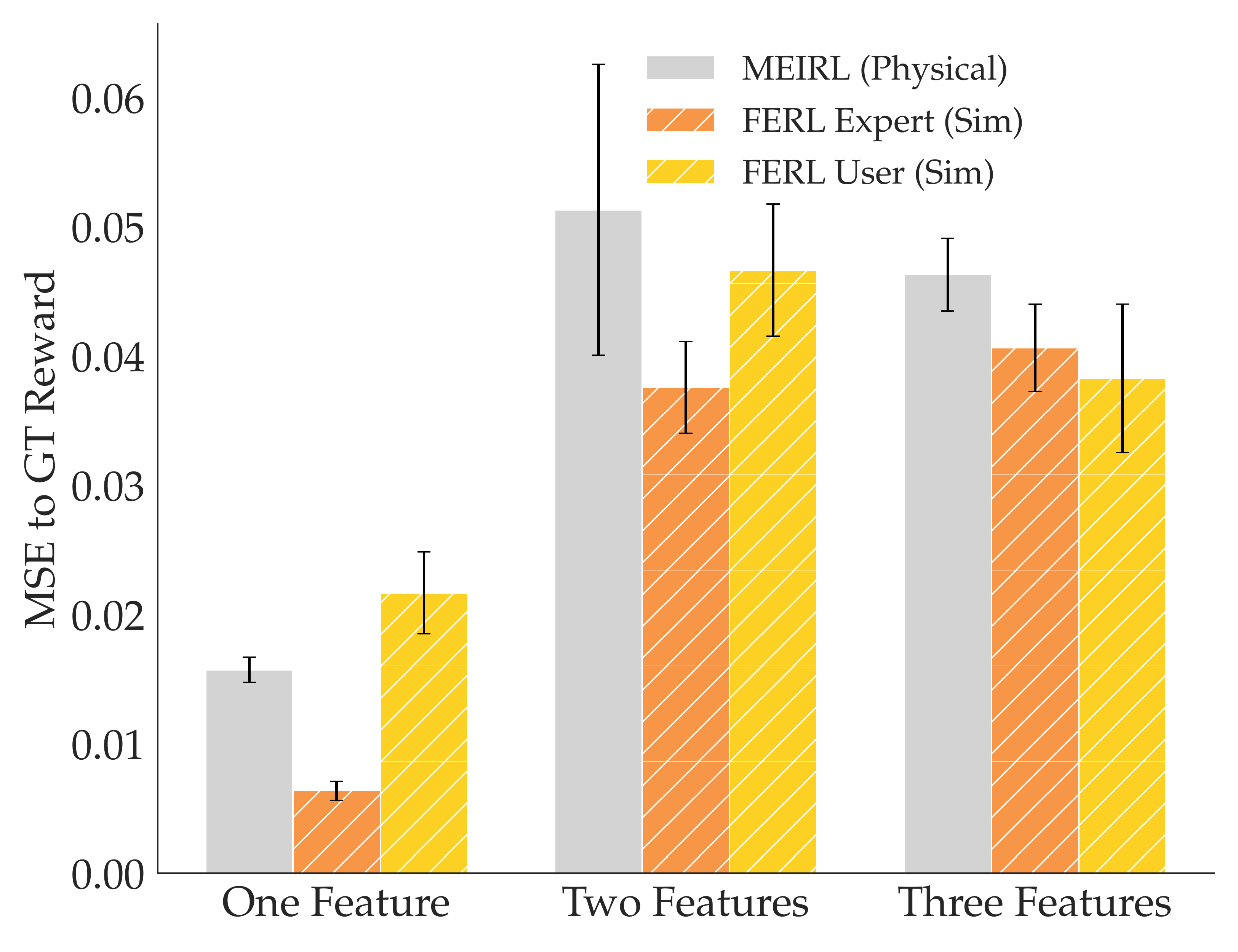}
   \caption{MSE to GT reward for the three tasks, comparing \ac{ME-IRL} from expert physical demonstrations (gray) to offline \ac{FERL} from expert (orange) and non-expert (yellow) features learned in simulation and combined via corrections.}
   \label{fig:study_offline_rewardMSE}
\end{figure}

Fig. \ref{fig:study_offline_rewardMSE} illustrates our findings for the reward comparison. We also added the offline \ac{FERL} reward using expert-taught simulation features for reference, where we \change{randomly} subsampled sets of 10 traces and trained 12 expert features for each of the three features.
This time, we find that the user features are noisy enough that, when combined into a reward, they do not reliably provide an advantage over \ac{ME-IRL}. This could be attributed to the difficulty of teaching features in a simulator, especially given that there is no easy way to approximate distances and traces in 3D space with a 2D interface are hard. 
We verified this result with an ANOVA with the learning method as a factor and the task as a covariate, and, as expected, we found no significant main effect.

\subsubsection{Summary.} 
Previously, we have seen how structure can indeed help reward learning generalizability and sample efficiency; but we now see that the \emph{wrong} -- or very noisy -- structure obtained from traces from simulation may diminish the benefits that our divide-and-conquer approach promises. However, we suggest taking this result with a grain of salt, since \ac{ME-IRL} had the advantage of all-expert, all-physical data, whereas our method was limited to data collected in simulation from novice users. While not possible during the pandemic, we are optimistic that with physical demonstrations the benefits would be more prominent. 

\section{Discussion}
\label{sec:discussion}

Learning reward functions is a popular way to help robots generate behaviors that adapt to new situations or human preferences.
In this work, we propose that robots can learn more generalizable rewards by using a divide-and-conquer approach, focusing on learning features separately from learning how to combine them.
We introduced feature traces as a novel type of human input that allows for intuitive teaching of non-linear features from high-dimensional state spaces. 
We then presented two instantiations of our \ac{FERL} algorithm: one that enables expanding the robot's feature set in online reward learning situations, and one that lets the user sequentially teach every feature and then combine them into a reward.
In extensive experiments with a real robot arm and a user study in simulation, we showed that online \ac{FERL} outperforms deep reward learning from demonstrations (\ac{ME-IRL}) in data-efficiency and generalization. Offline \ac{FERL} similarly beats \ac{ME-IRL} when the features used are of high enough quality, but the results are less conclusive when using very noisy features.

\paragraph{Implications for Online Reward Learning.}

Because they have to perform updates in real time from very little
input, online reward learning methods represent the reward as a linear function of a small set of hand-engineered features. 
As discussed, exhaustively choosing such a set a priori puts too much burden on system designers, and using an incomplete set of features can lead to learning the wrong reward.
Prior work enabled robots to at least detect that its feature space is insufficient to explain the human's input \citep{bobu2018learning}, but then the robot's only option was to either not update the reward or completely stop task execution.
Our online \ac{FERL} approach provides an alternative that allows people to teach features when the robot detects it is missing something, and then update the reward using the new feature set. 
Although in this paper we presented experiments where the robot learns rewards from corrections, our framework can conceivably be adapted to any online reward learning method, provided there is a way to detect the feature set is insufficient. Recent work on confidence estimation from human demonstrations \citep{bobu2020quantifying} and teleoperation \citep{zurek2021situational} offers encouraging pathways to adapting \ac{FERL} to other online human-robot collaborative settings.

\paragraph{Implications for Learning Complex Rewards from Demonstrations.}

Reward learning from raw state space with expressive function approximators is considered difficult because there exists a large set of functions $r_\theta(\sysstate)$ that could explain the human input. For example, in the case of learning from demonstrations, many functions $r_\theta(\sysstate)$ induce policies that match the demonstrations' state expectation. 
The higher dimensional the state $\sysstate$, the more human input is needed to disambiguate between those functions sufficiently to find a reward $r_\theta$ that accurately captures human preferences. Without that, the learned reward is unlikely to generalize to states not seen during training and might simply replicate the demonstrations' state expectations. In this paper, we presented evidence that offline \ac{FERL} may provide an alternative to better disambiguate the reward and improve generalization. 

The reason our divide-and-conquer approach can help relative to relying on demonstrations for everything is that demonstrations aggregate a lot of information. First, by learning features, we can isolate learning what matters from learning how to trade off what matters into a single value (the features vs. their combination) -- in contrast, demonstrations have to teach the robot about both at once. Second, feature traces give information about states that are not on optimal trajectories, be it states with high feature values that are undesirable, or states with low feature values where other, more important features have high values.
Third, feature traces are also structured by the monotonicity assumption: they tell us relative feature values of the states along a trace, whereas demonstrations only tell us about the aggregate reward across a trajectory. 
Thus, by focusing on learning features first before combining them into a reward, the robot can incorporate all three benefits and ultimately improve reward learning from demonstrations.


\paragraph{Limitations and Future Work.}

Our work is merely a step towards understanding how explicitly focusing on learning features can impact reward learning generalization and sample complexity. While \ac{FERL} enables robots to learn features and induce structure in reward learning, there are several limitations that may affect its usability.

Our user study provides evidence that non-expert users can, in fact, use \ac{FERL} to teach good features. However, due to the current pandemic, we conducted the study in a simulated environment instead of in person with the real robot, and most of our users had technical background. It is unclear how people without technical background would perform, and especially how kinesthetically providing feature traces (instead of clicking and dragging in a simulator) would affect their perception of the protocol's usability. 
Further, we only tested whether users could teach features we tell them about, so we still need to test whether users can teach features they implicitly know about (as would happen when intervening to correct the robot or designing a reward from scratch).

Even if people know the feature they want to teach, it might be so abstract (e.g. comfort) that they would not know how to teach it.
Moreover, with the current feature learning protocol, they might find it cumbersome to teach discontinuous features like constraints.
We could ease the human supervision burden by developing an active learning approach where the robot autonomously picks starting states most likely to result in informative feature traces. 
For instance, the robot could fit an ensemble of functions from traces online, and query for new traces from states where the ensemble disagrees \citep{reddy2019learning}. 
But for such complex features, it may be more effective to investigate combining feature traces with other types of structured human input.

The quality of the learned rewards depends directly on the quality of the learned features. When the human provides feature traces that lead to good features, many of our experiments demonstrate that they induce structure in the reward learning procedure that helps generalization and sample complexity. However, if the robot learns features that are too noisy or simply incorrect, that (wrong) structure may impair performance. 
We saw an example of this when we tried to utilize the user study features for reward learning. In online \ac{FERL} where a single feature was missing, the structure captured by the (noisy) non-expert features was still helpful in learning a better reward than the baseline. However, when trying to combine multiple noisy features in offline \ac{FERL}, reward learning did not see a benefit.
Future work must investigate ways in which the robot can determine whether to accept or reject the newly learned feature. 
One idea is to use our current framework's confidence estimation capability in Sec. \ref{sec:confidence_estimation} to determine whether the learned feature set explains the human's reward input. Another idea is to visualize either the feature function or examples of behaviors induced by it, and let the person decide whether the learned feature is acceptable.

Lastly, while we show that \ac{FERL} works reliably in 27D, more work is necessary to extend it to higher dimensional state spaces, like images. In our discussion in App. \ref{app:rawstate}, we show how spurious correlations in large input spaces may affect the quality of the learned features in low data regimes. To counteract that, we could ask the person for more data, but after a certain point this becomes too burdensome on the user.
Alternatively, approaches that encode these spaces to lower dimensional representations or techniques from causal learning, such as Invariant Risk Minimization~\citep{Arjovsky2019InvariantRM}, could help tackle these challenges.



\begin{acks}
This research is supported by the Air Force Office of Scientific Research (AFOSR), the Office of Naval Research (ONR-YIP), the DARPA Assured Autonomy Grant, the CONIX Research Center, and the German Academic Exchange Service (DAAD).
We also thank Rohin Shah for providing guidance and feedback on our work.
\end{acks}

\bibliographystyle{SageH}
\bibliography{IJRR2022}

\begin{thebibliography}{49}
\providecommand{\natexlab}[1]{#1}
\providecommand{\url}[1]{\texttt{#1}}
\providecommand{\urlprefix}{URL }
\expandafter\ifx\csname urlstyle\endcsname\relax
  \providecommand{\doi}[1]{DOI:\discretionary{}{}{}#1}\else
  \providecommand{\doi}{DOI:\discretionary{}{}{}\begingroup
  \urlstyle{rm}\Url}\fi

\bibitem[{Abbeel and Ng(2004)}]{abbeel2004apprenticeship}
Abbeel P and Ng AY (2004) Apprenticeship learning via inverse reinforcement
  learning.
\newblock In: \emph{Machine Learning (ICML), International Conference on}. ACM.

\bibitem[{Amodei and Clark(2016)}]{amodei2016}
Amodei D and Clark J (2016) Faulty reward functions in the wild
  \urlprefix\url{https://blog.openai. com/faulty-reward-functions/}.

\bibitem[{Argall et~al.(2009)Argall, Chernova, Veloso and
  Browning}]{argall2009survey}
Argall BD, Chernova S, Veloso M and Browning B (2009) A survey of robot
  learning from demonstration.
\newblock \emph{Robotics and autonomous systems} 57(5): 469--483.

\bibitem[{Arjovsky et~al.(2019)Arjovsky, Bottou, Gulrajani and
  Lopez-Paz}]{Arjovsky2019InvariantRM}
Arjovsky M, Bottou L, Gulrajani I and Lopez-Paz D (2019) Invariant risk
  minimization.
\newblock \emph{ArXiv} abs/1907.02893.

\bibitem[{Bajcsy et~al.(2018)Bajcsy, Losey, O'Malley and
  Dragan}]{bajcsy2018onefeature}
Bajcsy A, Losey DP, O'Malley MK and Dragan AD (2018) Learning from physical
  human corrections, one feature at a time.
\newblock In: \emph{Proceedings of the 2018 ACM/IEEE International Conference
  on Human-Robot Interaction}, HRI '18. New York, NY, USA: ACM.
\newblock ISBN 978-1-4503-4953-6, pp. 141--149.
\newblock \doi{10.1145/3171221.3171267}.
\newblock \urlprefix\url{http://doi.acm.org/10.1145/3171221.3171267}.

\bibitem[{Bajcsy et~al.(2017)Bajcsy, Losey, O’Malley and
  Dragan}]{bajcsy2017phri}
Bajcsy A, Losey DP, O’Malley MK and Dragan AD (2017) Learning robot
  objectives from physical human interaction.
\newblock In: Levine S, Vanhoucke V and Goldberg K (eds.) \emph{Proceedings of
  the 1st Annual Conference on Robot Learning}, \emph{Proceedings of Machine
  Learning Research}, volume~78. PMLR, pp. 217--226.
\newblock \urlprefix\url{http://proceedings.mlr.press/v78/bajcsy17a.html}.

\bibitem[{Baker et~al.(2007)Baker, B~Tenenbaum and R~Saxe}]{baker2007goal}
Baker C, B~Tenenbaum J and R~Saxe R (2007) Goal inference as inverse planning.
\newblock In: \emph{Proceedings of the 29th Annual Conference of the Cognitive
  Science Society}.

\bibitem[{{Bobu} et~al.(2020){Bobu}, {Bajcsy}, {Fisac}, {Deglurkar} and
  {Dragan}}]{bobu2020quantifying}
{Bobu} A, {Bajcsy} A, {Fisac} JF, {Deglurkar} S and {Dragan} AD (2020)
  Quantifying hypothesis space misspecification in learning from human–robot
  demonstrations and physical corrections.
\newblock \emph{IEEE Transactions on Robotics} : 1--20.

\bibitem[{Bobu et~al.(2018)Bobu, Bajcsy, Fisac and Dragan}]{bobu2018learning}
Bobu A, Bajcsy A, Fisac JF and Dragan AD (2018) Learning under misspecified
  objective spaces.
\newblock In: Billard A, Dragan A, Peters J and Morimoto J (eds.)
  \emph{Proceedings of The 2nd Conference on Robot Learning}, \emph{Proceedings
  of Machine Learning Research}, volume~87. PMLR, pp. 796--805.
\newblock \urlprefix\url{http://proceedings.mlr.press/v87/bobu18a.html}.

\bibitem[{Bobu et~al.(2021)Bobu, Wiggert, Tomlin and Dragan}]{bobu2021ferl}
Bobu A, Wiggert M, Tomlin C and Dragan AD (2021) Feature expansive reward
  learning: Rethinking human input.
\newblock In: \emph{Proceedings of the 2021 ACM/IEEE International Conference
  on Human-Robot Interaction}, HRI '21. New York, NY, USA: Association for
  Computing Machinery.
\newblock ISBN 9781450382892, p. 216–224.
\newblock \doi{10.1145/3434073.3444667}.
\newblock \urlprefix\url{https://doi.org/10.1145/3434073.3444667}.

\bibitem[{Braziunas and Boutilier(2008)}]{Braziunas2008elicitation}
Braziunas D and Boutilier C (2008) Elicitation of factored utilities.
\newblock \emph{AI Magazine} 29(4): 79.
\newblock \doi{10.1609/aimag.v29i4.2203}.
\newblock
  \urlprefix\url{https://ojs.aaai.org/index.php/aimagazine/article/view/2203}.

\bibitem[{Brown et~al.(2020)Brown, Coleman, Srinivasan and
  Niekum}]{brown2020brex}
Brown D, Coleman R, Srinivasan R and Niekum S (2020) Safe imitation learning
  via fast {B}ayesian reward inference from preferences.
\newblock In: III HD and Singh A (eds.) \emph{Proceedings of the 37th
  International Conference on Machine Learning}, \emph{Proceedings of Machine
  Learning Research}, volume 119. PMLR, pp. 1165--1177.
\newblock \urlprefix\url{http://proceedings.mlr.press/v119/brown20a.html}.

\bibitem[{Brown et~al.(2019)Brown, Goo, Nagarajan and
  Niekum}]{brown2019extrapolating}
Brown D, Goo W, Nagarajan P and Niekum S (2019) Extrapolating beyond suboptimal
  demonstrations via inverse reinforcement learning from observations.
\newblock In: \emph{International Conference on Machine Learning}. PMLR, pp.
  783--792.

\bibitem[{Brown et~al.(2018)Brown, Cui and Niekum}]{brown2018risk}
Brown DS, Cui Y and Niekum S (2018) Risk-aware active inverse reinforcement
  learning.
\newblock In: \emph{Conference on Robot Learning}. PMLR, pp. 362--372.

\bibitem[{Choi and Kim(2011)}]{choi2011inverse}
Choi J and Kim KE (2011) {Inverse reinforcement learning in partially
  observable environments}.
\newblock \emph{Journal of Machine Learning Research} 12(Mar): 691--730.

\bibitem[{Choi and Kim(2013)}]{choi2013bayesian}
Choi J and Kim KE (2013) {Bayesian nonparametric feature construction for
  inverse reinforcement learning}.
\newblock In: \emph{Twenty-Third International Joint Conference on Artificial
  Intelligence}.

\bibitem[{Christiano et~al.(2017)Christiano, Leike, Brown, Martic, Legg and
  Amodei}]{christiano2017preferences}
Christiano PF, Leike J, Brown T, Martic M, Legg S and Amodei D (2017) Deep
  reinforcement learning from human preferences.
\newblock In: Guyon I, Luxburg UV, Bengio S, Wallach H, Fergus R, Vishwanathan
  S and Garnett R (eds.) \emph{Advances in Neural Information Processing
  Systems}, volume~30. Curran Associates, Inc.
\newblock
  \urlprefix\url{https://proceedings.neurips.cc/paper/2017/file/d5e2c0adad503c91f91df240d0cd4e49-Paper.pdf}.

\bibitem[{Coumans and Bai(2016--2019)}]{coumans2019}
Coumans E and Bai Y (2016--2019) Pybullet, a python module for physics
  simulation for games, robotics and machine learning.
\newblock \url{http://pybullet.org}.

\bibitem[{Dragan et~al.(2015)Dragan, Muelling, Bagnell and
  Srinivasa}]{deformation}
Dragan AD, Muelling K, Bagnell JA and Srinivasa SS (2015) Movement primitives
  via optimization.
\newblock In: \emph{2015 IEEE International Conference on Robotics and
  Automation (ICRA)}. pp. 2339--2346.
\newblock \doi{10.1109/ICRA.2015.7139510}.

\bibitem[{Finn et~al.(2016)Finn, Levine and Abbeel}]{finn2016gcl}
Finn C, Levine S and Abbeel P (2016) Guided cost learning: Deep inverse optimal
  control via policy optimization.
\newblock In: \emph{Proceedings of the 33rd International Conference on
  International Conference on Machine Learning - Volume 48}, ICML’16.
  JMLR.org, p. 49–58.

\bibitem[{Fisac et~al.(2018)Fisac, Bajcsy, Herbert, Fridovich-Keil, Wang,
  Tomlin and Dragan}]{fisac2018probabilistically}
Fisac JF, Bajcsy A, Herbert SL, Fridovich-Keil D, Wang S, Tomlin CJ and Dragan
  AD (2018) Probabilistically safe robot planning with confidence-based human
  predictions.
\newblock \emph{Robotics: Science and Systems (RSS)} .

\bibitem[{Fridovich-Keil et~al.(2019)Fridovich-Keil, Bajcsy, Fisac, Herbert,
  Wang, Dragan and Tomlin}]{fridovich-keil2019confidence}
Fridovich-Keil D, Bajcsy A, Fisac JF, Herbert SL, Wang S, Dragan AD and Tomlin
  CJ (2019) Confidence-aware motion prediction for real-time collision
  avoidance.
\newblock \emph{International Journal of Robotics Research} .

\bibitem[{Fu et~al.(2018{\natexlab{a}})Fu, Luo and Levine}]{fu2018learning}
Fu J, Luo K and Levine S (2018{\natexlab{a}}) Learning robust rewards with
  adverserial inverse reinforcement learning.
\newblock In: \emph{International Conference on Learning Representations}.
\newblock \urlprefix\url{https://openreview.net/forum?id=rkHywl-A-}.

\bibitem[{Fu et~al.(2018{\natexlab{b}})Fu, Singh, Ghosh, Yang and
  Levine}]{fu2018variational}
Fu J, Singh A, Ghosh D, Yang L and Levine S (2018{\natexlab{b}}) Variational
  inverse control with events: A general framework for data-driven reward
  definition.
\newblock \emph{arXiv preprint} arXiv:1805.11686.

\bibitem[{Hadfield-Menell et~al.(2017)Hadfield-Menell, Milli, Abbeel, Russell
  and Dragan}]{HadfieldMenell2017InverseRD}
Hadfield-Menell D, Milli S, Abbeel P, Russell SJ and Dragan A (2017) Inverse
  reward design.
\newblock In: Guyon I, Luxburg UV, Bengio S, Wallach H, Fergus R, Vishwanathan
  S and Garnett R (eds.) \emph{Advances in Neural Information Processing
  Systems}, volume~30. Curran Associates, Inc.
\newblock
  \urlprefix\url{https://proceedings.neurips.cc/paper/2017/file/32fdab6559cdfa4f167f8c31b9199643-Paper.pdf}.

\bibitem[{Hart and Staveland(1988)}]{HART1988NASATLX}
Hart SG and Staveland LE (1988) Development of nasa-tlx (task load index):
  Results of empirical and theoretical research.
\newblock In: Hancock PA and Meshkati N (eds.) \emph{Human Mental Workload},
  \emph{Advances in Psychology}, volume~52. North-Holland, pp. 139 -- 183.
\newblock \doi{https://doi.org/10.1016/S0166-4115(08)62386-9}.
\newblock
  \urlprefix\url{http://www.sciencedirect.com/science/article/pii/S0166411508623869}.

\bibitem[{Haug et~al.(2018)Haug, Tschiatschek and Singla}]{haug2018teaching}
Haug L, Tschiatschek S and Singla A (2018) Teaching inverse reinforcement
  learners via features and demonstrations.
\newblock In: \emph{Advances in Neural Information Processing Systems}. pp.
  8464--8473.

\bibitem[{Ibarz et~al.(2018)Ibarz, Leike, Pohlen, Irving, Legg and
  Amodei}]{Ibarz2018reward}
Ibarz B, Leike J, Pohlen T, Irving G, Legg S and Amodei D (2018) Reward
  learning from human preferences and demonstrations in atari.
\newblock In: Bengio S, Wallach H, Larochelle H, Grauman K, Cesa-Bianchi N and
  Garnett R (eds.) \emph{Advances in Neural Information Processing Systems},
  volume~31. Curran Associates, Inc., pp. 8011--8023.
\newblock
  \urlprefix\url{https://proceedings.neurips.cc/paper/2018/file/8cbe9ce23f42628c98f80fa0fac8b19a-Paper.pdf}.

\bibitem[{Jain et~al.(2015)Jain, Sharma, Joachims and
  Saxena}]{jain2015learning}
Jain A, Sharma S, Joachims T and Saxena A (2015) Learning preferences for
  manipulation tasks from online coactive feedback.
\newblock \emph{The International Journal of Robotics Research} 34(10):
  1296--1313.

\bibitem[{Javdani et~al.(2018)Javdani, Admoni, Pellegrinelli, Srinivasa and
  Bagnell}]{javdani2015shared}
Javdani S, Admoni H, Pellegrinelli S, Srinivasa SS and Bagnell JA (2018) Shared
  autonomy via hindsight optimization for teleoperation and teaming.
\newblock \emph{The International Journal of Robotics Research} 37(7):
  717--742.
\newblock \doi{10.1177/0278364918776060}.
\newblock \urlprefix\url{https://doi.org/10.1177/0278364918776060}.

\bibitem[{Jaynes(1957)}]{jaynes1957infotheory}
Jaynes ET (1957) Information theory and statistical mechanics.
\newblock American Physical Society, pp. 620--630.
\newblock \doi{10.1103/PhysRev.106.620}.
\newblock \urlprefix\url{https://link.aps.org/doi/10.1103/PhysRev.106.620}.

\bibitem[{Levine et~al.(2010)Levine, Popovic and Koltun}]{levine2010feature}
Levine S, Popovic Z and Koltun V (2010) {Feature construction for inverse
  reinforcement learning}.
\newblock In: \emph{Advances in Neural Information Processing Systems}. pp.
  1342--1350.

\bibitem[{Levine et~al.(2011)Levine, Popovic and Koltun}]{levine2011nonlinear}
Levine S, Popovic Z and Koltun V (2011) {Nonlinear inverse reinforcement
  learning with gaussian processes}.
\newblock In: \emph{Advances in Neural Information Processing Systems}. pp.
  19--27.

\bibitem[{Lopes et~al.(2009)Lopes, Melo and Montesano}]{lopes2009active}
Lopes M, Melo F and Montesano L (2009) Active learning for reward estimation in
  inverse reinforcement learning.
\newblock In: \emph{Joint European Conference on Machine Learning and Knowledge
  Discovery in Databases}. Springer, pp. 31--46.

\bibitem[{Ng and Russell(2000)}]{Ng2000inverse}
Ng A and Russell S (2000) {Algorithms for inverse reinforcement learning}.
\newblock \emph{International Conference on Machine Learning (ICML)} 0:
  663--670.
\newblock \doi{10.2460/ajvr.67.2.323}.
\newblock
  \urlprefix\url{http://www-cs.stanford.edu/people/ang/papers/icml00-irl.pdf}.

\bibitem[{Osa et~al.(2018)Osa, Pajarinen, Neumann, Bagnell, Abbeel, Peters
  et~al.}]{osa2018algorithmic}
Osa T, Pajarinen J, Neumann G, Bagnell JA, Abbeel P, Peters J et~al. (2018) An
  algorithmic perspective on imitation learning.
\newblock \emph{Foundations and Trends in Robotics} 7(1-2): 1--179.

\bibitem[{Ratliff et~al.(2007)Ratliff, Bradley, Chestnutt and
  Bagnell}]{ratliff2007boosting}
Ratliff N, Bradley DM, Chestnutt J and Bagnell JA (2007) {Boosting structured
  prediction for imitation learning}.
\newblock In: \emph{Advances in Neural Information Processing Systems}. pp.
  1153--1160.

\bibitem[{Ratliff et~al.(2006)Ratliff, Bagnell and Zinkevich}]{ratliff2006MMP}
Ratliff ND, Bagnell JA and Zinkevich MA (2006) Maximum margin planning.
\newblock In: \emph{Proceedings of the 23rd International Conference on Machine
  Learning}, ICML ’06. New York, NY, USA: Association for Computing
  Machinery.
\newblock ISBN 1595933832, p. 729–736.
\newblock \doi{10.1145/1143844.1143936}.
\newblock \urlprefix\url{https://doi.org/10.1145/1143844.1143936}.

\bibitem[{Reddy et~al.(2020{\natexlab{a}})Reddy, Dragan, Levine, Legg and
  Leike}]{reddy2019learning}
Reddy S, Dragan A, Levine S, Legg S and Leike J (2020{\natexlab{a}}) Learning
  human objectives by evaluating hypothetical behavior.
\newblock In: \emph{ICML}.

\bibitem[{Reddy et~al.(2020{\natexlab{b}})Reddy, Dragan and
  Levine}]{Reddy2020SQILIL}
Reddy S, Dragan AD and Levine S (2020{\natexlab{b}}) {SQIL:} imitation learning
  via reinforcement learning with sparse rewards.
\newblock In: \emph{8th International Conference on Learning Representations,
  {ICLR} 2020, Addis Ababa, Ethiopia, April 26-30, 2020}. OpenReview.net.
\newblock \urlprefix\url{https://openreview.net/forum?id=S1xKd24twB}.

\bibitem[{Russell and Norvig(2002)}]{russell2002artificial}
Russell S and Norvig P (2002) Artificial intelligence: a modern approach .

\bibitem[{{Sadigh} et~al.(2016){Sadigh}, {Sastry}, {Seshia} and
  {Dragan}}]{sadigh2016infogather}
{Sadigh} D, {Sastry} SS, {Seshia} SA and {Dragan} A (2016) Information
  gathering actions over human internal state.
\newblock In: \emph{2016 IEEE/RSJ International Conference on Intelligent
  Robots and Systems (IROS)}. pp. 66--73.
\newblock \doi{10.1109/IROS.2016.7759036}.

\bibitem[{Schulman et~al.(2013)Schulman, Ho, Lee, Awwal, Bradlow and
  Abbeel}]{schulman2013trajopt}
Schulman J, Ho J, Lee AX, Awwal I, Bradlow H and Abbeel P (2013) Finding
  locally optimal, collision-free trajectories with sequential convex
  optimization.
\newblock In: \emph{Robotics: science and systems}, volume~9. Citeseer, pp.
  1--10.

\bibitem[{Vapnik(2013)}]{vapnik2013nature}
Vapnik V (2013) \emph{The nature of statistical learning theory}.
\newblock Springer science \& business media.

\bibitem[{Vernaza and Bagnell(2012)}]{vernaza2012efficient}
Vernaza P and Bagnell D (2012) {Efficient high dimensional maximum entropy
  modeling via symmetric partition functions}.
\newblock In: \emph{Advances in Neural Information Processing Systems}. pp.
  575--583.

\bibitem[{Von~Neumann and Morgenstern(1945)}]{von1945theory}
Von~Neumann J and Morgenstern O (1945) \emph{Theory of games and economic
  behavior}.
\newblock Princeton University Press Princeton, NJ.

\bibitem[{{Wulfmeier} et~al.(2016){Wulfmeier}, {Wang} and
  {Posner}}]{wulfmeier2016maxentirl}
{Wulfmeier} M, {Wang} DZ and {Posner} I (2016) Watch this: Scalable
  cost-function learning for path planning in urban environments.
\newblock In: \emph{2016 IEEE/RSJ International Conference on Intelligent
  Robots and Systems (IROS)}. pp. 2089--2095.

\bibitem[{Ziebart et~al.(2008)Ziebart, Maas, Bagnell and
  Dey}]{ziebart2008maximum}
Ziebart BD, Maas A, Bagnell JA and Dey AK (2008) Maximum entropy inverse
  reinforcement learning.
\newblock In: \emph{Proceedings of the 23rd National Conference on Artificial
  Intelligence - Volume 3}, AAAI'08. AAAI Press.
\newblock ISBN 978-1-57735-368-3, pp. 1433--1438.
\newblock \urlprefix\url{http://dl.acm.org/citation.cfm?id=1620270.1620297}.

\bibitem[{Zurek et~al.(2021)Zurek, Bobu, Brown and
  Dragan}]{zurek2021situational}
Zurek M, Bobu A, Brown DS and Dragan AD (2021) Situational confidence
  assistance for lifelong shared autonomy.

\end{thebibliography}

\clearpage

\appendix

\section{Method Details}

\subsection{Incorporating Relative Values in Training}
\label{app:relative_values}

\change{Concretely, given start state $\sysstate_0$, a relative value $v_0$ acts as a modifier for what $\phi_{\psi}(\sysstate_{0})$ should be relative to  $\phi_\psi$'s minimum value. 
If we consider the maximum feature value to be $\phi_\psi^{max}$ and the minimum one $\phi_\psi^{min}$, we can define the feature range $\phi_\psi^{range} = \phi_\psi^{max} - \phi_\psi^{min}$. Then, $v_0$ shifts the desired feature value $\phi_{\psi}(\sysstate_{0})$ in proportion to this range. 
When comparing $\phi_{\psi}(\sysstate_{0})$ to the maximum value $\phi_\psi^{max}$, their difference should be $\phi_\psi^{max} - \phi_{\psi}(\sysstate_{0}) = (1-v_{0})*\phi_\psi^{range}$. 
For example, if $v_0 = 0.3$, meaning the trace starts somewhere with a feature value 30\% higher than the minimum, their difference is 70\% of the feature range. If $v_0$ is the default 1, their difference becomes 0, meaning $\phi_{\psi}(\sysstate_{0})$ is the maximum.}

\change{Similarly, a relative value $v_n$ would also shift the feature value of an end state $\sysstate_n$ in proportion to $\phi_\psi^{range}$. This time, when comparing $\phi_{\psi}(\sysstate_{n})$ to the minimum value $\phi_\psi^{min}$, their difference will be $\phi_{\psi}(\sysstate_{n}) - \phi_\psi^{min} = v_{n}*\phi_\psi^{range}$. 
For example, if $v_n = 0.3$, meaning the trace ends somewhere with a feature value 30\% higher than the minimum, their difference is 30\% of the feature range. If $v_n$ is the default 0, their difference becomes 0, meaning $\phi_{\psi}(\sysstate_{n})$ is the minimum.}

\change{To incorporate the relative values $v_0$ and $v_n$ into the training procedure, we have to use them to modify the feature values that the predictor in Eq. \eqref{eq:softmax} is applied to.
Given start states $\sysstate_0$ and $\sysstate_0'$,  instead of comparing $\phi_{\psi}(\sysstate_{0})$ to $\phi_{\psi}(\sysstate_{0}')$ directly, we compare the altered feature values $\phi_{\psi}(\sysstate_{0})' = \phi_{\psi}(\sysstate_{0}) + (1-v_{0})*\phi_\psi^{range}$ and $\phi_{\psi}(\sysstate_{0}')' = \phi_{\psi}(\sysstate_{0}') + (1-v_{0})*\phi_\psi^{range}$. As such, the training loss uses $P(\phi_\psi(\sysstate_0)' > \phi_\psi(\sysstate_0')')$ as a predictor.
Similarly, given end states $\sysstate_n$ and $\sysstate_n'$,  instead of comparing $\phi_{\psi}(\sysstate_{n})$ to $\phi_{\psi}(\sysstate_{n}')$ directly, we compare the altered feature values $\phi_{\psi}(\sysstate_{n})' = \phi_{\psi}(\sysstate_{n}) - v_{n}*\phi_\psi^{range}$ and $\phi_{\psi}(\sysstate_{n}')' = \phi_{\psi}(\sysstate_{n}') - v_{n}*\phi_\psi^{range}$. As such, the training loss uses $P(\phi_\psi(\sysstate_n)' > \phi_\psi(\sysstate_n')')$ as a predictor.}

\section{Experimental Details}

\subsection{Protocols for Feature Trace Collection}
\label{app:FERL_traces}

\begin{figure*}
\centering
\begin{subfigure}{.27\textwidth}
  \centering
  \includegraphics[width=\textwidth,left]{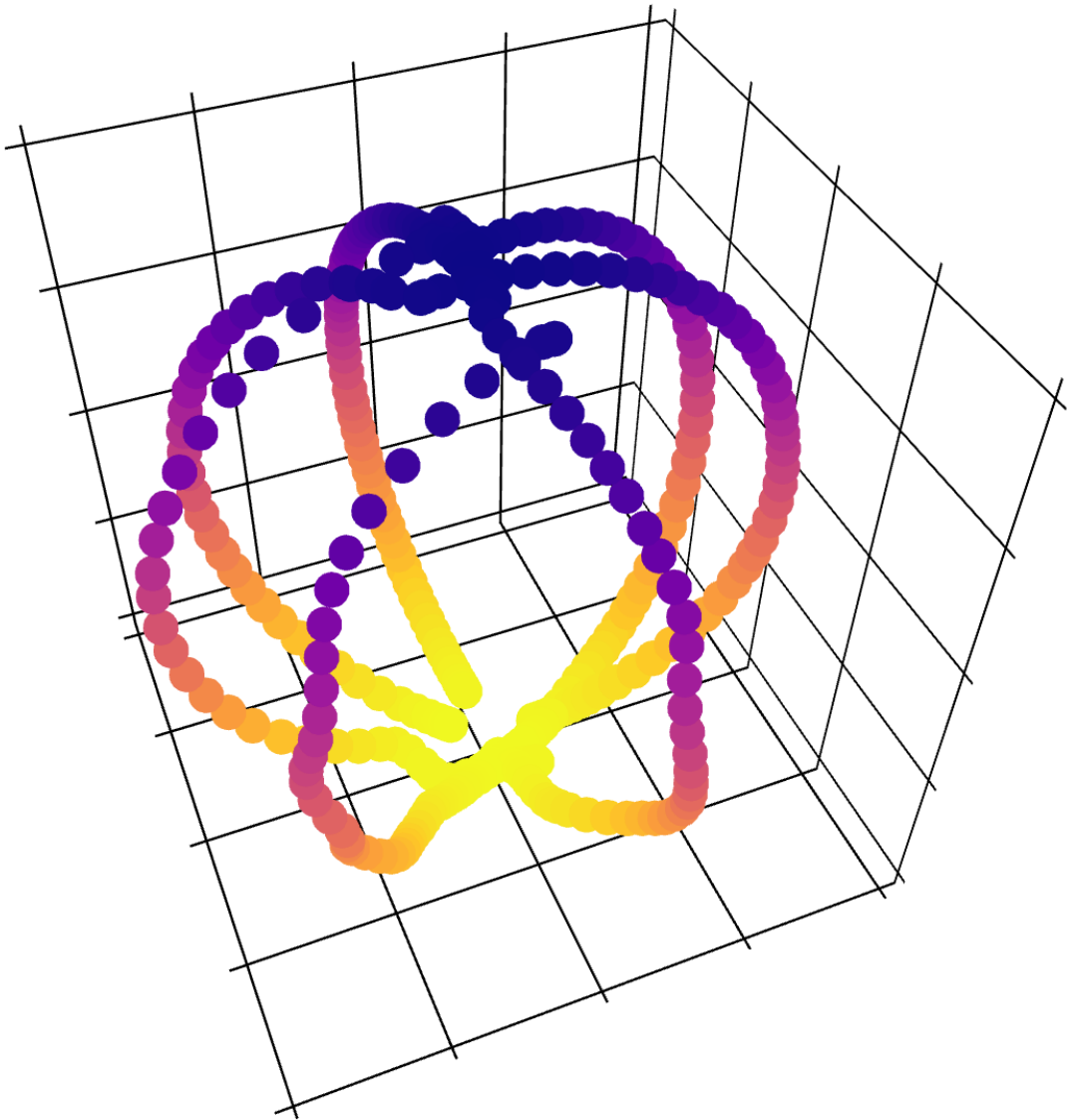}
\end{subfigure}%
\begin{subfigure}{.39\textwidth}
  \centering
  \includegraphics[width=\textwidth,left]{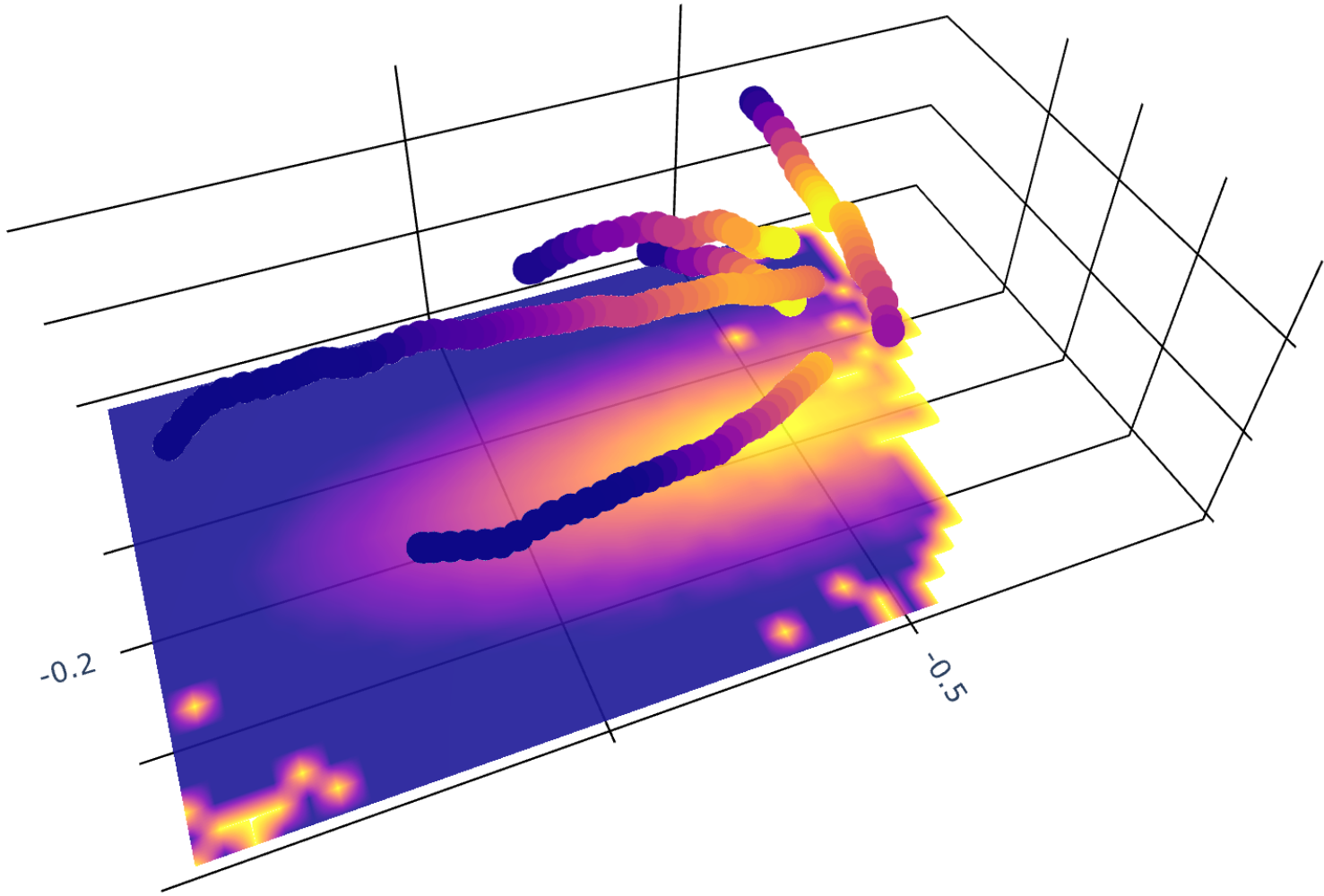}
\end{subfigure}
\begin{subfigure}{.33\textwidth}
  \centering
  \includegraphics[width=\textwidth,left]{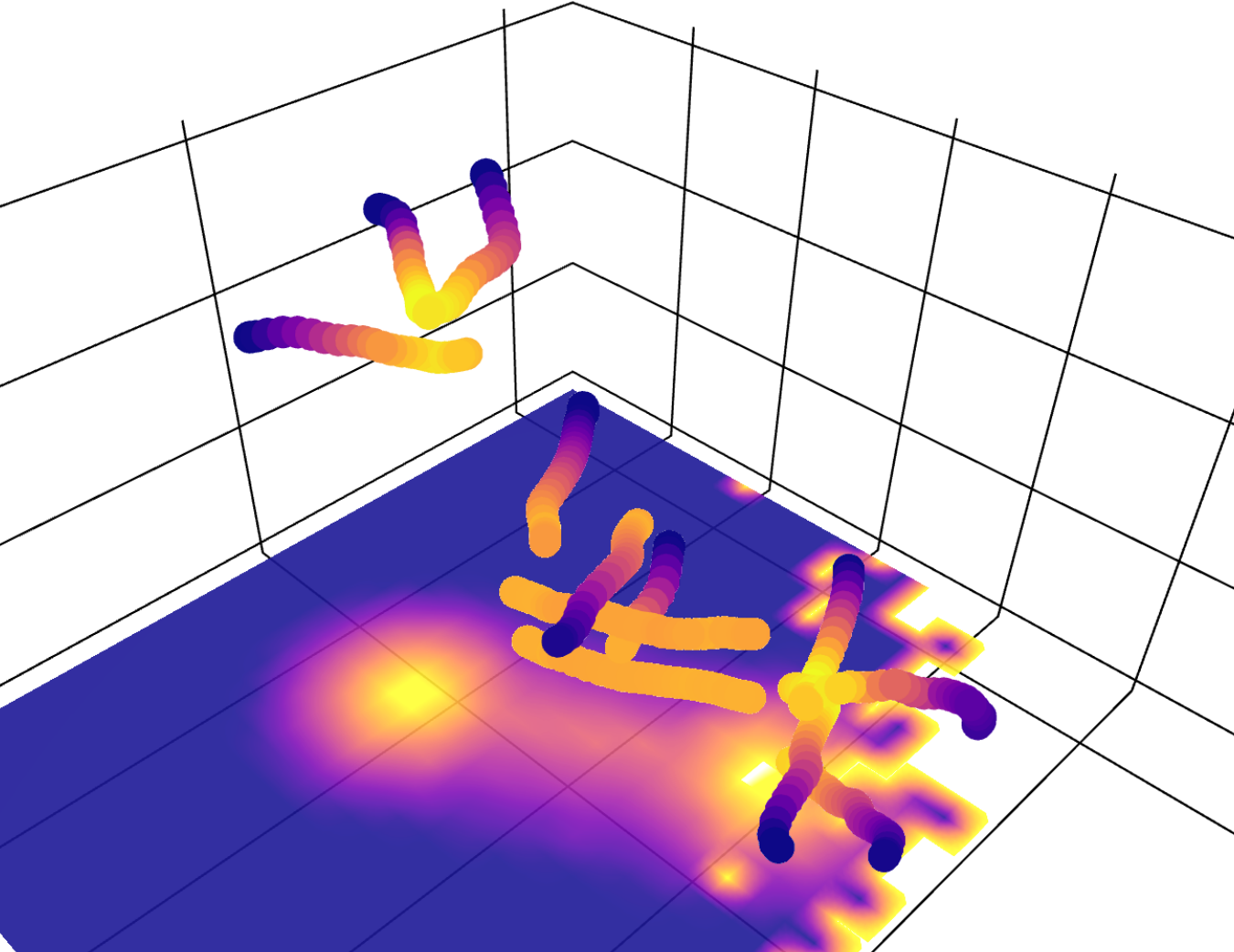}
\end{subfigure}
\caption{(Left) Feature traces for \textit{coffee}. We show the $xyz$ values of the $x$-axis base vector of the \acf{EE} orientation. The traces start with the \ac{EE} pointing downwards and move it upwards. (Middle) Feature traces for \textit{proxemics} with the human at $xy=[-0.2, -0.5]$, with \acf{GT} feature values projected on the $xy$-plane. Some traces are longer than others, to signal that the human dislikes the \ac{EE} being in front of them more than to the sides. (Right) Feature traces for \textit{between objects}, with \ac{GT} feature values projected on the $xy$-plane. Notice a mix of traces teaching about the two objects and about the space between them.}
\label{fig:traces}
\end{figure*}

\begin{figure*}
\centering
\begin{subfigure}{.3\textwidth}
  \centering
  \includegraphics[width=\textwidth,left]{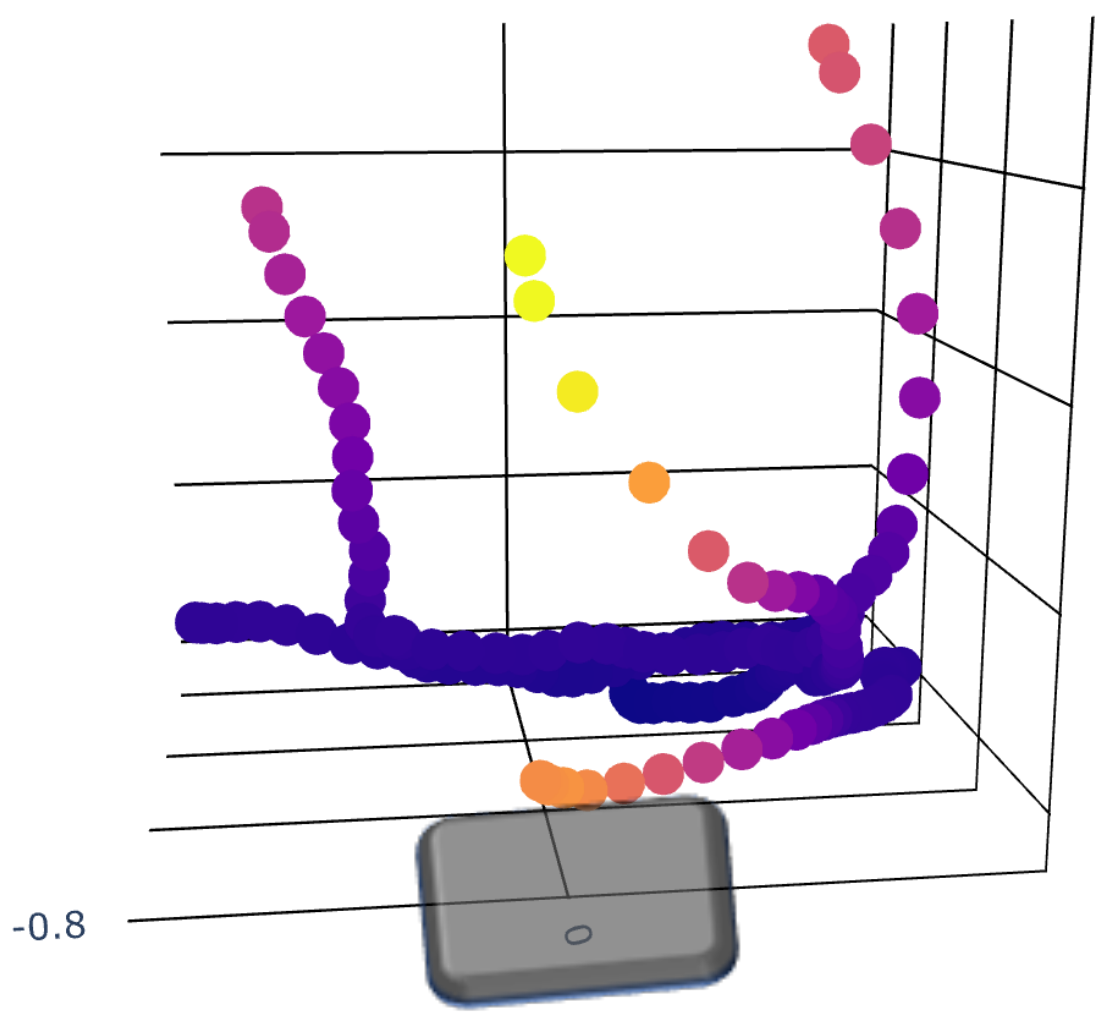}
\end{subfigure}%
\begin{subfigure}{.33\textwidth}
  \centering
  \includegraphics[width=\textwidth,left]{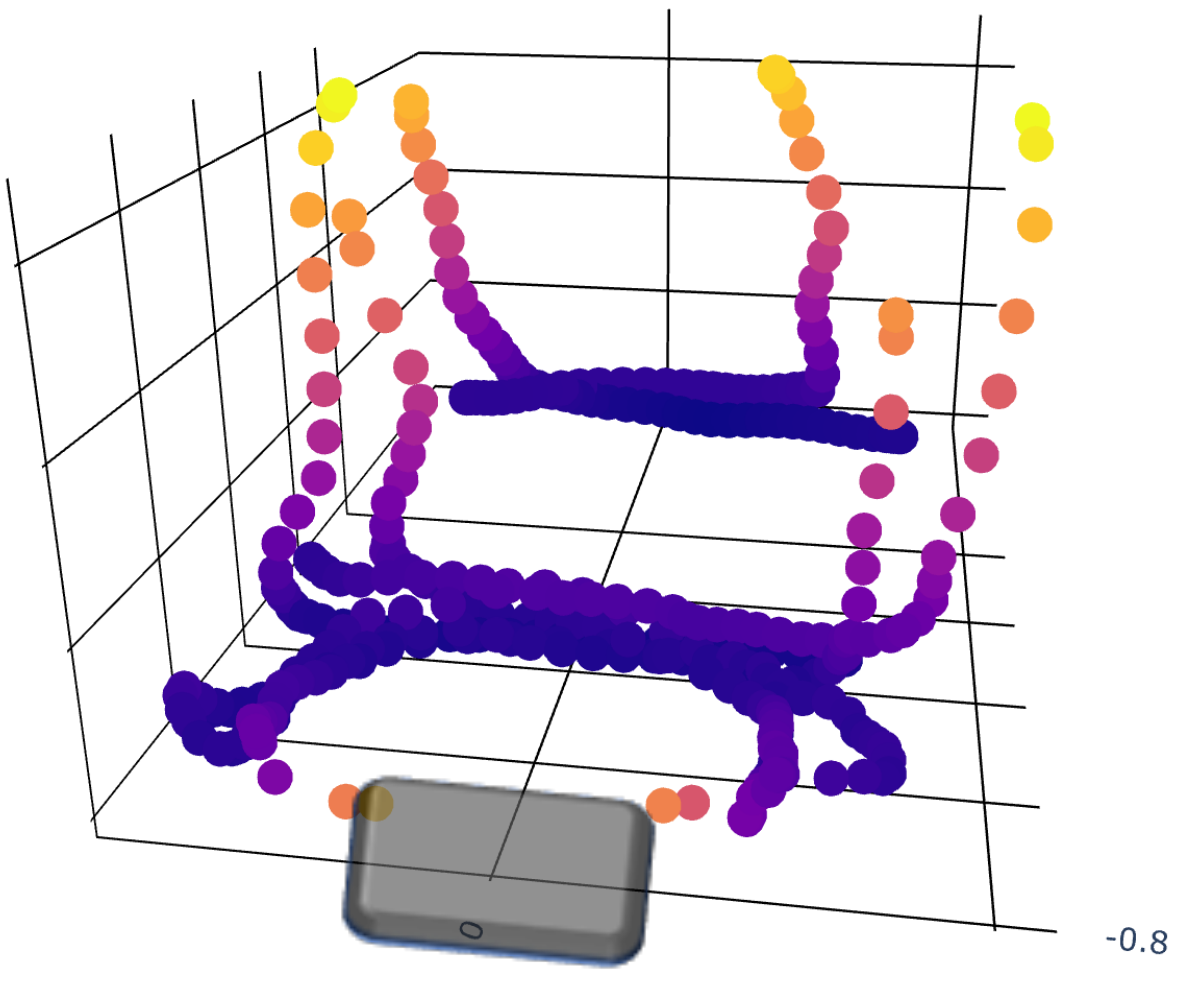}
\end{subfigure}
\begin{subfigure}{.33\textwidth}
  \centering
  \includegraphics[width=\textwidth,left]{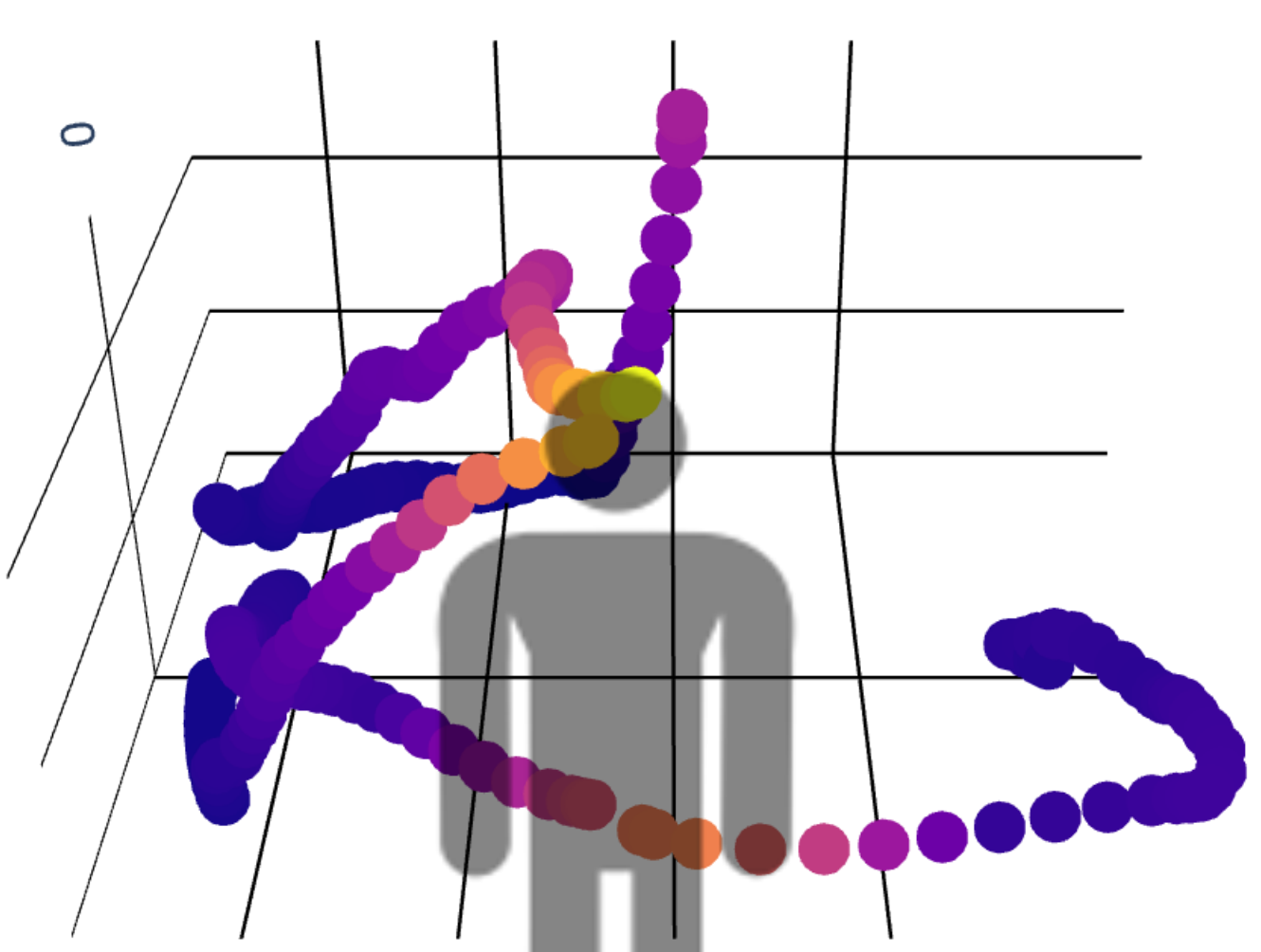}
\end{subfigure}
\caption{A few representative demonstrations collected for \textit{Laptop Missing} (left), \textit{Table Missing} (middle), and \textit{Proxemics Missing} (right). The colors signify the true reward values in each task, where yellow is low and blue is high.}
\label{fig:Demos}
\end{figure*}

In this section, we present our protocol for collecting feature traces for the six features discussed in Sec. \ref{sec:feature_expert_exps}. As we will see, the traces collected from the human only noisily satisfy the assumptions in Sec. \ref{sec:feature_training}. Nevertheless, as we showed in Sec. \ref{sec:feature_expert_exps}, \ac{FERL} is able to learn high quality feature functions.

For \textit{table}, the person teaches that being close to the table, anywhere on the $xy$ plane, is desirable, whereas being far away in height is undesirable. As such, in Fig. \ref{fig:Laptop_Feature_Exp} on the left traces traverse the space from up at a height, until reaching the table. A few different starting configurations are helpful, not necessarily to cover the whole state space, but rather to have signal in the data: having the same trace 10 times would not be different from having it once.

For \textit{laptop}, as described in the text and shown in Fig. \ref{fig:Laptop_Feature_Exp} on the right, the person starts in the middle of the laptop, and moves away a distance equal to the bump radius desired. Having traces from a few different directions and heights helps learn a more distinct feature. For \textit{test laptop location}, the laptop's location at test time is not seen during training. Thus, the training traces should happen with various laptop positions, also starting in the middle and moving away as much distance as desired. 

When teaching the robot to keep the cup upright (\textit{coffee}), the person starts their traces by placing the robot in a position where the cup is upside-down, then moving the arm or rotating the \acf{EE} such that it points upright. Doing this for a few different start configurations helps.
Fig. \ref{fig:traces} (left) shows example traces colored with the true feature values.

When learning \textit{proxemics}, the goal is to keep the \ac{EE} away from the human, more so when moving in front of their face, and less so when moving on their side. As such, when teaching this feature, the person places the robot right in front of the human, then moves it away until hitting the contour of some desired imaginary ellipsis: moving further in front of the human, and not as far to the sides, in a few directions. Fig. \ref{fig:traces} (middle) shows example traces colored with the \acf{GT} feature values.

Lastly, for \textit{between objects} there are a few types of traces, all shown in Fig. \ref{fig:traces} (right). First, to teach a high feature value on top of the objects, some traces need to start on top of them and move away radially. Next, the person has a few options: 1) record a few traces spanning the line between the objects, at different heights, and labeling the start and the end the same; 2) starting anywhere on the imaginary line between the objects and moving perpendicularly away the desired distance, and labeling the start; 3) starting on top of one of the objects, moving towards the other then turning away in the direction orthogonal to the line between the objects.

\subsection{Protocols for Demonstration Collection}
\label{app:MEIRL_demos}

In an effort to make the \ac{ME-IRL} comparison fair, we paid careful attention to collecting informative demonstrations for both reward learning settings in Sec. \ref{sec:online_FERL_exps} and Sec. \ref{sec:offline_FERL_exps}. 

In the online setting, for each unknown feature, we recorded a mix of 20 demonstrations about the unknown feature only (with a focus on learning about it), the known feature only (to learn a reward weight on it), and both of them (to learn a reward weight combination on them). We chose diverse start and goal configurations to trace the demonstrations.

For \textit{Laptop Missing}, we had a mix of demonstrations that start close to the table and focus on going around the laptop, ones that are far away enough from the laptop such that only staying close to the table matters, and ones where both features are considered. Fig. \ref{fig:Demos} (left) shows examples of such demonstrations: the two in the back start far away enough from the laptop but at a high height, and the two in the front start above the laptop at different heights.

For \textit{Table Missing}, we collected a similar set of trajectories, although we had more demonstrations attempting to stay close to the table when the laptop was already far away. Fig. \ref{fig:Demos} (middle) shows a few examples: the two in the back start far away from the laptop and only focus on staying close to the table, a few more start at a high height but need to avoid the laptop to reach the goal, and another two start above the laptop and move away from it.

For \textit{Proxemics Missing}, the most difficult one, some demonstrations had to avoid the person slightly to their side, while others needed to avoid the person more aggressively in the front. We also varied the height and start-goal locations, to ensure that we learned about each feature separately, as well as together. Fig. \ref{fig:Demos} (right) shows a few of the collected demonstrations.

In the offline setting, we took a similar approach to collecting demonstrations. For \textit{One Feature}, we recorded 20 demonstrations starting far from the table and moving close to it, making sure to vary the start and end configurations. For \text{Two Features}, we collected 40 demonstrations (double the amount for two features) with a similar protocol to the \textit{Laptop Missing} and \textit{Table Missing} tasks in the online setting. Lastly, for the \textit{Three Features} task we obtained 60 demonstrations, focusing on each feature separately, every pair of two, and the full combination of three features.

\subsection{Raw State Space Dimensionality}
\label{app:rawstate}

\begin{figure*}
\centering
\begin{subfigure}[b]{1\textwidth}
  \centering
  \includegraphics[width=0.9\linewidth]{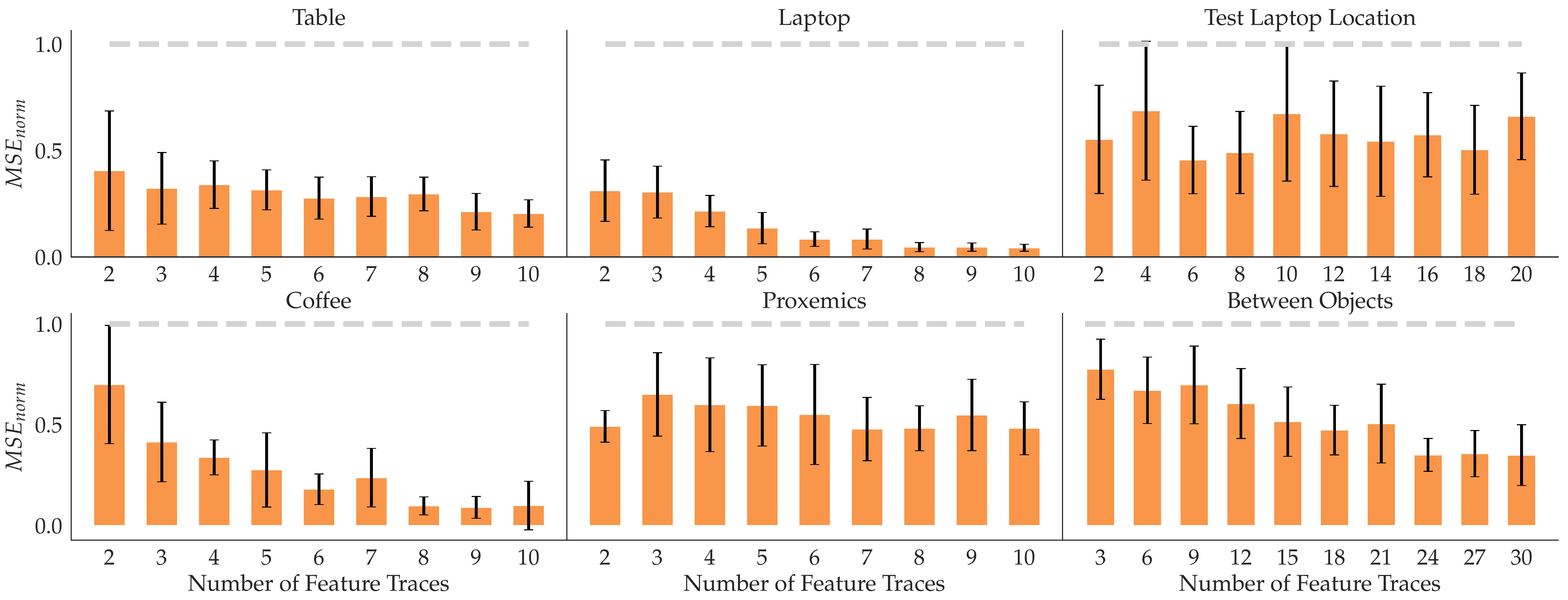}
\end{subfigure}%
\\
\begin{subfigure}[b]{1\textwidth}
  \centering
  \includegraphics[width=0.9\linewidth]{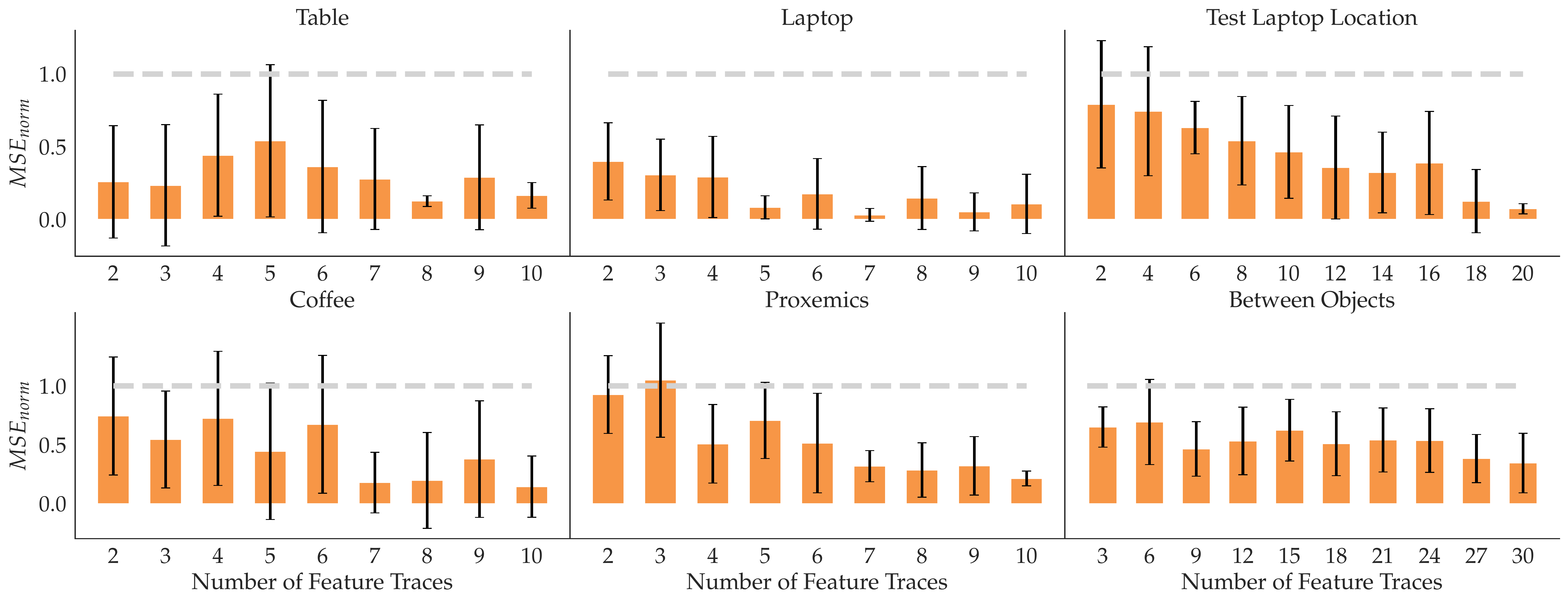}
\end{subfigure}
\caption{Quantitative feature learning results for 36D without (above) and with (below) the subspace selection heuristic. For each feature, we show the $\text{MSE}_{\text{norm}}$ mean and standard error across 10 random seeds with an increasing number of traces (orange) compared to random performance (gray).}
\label{fig:FERL_36_D}
\end{figure*}

Throughout our experiments, we chose a 36D input space made out of 27 Euclidean coordinates ($xyz$ positions of all robot joints and environment objects) and 9 entries in the \ac{EE}'s rotation matrix. We now explain how we chose this raw state space, how spurious correlations across different dimensions can reduce feature learning quality, and how this adverse effect can be alleviated.

First, note that the robot's 7 joint angles and the $xyz$ positions of the objects are the most succinct representation of the state, because the positions and rotation matrices of the joints can be determined from the angles via forward kinematics. With enough data, the neural network should be able to implicitly learn forward kinematics and the feature function on top of it. However, we found that applying forward kinematics a-priori and giving the network access to the $xyz$ positions and rotation matrices for each joint improve both data efficiency and feature quality significantly. In its most comprehensive setting, thus, the raw state space can be 97D (7 angles, 21 $xyz$ coordinates of the joints, 6 $xyz$ coordinates of the objects, and 63 entries in rotation matrices of all joints).


Unfortunately, getting neural networks to generalize on such high dimensional input spaces, especially with the little data that we have access to, is very difficult. Due to the redundancy of the information in the 97D state space, the feature network $\phi_{\psi}$ frequently picks up on spurious correlations in the input space, which decreases the generalization performance of the learned feature. In principle, this issue could be resolved with more diverse and numerous data. Since we want feature learning to be as effortless as possible for the human, we instead opted for the reduced 36D state space, focusing directly on the $xyz$ positions and the \ac{EE} orientation.



Now, as noted in Sec. \ref{sec:feature_expert_exps}, the spurious correlations in the 36D space still made it difficult to train on both the position and orientation subspaces. To better separate redundant information, we devised a heuristic to automatically select the appropriate subspace for a feature. For each subspace, the algorithm first trains a separate network for 10 epochs on half of the input traces and evaluates its generalization ability on the other half using the \ac{FERL} loss. The subspace model with the lower loss (better generalization) is then used for $\phi_{\psi}$ and trained on all traces. We found this heuristic to work fairly well, selecting the right subspace on average in about 85\% of experiments.

To test how well it works in feature learning, we replicated the experiment in Fig. \ref{fig:FERL_MSE} on the 36D state space, both with and without the subspace selection heuristic. A first obvious observation from this experiment is that performing feature learning on separate subspaces (Fig. \ref{fig:FERL_MSE}) results in lower MSEs for all features and $N$ number of traces than learning from all 36 raw states (Fig. \ref{fig:FERL_36_D}). Without the heuristic (Fig. \ref{fig:FERL_36_D} above), we notice that, while spurious correlations in the raw state space are not problematic for some features (\textit{table}, \textit{coffee}, \textit{laptop}, \textit{between objects}), they can reduce the quality of the learned feature significantly for \textit{proxemics} and \textit{test laptop location}. Adding our imperfect heuristic (Fig. \ref{fig:FERL_36_D} below) solves this issue, but increases the variance on each error bar: while our heuristic can improve learning when it successfully chooses the correct raw state subspace, feature learning worsens when it chooses the wrong one. 

In practice, when the subspace is not known, the robot could either use this heuristic or it could ask the human which subspace is relevant for teaching the desired feature. While this is a first step towards dealing with correlated input spaces, more work is needed to find more reliable solutions. A better alternative to our heuristic could be found in methods for causal learning, such as Invariant Risk Minimization~\citep{Arjovsky2019InvariantRM}. We defer such explorations to future work.

\subsection{User Study Instructions}
\label{app:user_instructions}

\change{The familiarization phase of the user study is crucial for making sure our participants are equipped to provide pedagogic feature traces.
To properly train our participants, we provided them with an instruction video and a user manual prior to the study. 
The manual outlined the task they were going to be trained on (\textit{human}), how features are visualized in the study, how to utilize the simulator interface, how to give feature traces in practice, and provided visual examples of traces that lead to high quality and low quality teaching.
The user video essentially followed the outline of the manual, but we found that it provided a more practical illustration of the interface and the teaching procedure. If interested in the instruction video, see \url{https://youtu.be/y36hhb9DI24}.}

\change{During the study, the familiarization phase was 10 minutes long and it gave participants the opportunity to try out the instructions from the manual in practice. 
First, a window appears visualizing the feature on 10,000 states sampled in the reachable set of the robot. We explain verbally what the feature represents and how that definition ties into the 3D visualization. This step was crucial to ensure that all participants have a standardized understanding on the features they teach.
After closing this window, the simulator interface opens up for training the feature.
Because this was a familiarization phase, we guided the participants through the steps, answered questions about the simulator interface, and explained how to give diverse and pedagogic feature traces. Once the algorithm trained the \textit{human} feature, the 3D visualization of the learned feature along the given traces appeared.
We walked the participants through what went right and wrong in their teaching, and explained how they could have improved their traces. We offered them the opportunity to try again, but all users chose to begin the study. Once the second phase of the study began, we offered participants no feedback on their teaching.}

\section{Implementation Details}
We report details of our training procedures, as well as any hyperparameters used. We tried a few different settings but no extensive hyperparameter tuning was performed. Here we present the settings that worked best for each method. 
The code can be found at \href{https://github.com/andreea7b/FERL}{https://github.com/andreea7b/FERL}.

\subsection{Feature Learning Training Details}
\label{app:FERL_implementation}

The feature function $\phi_{\psi}(\sysstate)$ is approximated by a 2 layer, 64 hidden units neural network. We used a leaky ReLu non-linearity for all but the output layer, for which we used the softplus non-linearity. We normalized the output of $\phi_{\psi}(\sysstate)$ every epoch by keeping track of the maximum and minimum output logit over the entire training data.
Following the description in Sec. \ref{sec:feature_training}, the full dataset consists of $|\mathcal{T}| = \sum_{i=1}^{N}\binom{(n^i+1)}{2} + 2 \binom{N}{2}$ tuples, where the first part is all tuples encoding monotonicity and the second part is all tuples encouraging indistinguishable feature values at the starts and ends of traces. Note that $\sum_{i=1}^{N}\binom{(n^i+1)}{2} >> 2 \binom{N}{2}$, hence in the dataset there are significantly fewer tuples of the latter than the former type. This imbalance can lead to the training converging to local optima where the start and end values of traces are significantly different across traces. 
\change{We addressed this by using data augmentation on the equivalence tuples (adding each tuple 5 times) and weighing the loss $L_{equiv}$ by a factor of 10, i.e. we picked $\lambda=10$ in Eq. \eqref{eq:loss_total}.} We optimized our final loss function using Adam for $K=100$ epochs with a learning rate and weight decay of 0.001, and a batch-size of 32 tuples over all tuples.

\subsection{Online \ac{FERL} Details}
\label{app:online_FERL_implementation}

In the Online FERL implementation of Alg. \ref{alg:E-FeRL}, the robot uses TrajOpt \citep{schulman2013trajopt} to plan a path from the start to the goal configuration using the initial parameters $\weight$, then starts tracking it.  
When a person applies a correction, the robot records the instantaneous deviation at 100Hz frequency until the interaction concludes. Then, the robot uses the first of these deviations to estimate the confidence $\hat\beta$ in its ability to explain the push. If the robot needs to learn a new feature, i.e. $\hat\beta<\epsilon$, it pauses trajectory execution. We used $\epsilon=0.1$ but we did not perform extensive parameter tuning.

After potentially learning the feature using Alg. \ref{alg:feature_learning}, the robot uses its recorded sequence of instantaneous deviations to update $\weight$ and replan $\traj$. 
If the robot did learn a new feature, by now its configuration has changed as a result of collecting feature traces, so we place it at the last recorded configuration before feature learning happened, then resume new trajectory execution.
More details on estimating $\hat\beta$, deforming the trajectory $\traj$ by the correction, and parameters for updating $\weight$ can be found in App. A of \citet{bobu2020quantifying}.

\subsection{Offline \ac{FERL} Details}
\label{app:offline_FERL_implementation}

We optimize the loss in Alg. \ref{alg:S-FeRL} with stochastic gradient descent using a learning rate of 1.0 and $K=50$ iterations.
At each iteration we have to generate a set of near optimal trajectories $D^\weight$ for the current reward. To do so, we take the start and goal pairs of the demonstrations and use TrajOpt \citep{schulman2013trajopt} to generate an optimal trajectory for each start-goal pair, hence $|D^*|=|D^\weight|$. At every iteration,
we estimate the gradient using the full batch of $|D^*|$ demonstration tuples.

\subsection{ME-IRL Training Details}
\label{app:MEIRL_implementation}

We approximate the reward $r_{\omega}(\sysstate)$ by a 2 layer, 128 hidden units neural network, with ReLu non-linearities. 
In the online reward learning experiments in Sec. \ref{sec:online_FERL_expert_exps}, we also add the known features to the output of this network before linearly mapping them to $r_{\omega}(\sysstate)$ with a softplus non-linearity. 
While $D^*$ is given, at each iteration we have to generate a set of near optimal trajectories $D^\omega$ for the current reward $r_{\omega}(\sysstate)$. To do so, we take the start and goal pairs of the demonstrations and use TrajOpt \citep{schulman2013trajopt} to generate an optimal trajectory for each start-goal pair, hence $|D^*|=|D^\omega|$. At each of the 50 iterations, we go through all start-goal pairs with one batch consisting of the $D^*$ and $D^\omega$ trajectories of one randomly selected start-goal pair from which we estimate the gradient as detailed in Sec. \ref{sec:online_FERL_expert_exps}. We optimize the loss with Adam using a learning rate and weight decay of 0.001.

\begin{figure}
\includegraphics[width=.45\textwidth]{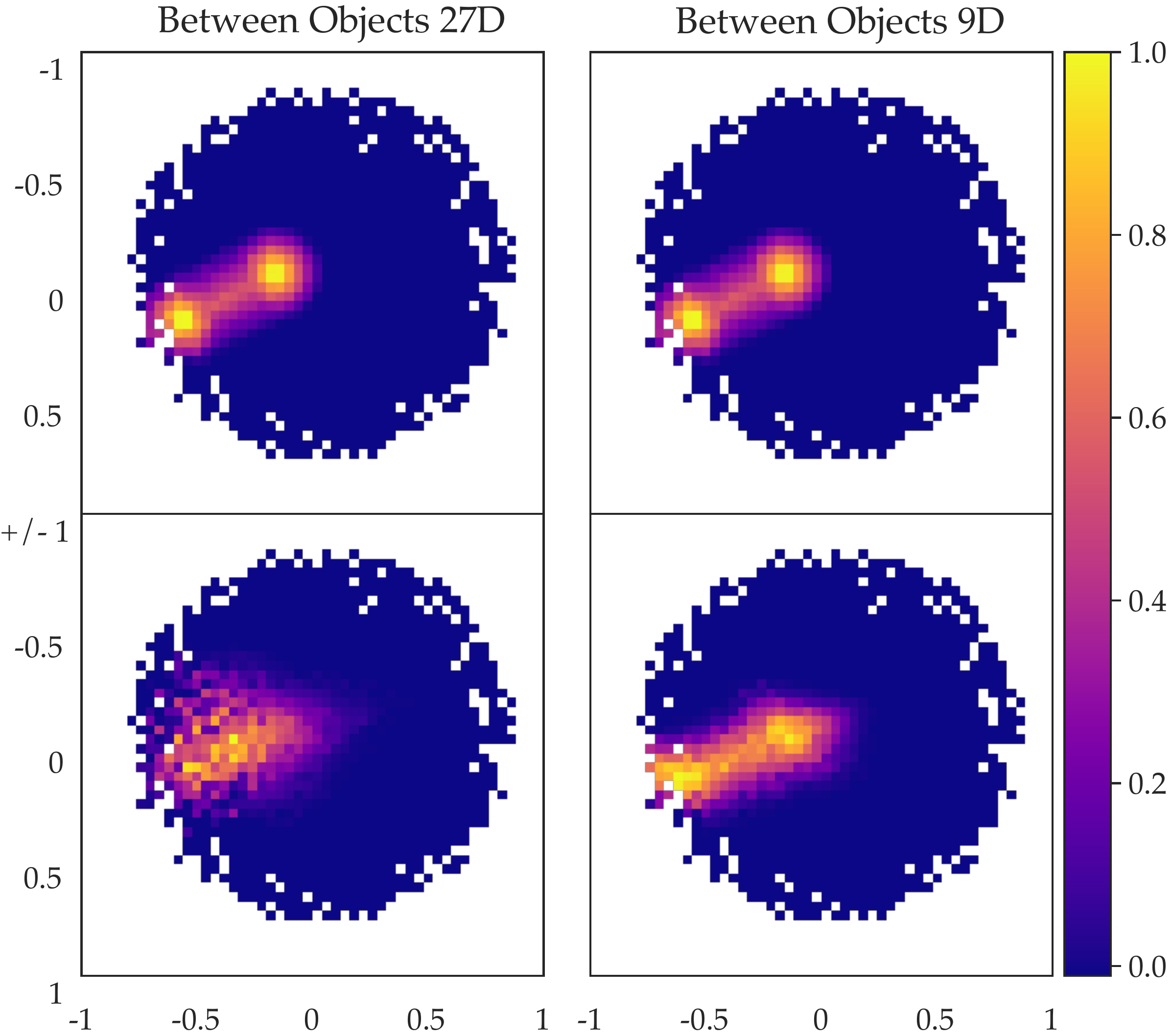}
\centering
\caption{The \textit{between objects} feature. (Left) Using a 27D highly correlated raw state space ($xyz$ positions of all robot joints and objects), the learned feature (Down) does not capture the fine-grained structure of the ground truth (Up). (Right) When using only 9D ($xyz$ positions of the \ac{EE} and objects), the quality of the learned feature improves.}
\label{fig:VisualComparisonC1}
\end{figure}

\begin{figure*}
\centering
\begin{subfigure}{.33\textwidth}
  \centering
  \includegraphics[width=\textwidth,left]{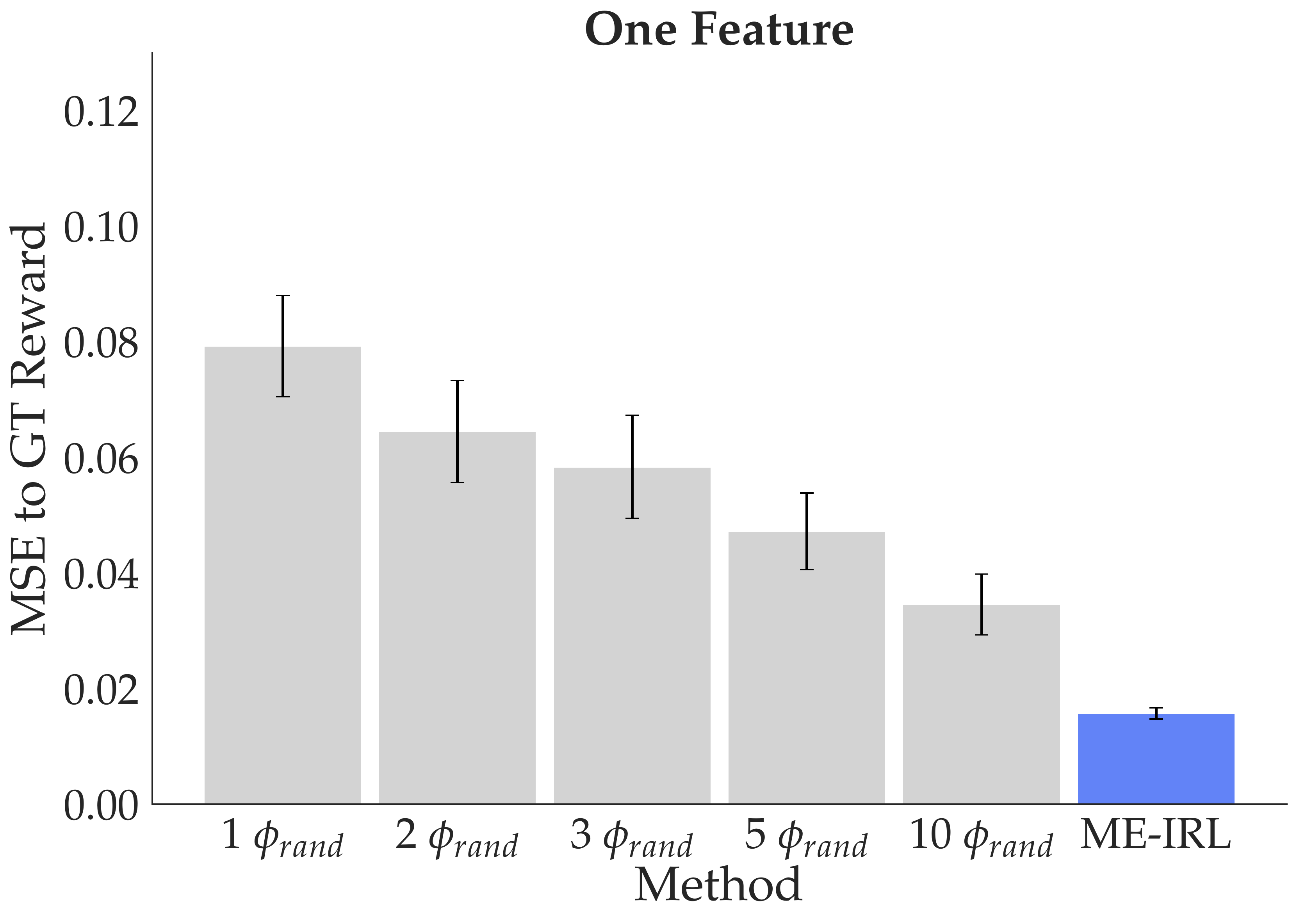}
\end{subfigure}%
\begin{subfigure}{.33\textwidth}
  \centering
  \includegraphics[width=\textwidth,left]{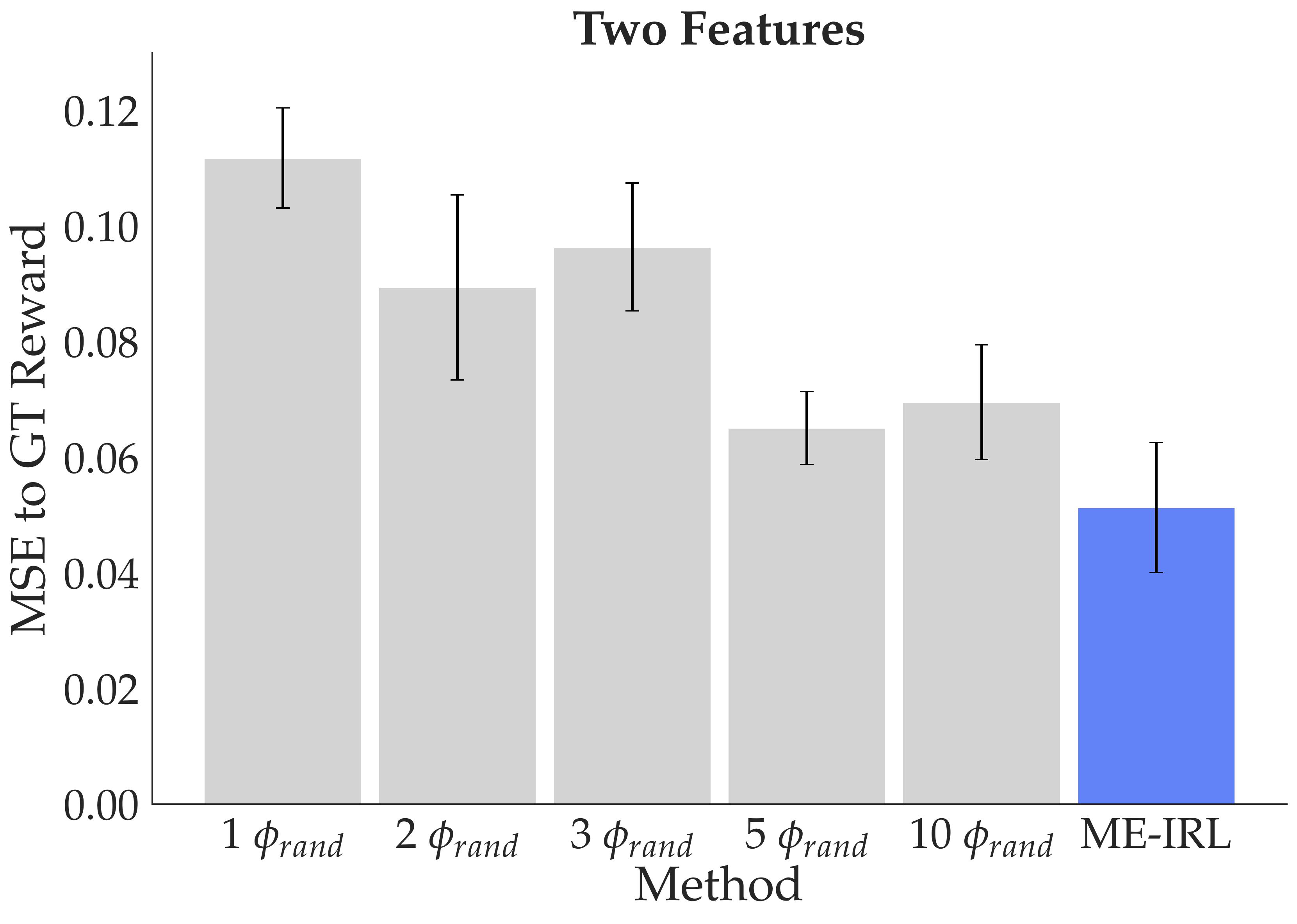}
\end{subfigure}%
\begin{subfigure}{.33\textwidth}
  \centering
  \includegraphics[width=\textwidth,left]{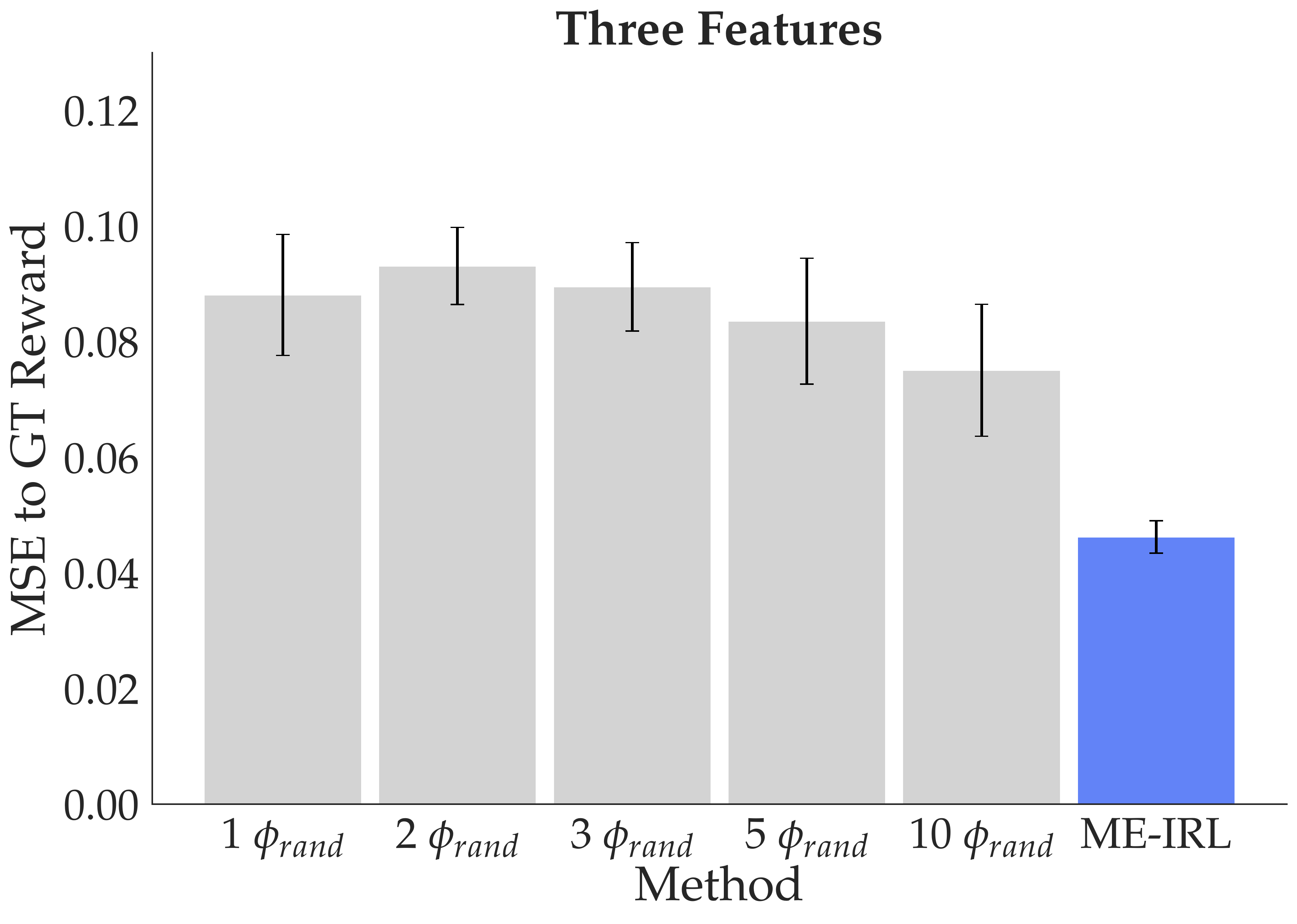}
\end{subfigure}
\caption{\change{MSE of shallow \ac{ME-IRL} with 1, 2, 3, 5, and 10 random features (gray) and deep \ac{ME-IRL} (blue) to \ac{GT} reward for \textit{One Feature} (Left), \textit{Two Features} (Middle), and \textit{Three Features} (Right). The deep \ac{ME-IRL} variant outperforms the shallow one.}}
\label{fig:random_vs_MEIRL_MSE}
\end{figure*}

\section{Additional Results}

\subsection{\textit{Between Objects} with 9D State Space}
\label{app:betweenobjects}

In Fig. \ref{fig:FERL_Qual} we saw that for \textit{between features}, while \ac{FERL} learned the approximate location of the objects to be avoided, it could not learn the more fine-grained structure of the ground truth feature. This could be an artefact of the spurious correlations in the high dimensional state space. To analyze this result, we trained a network with only the dimensions necessary for learning this feature: the $xyz$ positions of the \ac{EE} and of the two objects. The result in Fig. \ref{fig:VisualComparisonC1} illustrates that, in fact, our method is capable of capturing the fine-grained structure of the ground truth; however, more dimensions in the state space induce more spurious correlations that decrease the quality of the features learned.

\subsection{Baseline Comparison}
\label{app:randomfeatures}

\change{Throughout our reward learning experiments in Secs. \ref{sec:online_FERL_exps} and \ref{sec:offline_FERL_exps}, we compare \ac{FERL} to a deep implementation of \ac{ME-IRL}. 
Here, we investigate using a shallow variant as a baseline instead, where the reward is modeled as a linear combination of random features $\phi_{rand}$.
We model each random feature as frozen randomly initialized neural networks with 2 layers and 256 units.
This comparison should tell us whether using a deep architecture for reward learning provides an advantage when compared to simply learning rewards on top of random transformations of the input space.}

\change{For this experiment, we looked at three tasks where the \ac{GT} reward is increasingly complex in the number of features: \textit{One Feature}, \textit{Two Features}, and \textit{Three Features} from Sec. \ref{sec:offline_FERL_exps}. We compare deep \ac{ME-IRL} to a shallow implementation that has access to 1, 2, 3, 5, or 10 random features in the linear layer.
We use as a metric the same \textit{Reward Accuracy} metric as in Secs. \ref{sec:online_FERL_exps} and \ref{sec:offline_FERL_exps}, which computes the \ac{MSE} between the normalized learned rewards and the \ac{GT}.}

\change{Fig. \ref{fig:random_vs_MEIRL_MSE} illustrates the differences between the 5 shallow variants (gray) and the deep \ac{ME-IRL} (blue). 
For \textit{One Feature}, the easiest case where the reward relies only on the \textit{table} feature, we see that increasing the number of random features does improve the performance, but never beyond that of deep \ac{ME-IRL}.
The same trend disappears in the other, more complex, two cases. 
Overall, the deep variant consistently outperforms shallow \ac{ME-IRL} with any of the tested number of random features, so we chose deep \ac{ME-IRL} as the representative baseline in our main experiments.}




\end{document}